\documentclass[10pt,twocolumn,letterpaper]{article}

\usepackage{iccv}
\usepackage{times}
\usepackage{epsfig}
\usepackage{graphicx}
\usepackage{amsmath}
\usepackage{amssymb}

\usepackage{subcaption}
\usepackage{pgffor}
\usepackage{makecell}
\usepackage{tabularx}
\usepackage{arydshln}
\usepackage{multirow}
\usepackage{multicol}
\usepackage{booktabs}
\usepackage[accsupp]{axessibility}  

\usepackage{graphicx}
\usepackage{tikz}
\usepackage{gauss}

\usepackage{authblk}
\newcommand{\email}[1]{\texttt{\small{\{#1\}@samsung.com}}}

\usepackage{color}

\definecolor{darkgreen}{RGB}{30,150,30}
\definecolor{darkblue}{RGB}{0,0,127}
\definecolor{darkyellow}{RGB}{171,133,0}
\definecolor{darkred}{RGB}{180,20,20}
\definecolor{darkmagenta}{RGB}{127,0,127}
\definecolor{darkcyan}{RGB}{0,127,127}

\newcommand{\ourwork}{\text{ADIEE}}

\usepackage[pagebackref=true,breaklinks=true,letterpaper=true,colorlinks,bookmarks=false]{hyperref}

\iccvfinalcopy 

\def\iccvPaperID{5440} 

\ificcvfinal\fi

\begin{document}

\title{\ourwork: \underline{A}utomatic \underline{D}ataset Creation and Scorer for\\ Instruction-Guided \underline{I}mage \underline{E}diting \underline{E}valuation}


\author[1]{Sherry X. Chen\thanks{Work done during an internship at Samsung AI Center Mountain View}}
\author[2]{Yi Wei}
\author[2]{Luowei Zhou\thanks{Corresponding author}}
\author[2]{Suren Kumar}

\affil[1]{University of California, Santa Barbara}
\affil[2]{AI Center-Mountain View, Samsung Electronics}

\affil[ ]{\texttt{\small{xchen774@ucsb.edu}}\hspace{30pt}\email{yi.wei1,\hspace{5pt}luowei.zhou,\hspace{5pt}suren.kumar}}

\maketitle
\ificcvfinal\thispagestyle{empty}\fi

\begin{abstract}
Recent advances in instruction-guided image editing underscore the need for effective automated evaluation. While Vision-Language Models (VLMs) have been explored as judges, open-source models struggle with alignment, and proprietary models lack transparency and cost efficiency. Additionally, no public training datasets exist to fine-tune open-source VLMs, only small benchmarks with diverse evaluation schemes. To address this, we introduce \ourwork{}, an automated dataset creation approach which is then used to train a scoring model for instruction-guided image editing evaluation. We generate a large-scale dataset with over 100K samples and use it to fine-tune a LLaVA-NeXT-8B model modified to decode a numeric score from a custom token. The resulting scorer outperforms all open-source VLMs and Gemini-Pro 1.5 across all benchmarks, achieving a 0.0696 (+17.24\%) gain in score correlation with human ratings on AURORA-Bench, and improving pair-wise comparison accuracy by 4.03\% (+7.21\%) on GenAI-Bench and 4.75\% (+9.35\%) on AURORA-Bench, respectively, compared to the state-of-the-art. The scorer can act as a reward model, enabling automated best edit selection and model fine-tuning. Notably, the proposed scorer can boost MagicBrush model's average evaluation score on ImagenHub from 5.90 to 6.43 (+8.98\%). Our code and models are available at \href{https://github.com/SherryXTChen/ADIEE.git}{https://github.com/SherryXTChen/ADIEE.git}.
\end{abstract}

\section{Introduction}

\begin{figure}
    \centering
    \small
    \setlength{\tabcolsep}{0pt}
    \begin{tabular}{cc}
    \includegraphics[width=0.4\linewidth]{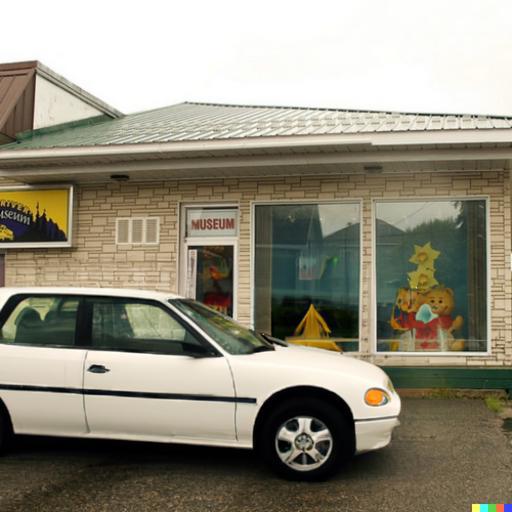} & \includegraphics[width=0.4\linewidth]{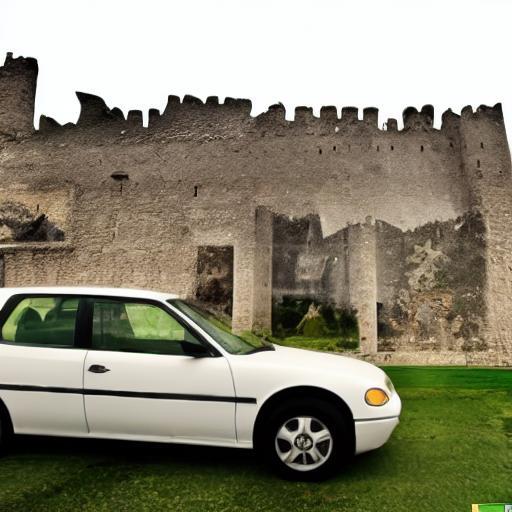}\\    
    \multicolumn{2}{l}{\makecell[l]{
    Can you rate how successful the edit instruction ``edit the\\ background by removing the museum and placing a castle''\\ have been executed from the first image to the second image\\ with a score from 0 to 10?}
    } \\

    \multicolumn{1}{l}{\textbf{Human:} 6.38} & \multicolumn{1}{l}{\hspace{13pt}\textbf{\ourwork{} (Ours):} 6.09} \\

    \hline\hline

    \multicolumn{2}{l}{\hspace{1.8em} Input \hspace{2.0em} MagicBrush \hspace{1.9em} Ours \hspace{3.9em} GT} \\
    \multicolumn{2}{c}{\includegraphics[width=0.9\linewidth, trim=0 30px 0 30px, clip]{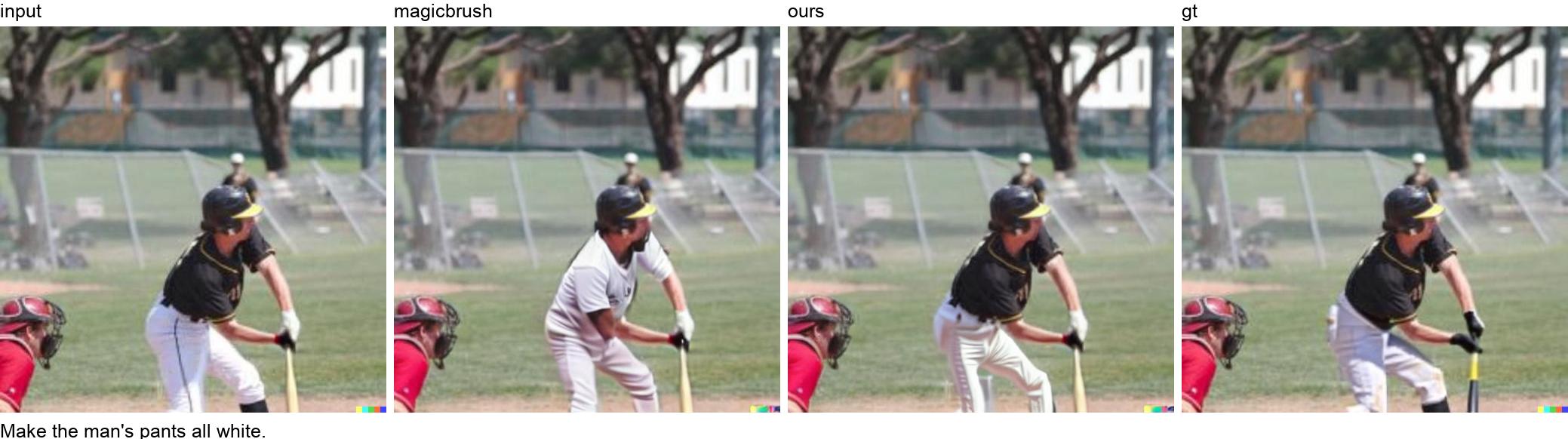}} \\
    \multicolumn{2}{l}{Edit instruction: ``Make the man's \underline{pants} all white''} \\
    \end{tabular}
    \vspace{-0.1in}
    \caption{\textbf{Top:} We fine-tune LLaVA-Next-8B~\cite{liu2024llavanext} on an image editing evaluation dataset automatically generated using our proposed methods to achieve good alignment with human judgment (faces are blocked due to privacy concerns). \textbf{Bottom:} The trained image editing quality scorer can also function as a reward model to fine-tune and enhance the editing capabilities of MagicBrush~\cite{zhang2024magicbrush}.}
    \label{fig:teaser}
\end{figure}

Recent advances in instruction-guided image editing~\cite{brooks2023instructpix2pix,zhang2024magicbrush,krojer2024aurora,zhao2024ultraedit,mirzaei2025watch,li2024zone} have enabled intuitive image modifications via natural language. These developments highlight the growing need for effective automated evaluation. Traditional metrics such as MSE, PSNR, SSIM~\cite{wang2004image}, and perceptual scores like LPIPS~\cite{zhang2018unreasonable} fall short in capturing semantic alignment. More recent approaches leverage vision models like CLIP~\cite{radford2021learning} and DINO~\cite{caron2021emerging,oquab2023dinov2}, yielding metrics such as CLIP score~\cite{hessel2021clipscore}, CLIP similarity~\cite{radford2021learning}, CLIP directional similarity~\cite{brooks2023instructpix2pix}, and DINO similarity~\cite{caron2021emerging}.

Traditional metrics, CLIP similarity and DINO similarity solely rely on ground-truth outputs from image editing benchmarks and overlook textual information. While CLIP score and CLIP directional similarity incorporate both text and images, they are limited to image descriptions/prompts rather than edit instructions. Additionally, prior studies highlight limitations of the CLIP text model, such as difficulties understanding long text~\cite{zhang2024long} and compositional relationships among objects and attributes~\cite{yuksekgonul2022and}, reducing its robustness for evaluation tasks.

To this end, recent work have explored Vision-Language Models (VLMs)~\cite{achiam2023gpt,team2024gemini,liu2024visual,liu2024llavanext,Qwen-VL,li2023blip,dai2023instructblip,fuyu-8b,wang2023cogvlm,awadalla2023openflamingo} as judges~\cite{ku2023viescore,jiang2024genai}, leveraging their abilities to process long text and images jointly with richer semantic understanding. However, open-source VLMs often struggle to align with human judgment, likely due to limitations in training data and computational resources. While proprietary models like ChatGPT~\cite{achiam2023gpt} and Gemini~\cite{team2024gemini} perform better, they lack transparency, customizability, and cost efficiency. 

A key obstacle in developing a robust image editing scorer is acquiring suitable training data. Prior approaches often rely on human-annotated ground-truth labels~\cite{richhf,lin2024evaluating,xu2024imagereward,he2024videoscore,kirstain2023pick,wu2023human,wu2023humanv2,zhang2024learning}, which are costly and limit dataset scalability. Others use proprietary models to generate labels~\cite{wei2024omniedit,wu2025multimodal}, but this approach is expensive for large-scale datasets and inherently constrains fine-tuned VLM performance to that of the proprietary models. Given these challenges, we introduce \ourwork, consisting of automated dataset creation approaches and scorer for instruction-guided image editing evaluation (Fig.\ref{fig:teaser} (top)). 

Specifically, we leverage resources from the image editing space, including instruction-guided training datasets~\cite{ge2024seeddataedit,brooks2023instructpix2pix,zhang2024magicbrush,krojer2024aurora} and text-guided image editing models~\cite{cyclediffusion,couairon2022diffedit,hertz2022prompt,parmar2023zero,brooks2023instructpix2pix,zhang2024magicbrush,krojer2024aurora}. A key observation is that image editing training datasets~\cite{brooks2023instructpix2pix,zhang2024magicbrush,ge2024seeddataedit,zhang2024hive} implicitly serve as training datasets for image editing evaluation. These datasets contain two distinct sample types: (1) ground-truth edited images, representing successful edits with high evaluation scores, and (2) input images, which, along with their augmented versions, should receive low scores if treated as edited outputs, as the edit instructions are not successfully executed. The latter is especially important because a common failure mode of image editing models is that they simply output the original input image. 

To introduce more sample diversity, we generate additional outputs using text-guided image editing models~\cite{cyclediffusion,couairon2022diffedit,hertz2022prompt,parmar2023zero,brooks2023instructpix2pix,zhang2024magicbrush,krojer2024aurora} and apply heuristics to assign corresponding scores. Additionally, we incorporate multi-turn edit datasets~\cite{zhang2024magicbrush,ge2024seeddataedit} as an auxiliary training signal, where intermediate edits provide incorrect or partially correct results with varying evaluation scores relative to the final edit.

Using these approaches, we construct a large-scale instruction-guided image editing evaluation dataset with over 100K samples and corresponding scores~\cite{wei2024omniedit,ge2024seeddataedit}. We then fine-tune LLaVA-NeXT-8B~\cite{liu2024llavanext} by expanding its vocabulary with a special token, whose embedding is decoded into the final image editing quality score~\cite{lai2023lisa}. We evaluate the resulting \ourwork{} scorer against proprietary and open-source VLMs across multiple benchmarks, including ImagenHub~\cite{ku2024imagenhub}, GenAI-Bench~\cite{jiang2024genai}, and AURORA-Bench~\cite{krojer2024aurora}, where our method surpasses all open-source VLMs and Gemini-Pro 1.5 across the board. 

Furthermore, our scorer can function as a reward model to enhance image editing models~\cite{zhang2024magicbrush} as illustrated in Fig.~\ref{fig:teaser} (bottom) by conditioning training on quality scores predicted by the scorer~\cite{zhang2024hive}.

In summary, our main contributions are:
\begin{enumerate} 
\item We propose novel approaches for automatically generating instruction-guided image editing evaluation scoring data for training and generate a training dataset with over 100k samples. 
\item We propose a reliable automated evaluation metric for instruction-guided image editing, leveraging a scoring model fine-tuned with the aforementioned dataset.
\item We showcase the effectiveness of an improved image editing model using our scorer as the reward model.
\end{enumerate}

\section{Related Work}

\subsection{Text-guided diffusion-based image editing}

Diffusion models~\cite{song2020denoising,dhariwal2021diffusion,rombach2022high,podell2023sdxl,luo2023latent,esser2403scaling} have set a new benchmark in image generation quality and form the backbone of many text-guided editing methods~\cite{hertz2022prompt, mokady2023null, wu2023uncovering, wallace2023edict, parmar2023zero, tumanyan2023plug}. These methods refine images to align with target prompts by anchoring to the input prompt, using strategies such as attention preservation~\cite{hertz2022prompt,tumanyan2023plug}, attention optimization~\cite{parmar2023zero}, empty-prompt inversion~\cite{mokady2023null}, or coupled inversion~\cite{wallace2023edict} during the process.

Another direction simplifies editing via single-step instructions rather than prompt pairs~\cite{brooks2023instructpix2pix,zhang2024magicbrush,zhang2024hive,zhao2024ultraedit,krojer2024aurora,sheynin2024emu}. InstructPix2Pix (IP2P)~\cite{brooks2023instructpix2pix} initiates this line by fine-tuning GPT-3~\cite{brown2020language} to synthesize prompt pairs and instructions from training data constructed via Prompt-to-Prompt~\cite{hertz2022prompt}. MagicBrush~\cite{zhang2024magicbrush} extends IP2P with a semi-manual dataset combining image masks and DALL$\cdot$E 2 outputs~\cite{ramesh2022hierarchical}. HIVE~\cite{zhang2024hive} introduces user feedback for reward modeling, while Emu-Edit~\cite{sheynin2024emu}, UltraEdit~\cite{zhao2024ultraedit}, and AURORA~\cite{krojer2024aurora} design alternative strategies for high-quality dataset curation.

\begin{figure*}
    \centering
    \includegraphics[width=0.8\linewidth]{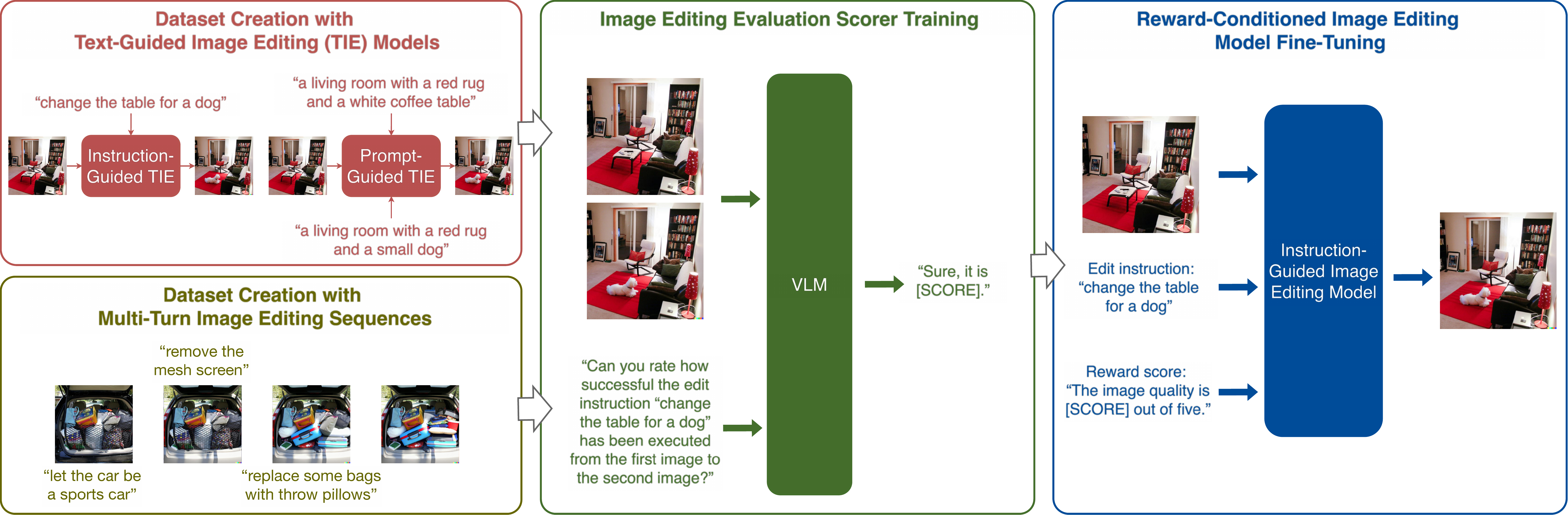}
    \caption{\ourwork{} consists of two image editing evaluation training data generation approaches using text-guided image editing models (Sec.~\ref{sec:data_creation_synthetic}) and multi-turn image editing sequences (Sec.~\ref{sec:data_creation_multiturn}) respectively. It also contains an image editing evaluation scorer trained using our generated data from a vision-language model (VLM) (Sec.~\ref{sec:evaluation_scorer}). Lastly, the trained scorer can act as a reward model to improve the performance of image editing models (Sec.~\ref{sec:reward_conditioned_image_editing}).}
    \vspace{-0.1in}
    \label{fig:overview}
\end{figure*}

\subsection{Text-guided image editing evaluation methods}

With the rise of text-guided image editing, automated evaluation has become increasingly important. Traditional metrics such as MSE, PSNR, SSIM~\cite{wang2004image}, and perceptual scores like LPIPS~\cite{zhang2018unreasonable} rely solely on pixel-wise comparison with ground-truth outputs, neglecting the associated text prompts or instructions.

Recent methods utilize vision(-language) models like CLIP~\cite{radford2021learning} and DINO~\cite{caron2021emerging,oquab2023dinov2} to measure alignment. CLIP score~\cite{hessel2021clipscore} captures semantic similarity between output images and target prompts; CLIP/DINO similarity~\cite{radford2021learning,caron2021emerging} compare image embeddings; CLIP directional similarity~\cite{brooks2023instructpix2pix} aligns textual and visual transformations via latent shifts.

Incorporating text has improved evaluation, yet CLIP struggles with long or compositional text~\cite{zhang2024long,yuksekgonul2022and}. Vision-language models (VLMs) offer stronger joint reasoning, and are increasingly used as evaluators. Some approaches pose binary alignment questions using GPT-4V~\cite{achiam2023gpt}~\cite{ma2024i2ebench}. MEGA-Bench~\cite{chen2024mega-bench} prompts models like GPT-4o~\cite{achiam2023gpt}, Gemini~\cite{team2024gemini}, Claude~\cite{anthropic2024claude3.5sonnet}, and open-source VLMs~\cite{wang2024qwen2,chen2024expanding,li2024llava,smith2023aria,yue2023mmmu} to classify outputs as bad, medium, or good.

ImagenHub~\cite{ku2024imagenhub} and part of AURORA-Bench~\cite{krojer2024aurora} adopt a 0–1 rating scale, while VIEScore~\cite{ku2023viescore} predicts 0–10 scores using off-the-shelf VLMs. GenAI-Bench~\cite{jiang2024genai} and the other part of AURORA-Bench~\cite{krojer2024aurora} use pairwise comparisons to determine human preference, with models like GPT-4o, Gemini, and open-source VLMs~\cite{liu2023llava,laurencon2024idefics2,liu2024llavanext,yao2024minicpm,li2023blip} predicting which output is better.

\subsection{Visual evaluation training data creation}

VLMs are commonly used as zero- or one-shot evaluators for visual tasks, but studies~\cite{chen2024mega-bench,ku2023viescore,jiang2024genai} reveal a consistent performance gap between proprietary and open-source models. Bridging this gap often requires fine-tuning open-source VLMs on tailored datasets.

For image editing, OmniEdit~\cite{wei2024omniedit} generates training data by prompting GPT-4o~\cite{achiam2023gpt} with VIEScore~\cite{ku2023viescore} templates to train InternVL2~\cite{chen2024internvl}. RewardEdit20K~\cite{gu2024multi} applies GPT-4o to score edits across instruction alignment, detail preservation, and generation quality. In text-to-image evaluation, VisionPrefer~\cite{wu2024multimodal} uses GPT-4V~\cite{achiam2023gpt} to rank outputs from Stable Diffusion~\cite{rombach2022high} and Dreamlike Photoreal 2.5~\cite{dreamlike2023photoreal25}, training ImageReward~\cite{xu2024imagereward} on the results. 

Other datasets, including ImageRewardDB~\cite{xu2024imagereward}, RichHF-18K~\cite{richhf}, Human Preference Score v2~\cite{wu2023human}, and IE-Bench~\cite{sun2025ie}, use human annotation instead, which improves quality but limits scalability.

\section{Method}

\begin{figure*}[]
    \centering

    \begin{small}
        \begin{tabbing}
            \hspace{3.6em} \= \hspace{3.3em} \= \hspace{4.8em} \= \hspace{4.7em} \= \hspace{4.1em} \= \hspace{4.1em} \= \hspace{4.9em} \= \hspace{4.0em} \= \hspace{2.8em} \= \hspace{4.9em} \= \hspace{5.4em} \=  \kill
            \> Input \> CycleDiff \> DiffEdit \> Pr2Pr \> P2P-0 \> SDEdit \> T2L \> I-P2P \> MagicBrush \> AURORA \> GT
        \end{tabbing}    
    \end{small}

    \vspace{-0.15in} 
    \foreach \file/\instruction in {
        9_3/{get rid of the bread},
        77_3/{add a plane in the sky},
        283_1/{Replace the red wine with white wine.}
    }{
        \begin{subfigure}{\linewidth}
            \centering
            \includegraphics[width=0.9\linewidth, trim=0 180px 0 30px, clip]{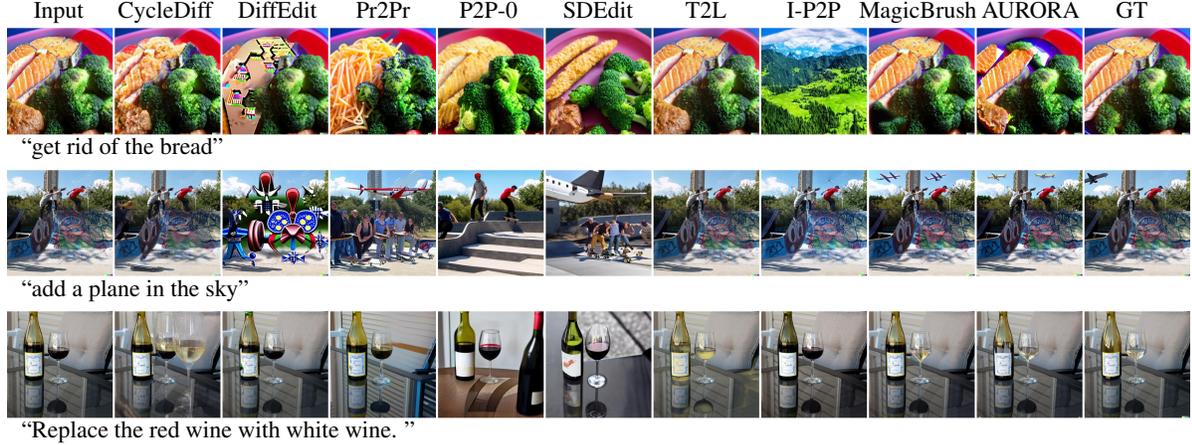}
            \parbox{\linewidth}{
                \parbox{\linewidth}{\vspace{-0.1in}\hspace{3em}{\small``\instruction''}\\\vspace{-0.15in}}
            }
        \end{subfigure}
    }
    \vspace{-0.25in}
    \caption{We use 9 text-guided image editing methods - CycleDiffusion (CycleDiff)~\cite{cyclediffusion}, DiffEdit~\cite{couairon2022diffedit}, Prompt-to-Prompt (Pr2Pr)~\cite{hertz2022prompt}, pix2pix-zero (P2P-0)~\cite{parmar2023zero}, SDEdit~\cite{meng2021sdedit}, Text2LIVE (T2L)~\cite{bar2022text2live}, InstructPix2Pix (I-P2P)~\cite{brooks2023instructpix2pix}, MagicBrush~\cite{zhang2024magicbrush}, and AURORA~\cite{krojer2024aurora} - to generate samples with various editing quality as part of our evaluation training data along with the ground-truth (GT) outputs. For methods that need input and target prompts to perform edits, we trained a VLM and follow the procedure detailed in Fig.~\ref{fig:image_captioning} to get these prompts. For more examples, please refer to the supplementary.}
    \label{fig:data_creation_synthetic}
\end{figure*}

\begin{figure*}[]
    \centering
    \setlength{\tabcolsep}{4pt}
    \small
    \begin{tabular}{ccc}
    & \textbf{Baseline} & \textbf{Ours} \\
    
    \makecell[c]{\includegraphics[width=0.08\linewidth]{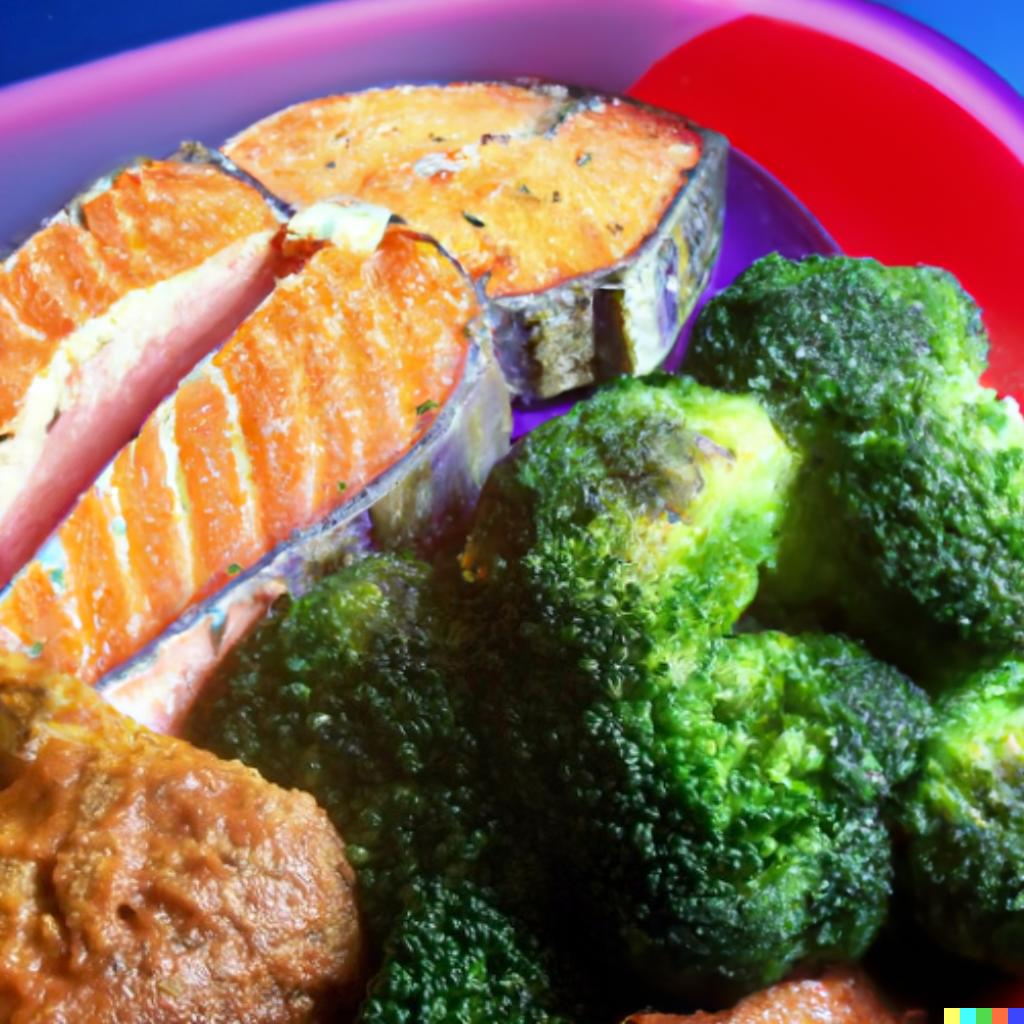}} & \makecell[l]{
    $\bullet$ \textbf{USER:} Describe this image in one sentence. \\ 
    $\bullet$ \textbf{ASSISTANT:} The image shows a plate with a serving\\\hspace{0.5em} of salmon, broccoli, and \textcolor{darkred}{\textbf{a meatball}}.
    } & \makecell[l]{
    $\bullet$ \textbf{USER:} Describe this image based on the edit ``get\\\hspace{0.5em} rid of the bread'' we want to perform on it later. \\
    $\bullet$ \textbf{ASSISTANT:} a plate of food with broccoli and \textcolor{darkgreen}{\textbf{bread}}
    } \\
    
    \makecell[c]{\includegraphics[width=0.08\linewidth]{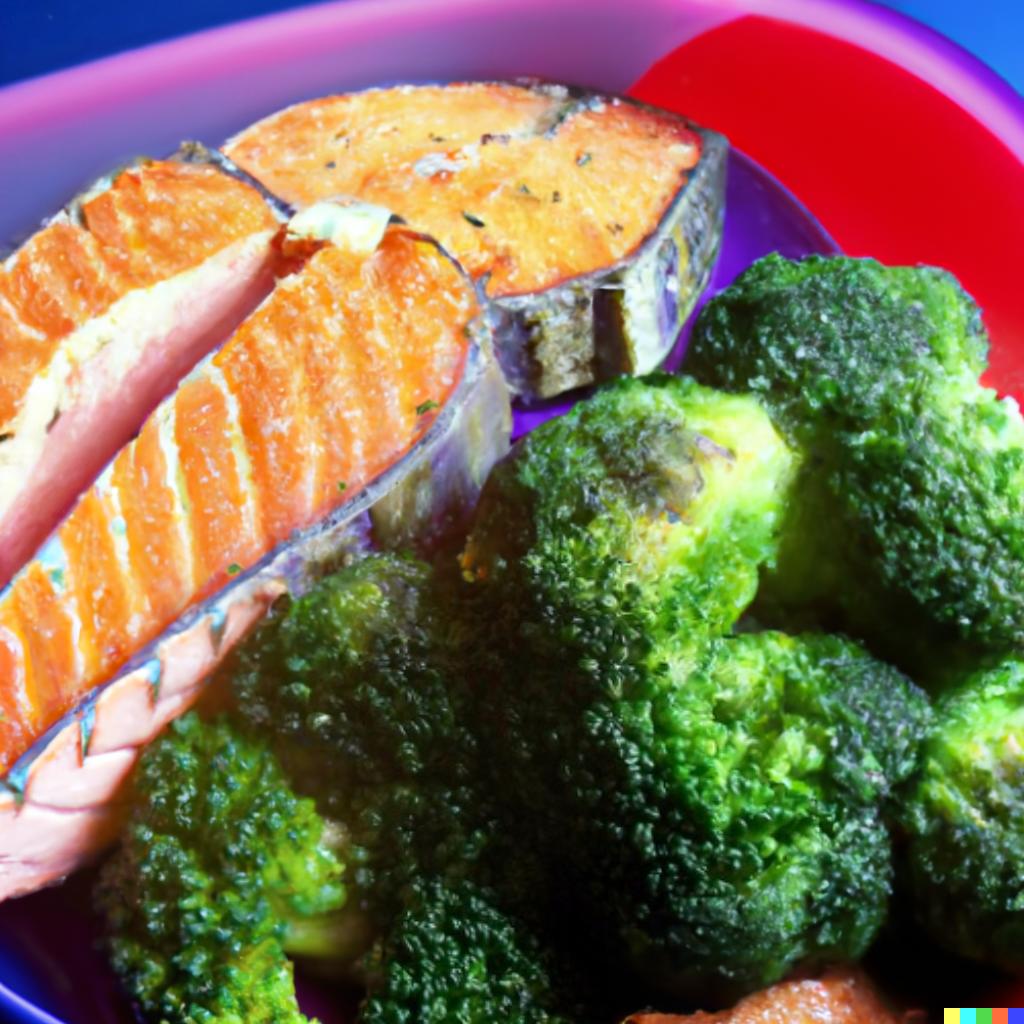}} & \makecell[l]{
    $\bullet$ \textbf{USER:} Describe this image in one sentence. \\ 
    $\bullet$ \textbf{ASSISTANT:} A plate of cooked salmon and broccoli.
    } &  \makecell[l]{
    $\bullet$ \textbf{USER:} Describe the edited image with respect to\\\hspace{0.5em} this image and the edit instruction. \\
    $\bullet$ \textbf{ASSISTANT:} a plate of food with broccoli
    } \\
    
    \makecell[c]{Edit instruction:\\``get rid of the bread''} & \textcolor{darkred}{\makecell[c]{\textbf{$\times$} The difference between the input and target\\ captions does not align with the edit instruction.}}  &  \textcolor{darkgreen}{\makecell{\textbf{$\checkmark$} The difference between the input and target\\ caption aligns with the edit instruction}}
    \end{tabular}
    \vspace{-0.1in}
    \caption{Compared to the input/target prompt generated by a baseline VLM~\cite{wang2024qwen2} with respect to the input (top) and the ground-truth output (bottom) image (Fig.~\ref{fig:data_creation_synthetic} Row 1), where their difference don't align with the edit instruction, our fine-tuned model takes only the input image and generate input-target prompt pairs that align with the edit instruction well.}
    \label{fig:image_captioning}
\end{figure*}

\ourwork{} consists of two image editing evaluation training data generation approaches using text-guided image editing models (Sec.~\ref{sec:data_creation_synthetic}) and multi-turn image editing sequences (Sec.~\ref{sec:data_creation_multiturn}) respectively. In Sec.~\ref{sec:evaluation_scorer}, we discuss the architecture of our image editing evaluation scorer, which is a vision-language model (VLM) trained on our generated training data. Lastly, the trained scorer can act as a reward model to improve the performance of image editing models as detailed in Sec.~\ref{sec:reward_conditioned_image_editing}.



\subsection{Data creation using image editing methods\label{sec:data_creation_synthetic}}

Instruction-guided image editing datasets inherently serve as evaluation datasets. They typically include (1) ground-truth edited images, representing successful edits with high scores, and (2) input images, which, if used as outputs, reflect failed edits with low scores. To align with existing evaluation benchmarks~\cite{ku2023viescore,krojer2024aurora} that use quantized scores of 0 (fail), 0.5 (partial), and 1 (success), we generate samples accordingly.

Specifically, we apply various text-guided editing methods~\cite{cyclediffusion,couairon2022diffedit,hertz2022prompt,parmar2023zero,meng2021sdedit,bar2022text2live,brooks2023instructpix2pix,zhang2024magicbrush} to the MagicBrush dataset~\cite{zhang2024magicbrush} (Fig.~\ref{fig:data_creation_synthetic}). Some methods require input/target captions, which are typically absent in instruction-guided datasets. While VLMs can generate captions, they often misalign with the edit intent (Fig.~\ref{fig:image_captioning}). To address this, we manually annotate a small subset of input images and edit instructions with input/target prompts, then fine-tune a VLM~\cite{wang2024qwen2} to generate instruction-aligned captions (Fig.~\ref{fig:image_captioning}). This prompt refinement is essential for consistent evaluation and fair method comparison.

\noindent\textbf{Negative samples.} A key aspect of score assignment is identifying failed edits. Since input images represent a special case of failure, we use them as anchors to compare against generated samples. Given an input/original image $I^o$, the edit instruction $p$, the ground-truth output $I^e$, the edited output $\hat{I^e}$, the input image caption $p^o$ and the target output caption $p^e$, there are two properties $\hat{I^e}$ must exhibit relative to $I^o$ to be considered at least a partial success edit. 

First, $\hat{I^e}$ should be semantically closer to $I^e$ than $I^o$, which can be measured by CLIP directional similarity (CLIP-D)~\cite{brooks2023instructpix2pix}. The CLIP-D score for an arbitrary edited output $\tilde{I^e}$ is defined as: 
\begin{equation}
\begin{split}
    \text{CLIP-D}(I^o, \tilde{I^e}, p^o, p^e) = \text{cos}(&\text{CLIP}_\text{vis}(\tilde{I^e}) - \text{CLIP}_\text{vis}(I^o),\\
    &\text{CLIP}_\text{text}(p^e) - \text{CLIP}_\text{text}(p^o)),
\end{split}
\end{equation}

where $\text{CLIP}_\text{vis}$ and $\text{CLIP}_\text{text}$ are the text and visual encoder of the CLIP model~\cite{radford2021learning}. If $\tilde{I^e} = I^o$, the CLIP-D score is zero. If $\hat{I^e}$ has a negative CLIP-D score, it is further from the ground-truth edit than the input image itself - a known failure case - so it should receive an evaluation score of 0.

Secondly, $\hat{I^e}$ should be visually closer to $I^e$ than $I^o$, measured by CLIP similarity (CLIP-I)~\cite{radford2021learning} and DINO similarity (DINO-I)~\cite{caron2021emerging}. The CLIP-I and DINO-I score of an arbitrary edited output $\tilde{I^e}$ are 
\begin{align}
    \text{CLIP-I}(\tilde{I^e}, I^e) &= \text{cos}(\text{CLIP}_\text{vis}(\tilde{I^e}), \text{CLIP}_\text{vis}(I^e)) \\
    \text{DINO-I}(\tilde{I^e}, I^e) &= \text{cos}(\text{DINO}_\text{vis}(\tilde{I^e}), \text{DINO}_\text{vis}(I^e)),
\end{align}
where $\text{DINO}_\text{vis}$ is the DINOv2 model~\cite{oquab2023dinov2}. However, assigning $\hat{I^e}$ an evaluation score 0 solely based on $\text{CLIP-I}(\hat{I^e}, I^e) \leq \text{CLIP-I}(I^o, I^e)$ and/or $\text{DINO-I}(\hat{I^e}, I^e) \leq \text{DINO-I}(I^o, I^e)$ is problematic because CLIP and DINO models prioritize high-level semantics, so they often overlook fine details~\cite{bianchi2024clip} and lack perceptual quality awareness~\cite{fu2023dreamsim}. Instead, we compute the CLIP-I and DINO-I scores for all input images in the dataset and take their $5^{th}$ percentile as thresholds, denoted as $\tau_\text{CLIP-I}$ and $\tau_\text{DINO-I}$. If $\text{CLIP-I}(\hat{I^e}) \leq \tau_\text{CLIP-I}$ and $\text{DINO-I}(\hat{I^e}) \leq \tau_\text{DINO-I}$, we are more confident that $\hat{I^e}$ is indeed more visually different from $I^e$, thus is assigned score 0.

Crucially, our approach does not directly use CLIP/DINO metrics as evaluation scores but instead leverages their fundamental characteristics and overall trends in the dataset to identify easy negative samples, avoiding reliance on these models' limitations in capturing subtle image differences.

We focus on four models that consistently produce lower-quality edits: DiffEdit~\cite{couairon2022diffedit}, Pix2Pix-Zero (P2P-0)~\cite{parmar2023zero}, SDEdit~\cite{meng2021sdedit}, and Text2LIVE (T2L)~\cite{bar2022text2live}. As shown in Fig.~\ref{fig:data_creation_synthetic}, DiffEdit uses mask guidance to localize edits but often introduces incoherent pixels. P2P-0 ideally requires isolating input-target differences to define edit directions, but in practice, full captions are used, leading to over-editing. SDEdit denoises from added noise using the target prompt, frequently losing visual details. T2L relies on region-specific prompts, and when full captions are used instead, the outputs are often sub-optimal. While a small subset of samples may show better quality, the overall inconsistency justifies assigning all outputs from these models a score of 0.

\noindent\textbf{Positive samples.} We identify two models with overall better editing performance: MagicBrush~\cite{zhang2024magicbrush} and AURORA~\cite{krojer2024aurora}, as the former is trained on this dataset and the latter fine-tuned from the former. To locate samples with a score of 0.5 (partial edit success), we apply a CLIP-D threshold of $\tau_\text{CLIP-D} = 0.2$, which is used in IP2P~\cite{brooks2023instructpix2pix} to keep successful edits, and assign this score to samples below the threshold. Notably, this approach generalizes beyond our selected datasets and models, as it is applicable to any editing dataset and models trained on them, such as SEED-Data-Edit~\cite{ge2024seeddataedit} and SEED-X-Edit~\cite{ge2024seedx}.

\subsection{Data creation from multi-turn edit sequences\label{sec:data_creation_multiturn}}

\begin{figure}
    \centering
    \includegraphics[width=\linewidth]{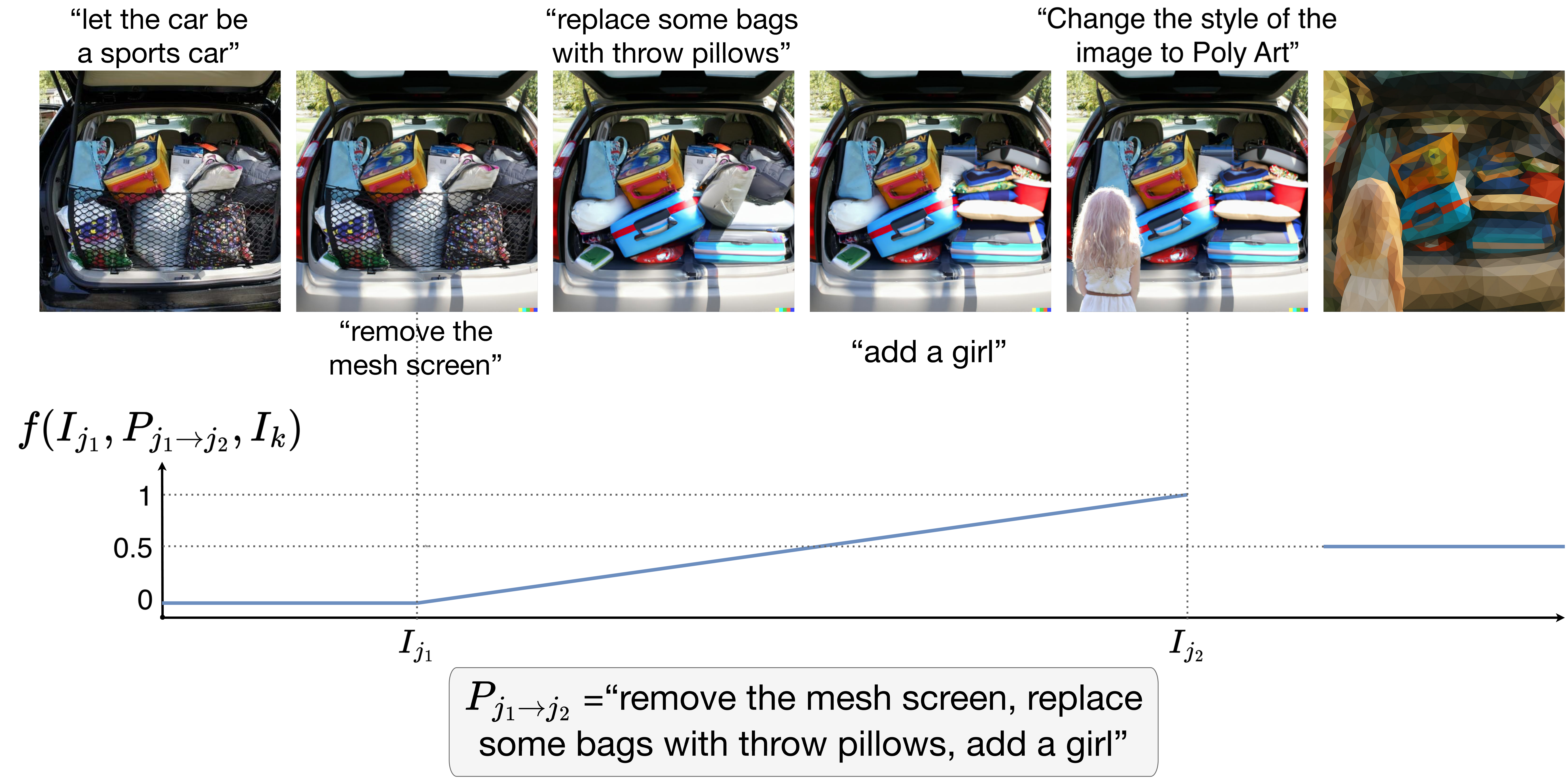}
    \vspace{-0.15in}
    \caption{For each sequence of edited images in a multi-turn image editing dataset~\cite{ge2024seeddataedit}, we can randomly sample two images as the input image and the ground truth output respectively ($I_{j_1}, I_{j_2}$) with respect to the set of edit instructions between them ($P_{j_1\rightarrow j_2}$). Then, we can evaluate the edit quality of any image $I_k$ in the same sequence and assign them score $f(I_{j_1}, P_{j_1\rightarrow j_2}, I_k)$ (y-axis).}
    \label{fig:data_creation_multiturn}
\end{figure}

To further diversify our training set in terms of edit instructions and score distribution, we leverage multi-turn edit sequences~\cite{zhang2024magicbrush,ge2024seeddataedit}, where successive edits are applied to the same image. A key insight is that any image in the sequence can serve as the ground-truth output for any preceding image, given the corresponding edit instructions. This enables a systematic construction of image editing evaluation samples as shown in Fig.~\ref{fig:data_creation_multiturn}.

Formally, define an edit sequence as $\mathcal{S} = [I_0, p_1, I_1, \dots, p_l, I_l]$ with $l$ edit turns, where the $j^{\text{th}}$ edit turn ($j \in [1, l]$) corresponds to performing the edit instruction $p_j$ on the input image $I_{j-1}$ to produce the output image $I_j$. To generate evaluation samples, we randomly sample two images $I_{j_1}, I_{j_2} \in \mathcal{S}$ where $j_1 < j_2$. We treat them as the input image and the ground-truth output image, with the corresponding edit instruction defined as $P_{j_1\rightarrow j_2} = \{p_j \mid j \in [j_1, j_2-1]\}$. The editing quality of any image $I_k \in \mathcal{S}$ with respect to the input image $I_{j_1}$ and the edit instruction $P_{j_1\rightarrow j_2}$ can be categorized as below. 

When $k \in [1, j_1]$, the selected image is either the input image itself or an image from an earlier edit turn. None of the changes from $I_{j_1}$ to $I_k$ align with the edit instruction $P_{j_1\rightarrow j_2}$, indicating a failure to apply the edit, thus should have score 0. When $k \in [j_1+1, j_2-1]$, the selected image corresponds to an intermediate edit turn between $I_{j_1}$ and $I_{j_2}$. Edit instructions $\{p_k \mid k \in [j_1, k-1]\}$ have been successfully performed, while instructions $\{p_j \mid j \in [k, j_2-1]\}$ remain incomplete. If we assume the level of success is directly proportional to the number of instructions that have been successfully performed, then the score should be $\frac{k - j_1}{j_2 - j_1}$. If $k = j_2$, the ground-truth output image is selected, which corresponds to a fully successful edit with score 1. Lastly, when $k \in [j_2+1, l]$, all edit instructions in $P_{j_1\rightarrow j_2}$ have been successfully applied, but additional irrelevant instructions $\{p_j \mid j \in [j_2+1, k]\}$ have also been performed. Since it is difficult to automatically evaluate the level of negative effect of excessive edit on edit quality, we assume that the over-edit outputs get score of 0.5. 

Collectively, the score assignment function for an output image $I_k$ with respect to the input $I_{j_1}$ and edit instruction $P_{j_1\rightarrow j_2}$ can be defined as:
\begin{equation}
f(I_{j_1}, P_{j_1\rightarrow j_2}, I_k) =
\begin{cases} 
0 & \text{if } k \in [1, j_1), \\
\frac{k - j_1}{j_2 - j_1} & \text{if } k \in [j_1, j_2], \\
0.5 & \text{if } k \in (j_2, l].
\end{cases}
\label{eq:label_assignment}
\end{equation}

\subsection{Evaluation scorer architecture\label{sec:evaluation_scorer}}

\begin{figure}
    \centering
    \includegraphics[width=\linewidth]{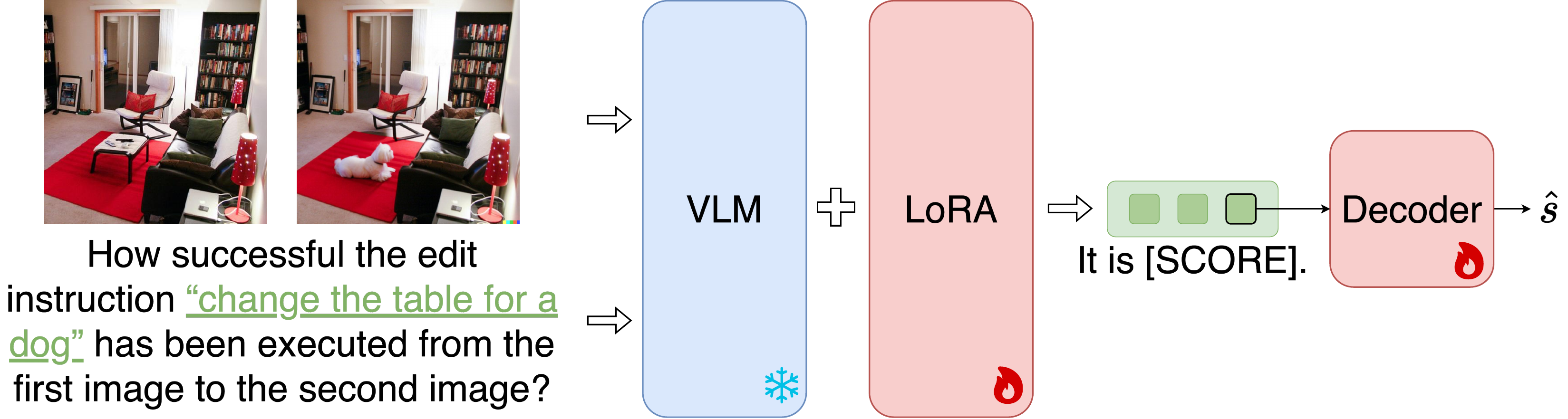}
    \vspace{-0.15in}
    \caption{To train a VLM~\cite{liu2023llava} as image editing evaluator, we concatenate the input and the output image and send them to the model along with the text query. The embedding corresponding to the special token \textsc{[SCORE]} is decoded into the final image editing quality evaluation score by a decoder consisting of several fully-connected layers. The weights of the VLM are fine-tuned using LoRA~\cite{hu2022lora}.}
    \label{fig:architecture}
    \vspace{-0.05in}
\end{figure}

After creating the training set using above approaches, we fine-tune LLaVA-Next-8B~\cite{liu2024llavanext} as an image editing evaluation scorer. We choose this VLM since it can handle multiple, high-resolution images as input. Specifically, we use a similar approach as LISA~\cite{lai2023lisa}, where we expand the vocabulary of the VLM with a special token \textsc{[SCORE]} whose embedding is decoded through a MLP-based decoder into the final evaluation score as shown in Fig.~\ref{fig:architecture}. This simple yet effective approach yields evaluation scores closely resemble human preferences in our empirical study. 

\subsection{Image editing model fine-tuning\label{sec:reward_conditioned_image_editing}}

\noindent\textbf{Reward-condition fine-tuning}  Beyond evaluating various image editing methods, the trained image editing scorer can act as a reward function and be used to fine-tune and improve image editing models such as MagicBrush~\cite{zhang2024magicbrush}. To condition image editing on the evaluation score, we first estimate the editing quality score $\hat{s}$ for each training sample, and append an additional text prompt ``The image quality is $\text{[SCORE]}$ out of five'' after the corresponding edit instruction, similar to HIVE~\cite{zhang2024hive}. Here the editing quality score is normalized to score 1-5. In the end, we fine-tune MagicBrush model on its own training data but with the new text inputs consisting of both the edit instructions and the edit quality scores. Once the model is calibrated with the image quality level, we prompt the model to generate images at the highest quality (\ie, 5) during inference.

\noindent\textbf{Reward feedback learning fine-tuning} We also experiment with reward feedback learning~\cite{xu2024imagereward}, where our scorer is used as part of the loss function to fine-tune MagicBrush~\cite{zhang2024magicbrush}. Specifically, for each training step, we randomly set the total number of diffusion denoising steps to $T \in \{5k | k \in [2, 20]\}$, where we denoise a random gaussian noise to an edited output for T-1 steps. Then, for the last diffusion step, we generate the pristine edited output and its evaluation score $s$ (normalized to 0-10) using the proposed scorer. The reward feedback learning loss is formulated as:
\begin{equation}
    \mathcal{L}_{reward} = 10 - s,
\end{equation}
which calculates the gap between the evaluation score and the maximum score 10. We use this with the diffusion model MSE loss as detailed in the supplementary material.
\section{Experiments}

\subsection{Dataset generation and implementation\label{sec:dataset_generation_implementation}}

We use 9,935 samples from MagicBrush~\cite{zhang2024magicbrush} and 1,100 additional instructions for global/style edits on Emu-Edit test images~\cite{sheynin2024emu}, generating 93,915 outputs via nine text-guided editing methods. For prompt generation, we fine-tune Qwen2-VL-7B-Instruct~\cite{wang2024qwen2} on 600 manually curated samples (see Sec.~\ref{sec:data_creation_synthetic}) using LLaMA-Factory~\cite{zheng2024llamafactory}. We also incorporate 21,382 multi-turn sequences from SEED-Data-Edit-Part3~\cite{ge2024seeddataedit}.

We train LLaVA-Next-8B~\cite{liu2024llavanext} for 30 epochs on 6×H100 GPUs (batch size 192, LoRA rank 8~\cite{hu2022lora}, learning rate 2e-5). The trained scorer is then used to predict quality scores for MagicBrush training data, which we use to fine-tune MagicBrush~\cite{zhang2024hive} for 10,000 steps (batch size 384, learning rate 5e-5). Further details are in the supplementary.

\subsection{Benchmarks and Baselines\label{sec:benchmarks_baselines}}

We evaluate our approach on ImagenHub~\cite{ku2024imagenhub}, GenAI-Bench~\cite{jiang2024genai}, and AURORA-Bench~\cite{krojer2024aurora}. ImagenHub and AURORA-Bench (point-wise) contain human-rated edits, with Spearman correlation used to assess alignment with human scores. ImagenHub includes three human ratings per sample, enabling Human-to-Human correlation computation as an upper bound~\cite{ku2023viescore}.

GenAI-Bench and AURORA-Bench (pair-wise) feature paired outputs with preference labels. Evaluation is based on prediction accuracy: if our model assigns a higher score to the preferred image, it is counted as correct; otherwise, a tie is predicted.


We also report CLIP score (CLIP-T)~\cite{hessel2021clipscore}, directional similarity (CLIP-D)~\cite{brooks2023instructpix2pix}, CLIP similarity (CLIP-I)~\cite{radford2021learning}, and DINO similarity (DINO-I)~\cite{caron2021emerging} for datasets with ground-truth outputs. See supplementary for further details.

\subsection{Image editing evaluation}

\begin{figure*}[]
    \centering
    \small
    \setlength{\tabcolsep}{2pt}
    \resizebox{\linewidth}{!}{
    \begin{tabular}{cc|cc|cc|cc}
    \multicolumn{1}{c}{Input} &  Output & \multicolumn{1}{c}{Input} &  Output & \multicolumn{1}{c}{Input} &  Output & \multicolumn{1}{c}{Input} &  Output \\ 
    
    \includegraphics[width=0.12\textwidth]{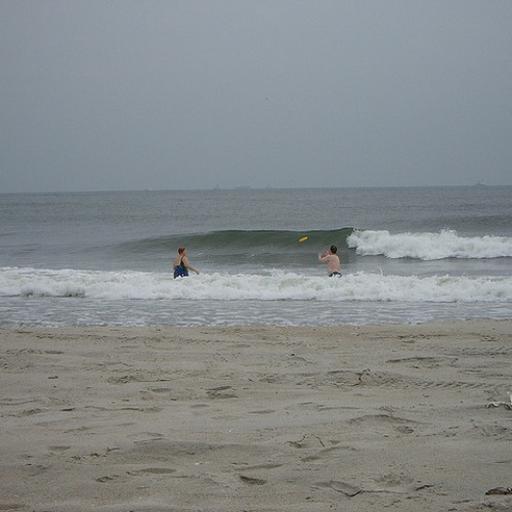} &  
    \includegraphics[width=0.12\textwidth]{fig/4_experiments/image_editing_evaluation/1_input.jpg} & 
    \includegraphics[width=0.12\textwidth]{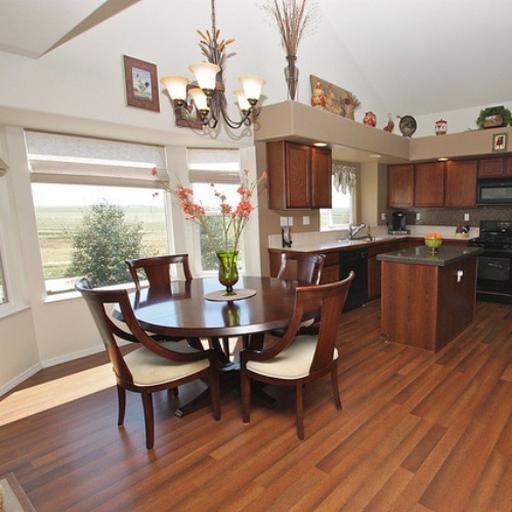} & 
    \includegraphics[width=0.12\textwidth]{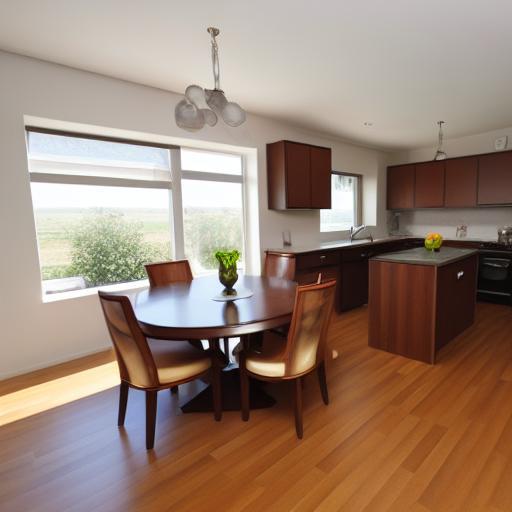} & 
    \includegraphics[width=0.12\textwidth]{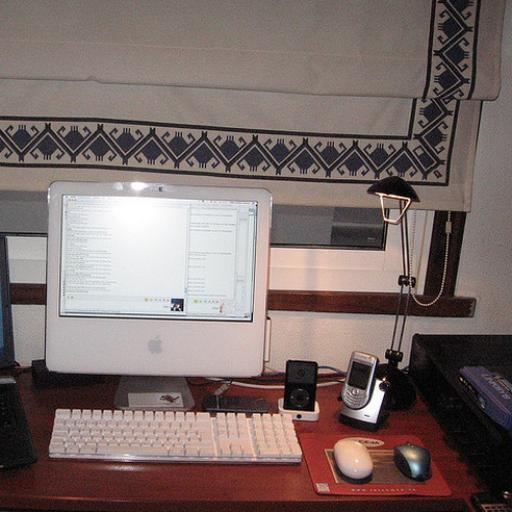} & 
    \includegraphics[width=0.12\textwidth]{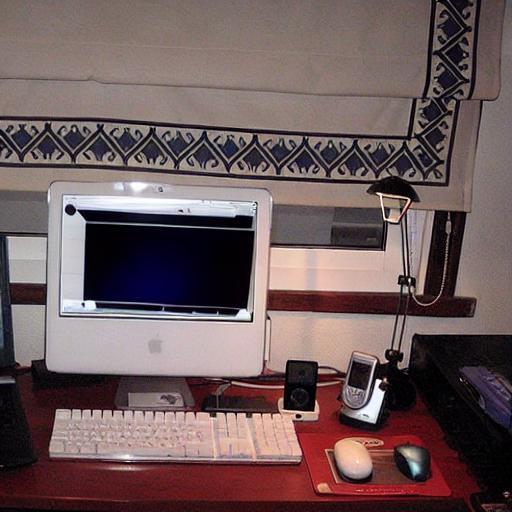} & 
    \begin{tikzpicture}
        \node[anchor=south west, inner sep=0] (image) at (0,0) {\includegraphics[width=0.12\linewidth]{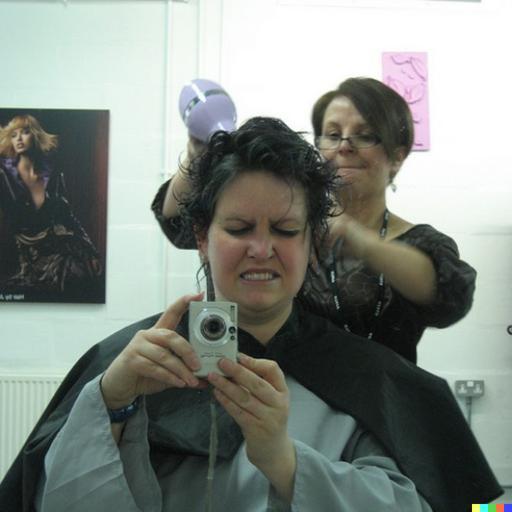}};

        \draw[fill=black, opacity=1.0] (1.06,1.1) circle [radius=0.17]; 
        \draw[fill=black, opacity=1.0] (1.45,1.5) circle [radius=0.15]; 
    \end{tikzpicture} & 
    \begin{tikzpicture}
        \node[anchor=south west, inner sep=0] (image) at (0,0) {\includegraphics[width=0.12\linewidth]{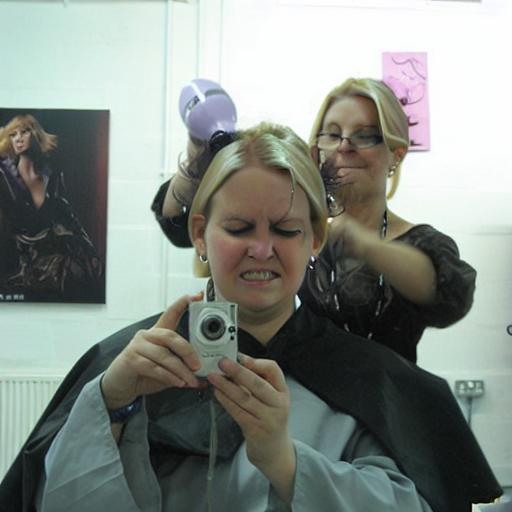}};

        \draw[fill=black, opacity=1.0] (1.06,1.1) circle [radius=0.17]; 
        \draw[fill=black, opacity=1.0] (1.45,1.5) circle [radius=0.15]; 
    \end{tikzpicture} \\
    
    \multicolumn{2}{l|}{\makecell[l]{
    ``change the frisbee into a ball''
    }} & 
    \multicolumn{2}{l|}{\makecell[l]{
    ``get rid of the vase on top of\\ the table''
    }} & 
    \multicolumn{2}{l|}{\makecell[l]{
    ``Let the monitor turn black.''
    }} & 
    \multicolumn{2}{l}{\makecell[l]{
    ``let the woman have blonde hair''
    }} \\

    \makecell[l]{\textbf{GT}: 1.67} & \makecell[l]{\textbf{Ours}: 1.48} & 
    \makecell[l]{\textbf{GT}: 0.0} & \makecell[l]{\textbf{Ours}: 0.85} & 
    \makecell[l]{\textbf{GT}: 3.33} & \makecell[l]{\textbf{Ours}: 5.70} &  
    \makecell[l]{\textbf{GT}: 10.0} & \makecell[l]{\textbf{Ours}: 8.28} \\

    \makecell[l]{\textbf{GPT-4o}: 1.41} & \makecell[l]{\textbf{Gemini}: 5.48} &
    \makecell[l]{\textbf{GPT-4o}: 3.0} & \makecell[l]{\textbf{Gemini}: 7.75} &
    \makecell[l]{\textbf{GPT-4o}: 6.71} & \makecell[l]{\textbf{Gemini}: 7.07} &
    \makecell[l]{\textbf{GPT-4o}: 6.32} & \makecell[l]{\textbf{Gemini}: 6.0} \\

    \midrule
    \includegraphics[width=0.12\textwidth]{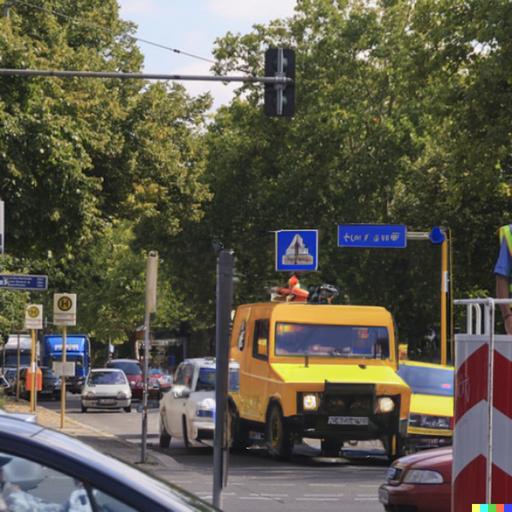} &  
    \includegraphics[width=0.12\textwidth]{fig/4_experiments/image_editing_evaluation/94_input.jpg} & 
    \includegraphics[width=0.12\textwidth]{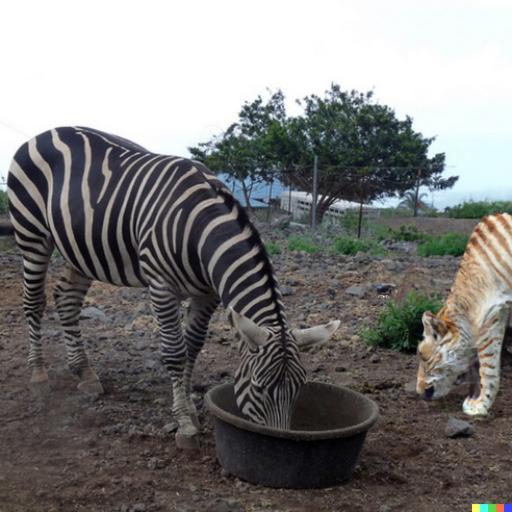} & 
    \includegraphics[width=0.12\textwidth]{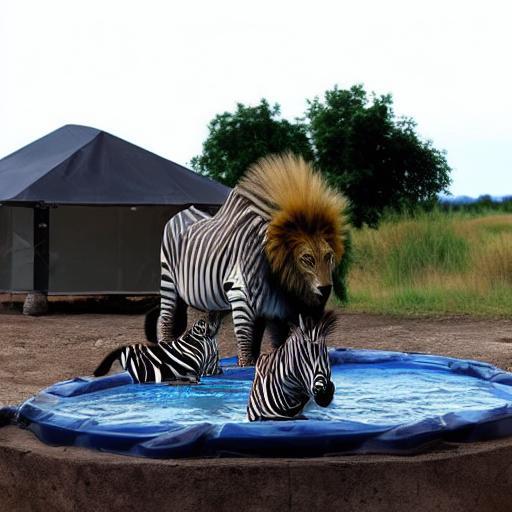} & 
    \includegraphics[width=0.12\textwidth]{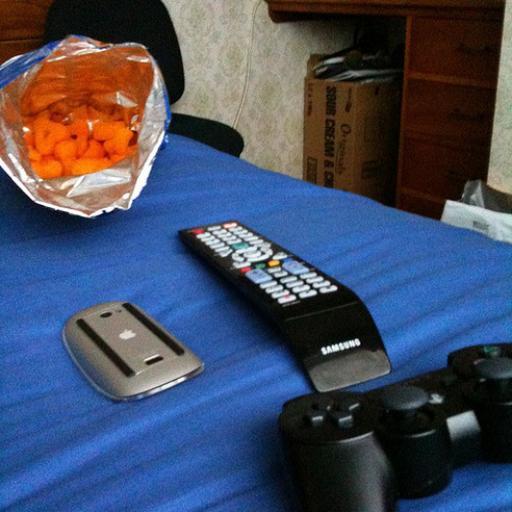} & 
    \includegraphics[width=0.12\textwidth]{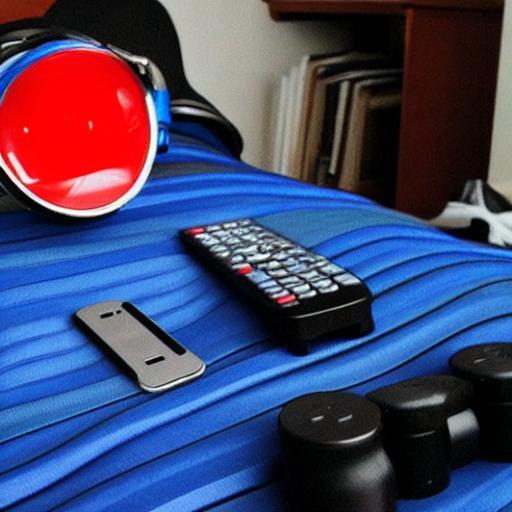} & 
    \includegraphics[width=0.12\textwidth]{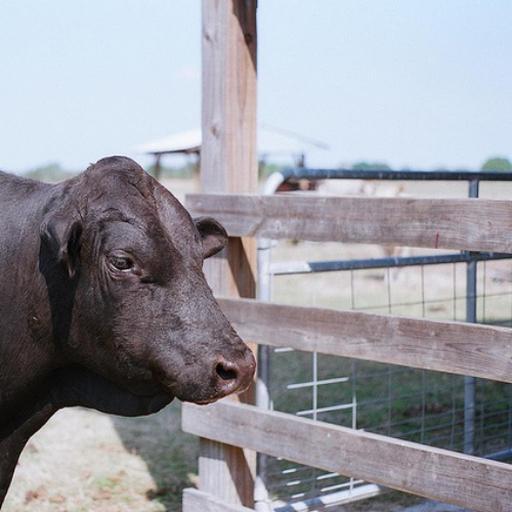} & 
    \includegraphics[width=0.12\textwidth]{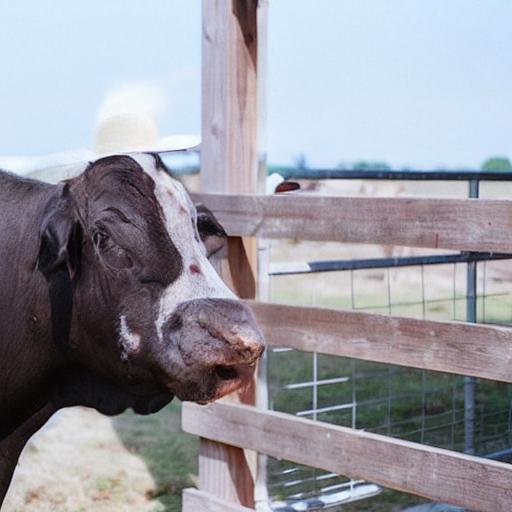} \\
    
    \multicolumn{2}{l|}{\makecell[l]{
    ``add a pedestrian''
    }} & 
    \multicolumn{2}{l|}{\makecell[l]{
    ``turn the basin into a plastic pool''
    }} & 
    \multicolumn{2}{l|}{\makecell[l]{
    ``change the bag of chips into\\ a backpack''
    }} & 
    \multicolumn{2}{l}{\makecell[l]{
    ``Have the cow wear a hat.''
    }} \\

    \makecell[l]{\textbf{GT}: 0.0} & \makecell[l]{\textbf{Ours}: 1.94} & 
    \makecell[l]{\textbf{GT}: 0.0} & \makecell[l]{\textbf{Ours}: 2.91} & 
    \makecell[l]{\textbf{GT}: 1.66} & \makecell[l]{\textbf{Ours}: 0.56} &  
    \makecell[l]{\textbf{GT}: 3.33} & \makecell[l]{\textbf{Ours}: 3.43} \\

    \makecell[l]{\textbf{GPT-4o}: 2.65} & \makecell[l]{\textbf{Gemini}: 6.0} &
    \makecell[l]{\textbf{GPT-4o}: 3.46} & \makecell[l]{\textbf{Gemini}: 4.0} &
    \makecell[l]{\textbf{GPT-4o}: 3.0} & \makecell[l]{\textbf{Gemini}: 0.0} &
    \makecell[l]{\textbf{GPT-4o}: 4.90} & \makecell[l]{\textbf{Gemini}: 0.0} \\
    \end{tabular}}
    \vspace{-0.1in}
    \caption{Evaluation examples (faces are blocked for privacy concern) from GPT-4o~\cite{achiam2023gpt}, Gemini-Pro 1.5 (Gemini)~\cite{team2024gemini}, and our method on ImagenHub~\cite{ku2024imagenhub}, with the ground-truth (GT) scores. See supplementary for more results.}
    \label{fig:image_editing_evaluation_results_qualitative}
\end{figure*}

We present the quantitative comparison in Table~\ref{tab:image_editing_evaluation_correlation} and Table~\ref{tab:image_editing_evaluation_accuracy} along with qualitative examples in Fig.~\ref{fig:image_editing_evaluation_results_qualitative}. On GenAI-Bench~\cite{jiang2024genai} and AURORA-Bench~\cite{krojer2024aurora}, the proposed \ourwork{} scorer out-performs all baselines. Compared to the state-of-the-art, it achieve a 0.0696 (+17.24\%) increase in correlation on AURORA-Bench (point-wise) while improving pair-wise comparison accuracy by 4.03\% (+7.21\%) on GenAI-Bench and 4.75\% (+9.35\%) on AURORA-Bench (pair-wise). For ImagenHub~\cite{ku2024imagenhub}, our scorer also surpasses all open-sourced VLMs and Gemini-Pro 1.5~\cite{team2024gemini}, while closely trailing GPT-4o. 

CLIP-D, CLIP-I, DINO-I performs better than all open-sourced VLMs , which supports our design choices of utilizing these metrics for dataset creation. For more results, please refer to the supplementary material.

\begin{table}[]
\renewcommand{\arraystretch}{0.9} 
\centering
\small
\resizebox{\linewidth}{!}{
\begin{tabular}{lcc}
\toprule
                        & ImagenHub & AURORA-Bench (point-wise) \\
\midrule
Human-to-Human          & 0.4184  &  -       \\
\midrule
CLIP-D                  & 0.2117  &  0.3080  \\ 
CLIP-T                  & 0.1894  &  0.1847  \\
CLIP-I                  & 0.1261  &  -       \\
DINO-I                  & 0.0441  &  -       \\ 
\midrule
GPT-4o                  & \textbf{0.3821}  &  \underline{0.4038}  \\
Gemini-Pro 1.5          & 0.2728  &  0.1052  \\
\midrule
LLaVA                   & 0.0273  &  0.0073  \\  
LLaVA-NeXT              & 0.0356  &  -0.0491 \\
LLaVA-OneVision         & 0.0829  & 0.0555   \\
Qwen-VL                 & 0.0404  &  0.0118  \\
Qwen2-VL                & 0.1445  & 0.1783   \\
Qwen2.5-VL              & 0.1859  & 0.2351   \\
Phi-3.5-vision-instruct & 0.1126  & -0.0107  \\
Pixtral                 & 0.0123  & -0.0005  \\
BLIP-2                  & 0.0378  & -0.0003  \\
InstructBLIP            &  0.0212 & -0.0351  \\
Fuyu                    & 0.0206  & -0.0044  \\
CogVLM                  & -0.0288 & 0.0199   \\
OpenFlamingo            & -0.0577 & 0.0065   \\
\textbf{\ourwork{} (Ours)}       & \underline{0.3450}  & \textbf{0.4734}   \\
\bottomrule
\end{tabular}}
\vspace{-0.1in}
\caption{
Correlations of predicted scores with human ratings. The best and second-to-best values are in bold and underlined respectively. 
Please refer to Sec.~\ref{sec:benchmarks_baselines} and the supplementary for more details and results.
}
\label{tab:image_editing_evaluation_correlation}
\end{table}

\begin{table}[]
\renewcommand{\arraystretch}{0.9} 
\centering
\small
\resizebox{\linewidth}{!}{
\begin{tabular}{lcc}
\toprule
                & GenAI-Bench & AURORA-Bench (pair-wise) \\
\midrule
random          & 25.90       &  33.43 \\
\midrule
CLIP-D          & 43.09       &  31.63  \\
CLIP-T          & 39.39       &  42.93  \\
CLIP-I          & 38.96       & -       \\  
DINO-I          & 36.78       & -       \\
\midrule
GPT-4o          & 53.54       & \underline{50.81}  \\
Gemini-Pro 1.5  & \underline{55.93}       & 28.13  \\
\midrule
LLaVA           & 26.12       & 27.50  \\
LLaVA-NeXT      & 25.35       & 27.19  \\
LLaVA-OneVision & 22.31       & 33.25  \\
Qwen-VL         & 14.91       & 12.69  \\
Qwen2-VL        & 26.12       & 27.38  \\
Qwen2.5-VL      & 32.10       & 30.69  \\
Phi-3.5-vision-instruct & 21.87 & 32.25\\
Pixtral         & 26.12       & 27.38  \\
BLIP-2          & 26.01       & 26.25  \\
InstructBLIP    & 19.80       & 16.69  \\
\textbf{\ourwork{} (Ours)} & \textbf{59.96}       & \textbf{55.56} \\
\bottomrule
\end{tabular}}
\vspace{-0.1in}
\caption{Accuracy of predicted comparison labels with human preference. The best and second-to-best values are in bold and underlined respectively. Please refer to Sec.~\ref{sec:benchmarks_baselines} and the supplementary for more details and results.
}
\label{tab:image_editing_evaluation_accuracy}
\end{table}

\begin{figure*}
    \centering
    \captionsetup[subfigure]{labelformat=empty}
    
    \begin{small}
        \begin{tabbing}
            \hspace{5.0em} \= \hspace{4.7em} \= \hspace{6.0em} \= \hspace{5.6em} \= \hspace{8.5em} \= \hspace{4.8em} \= \hspace{6.0em} \= \hspace{5.5em} \=  \kill
            \> Input \> Baseline \> Ours \> GT \> Input \> Baseline \> Ours \> GT
        \end{tabbing}    
    \end{small}
    \vspace{-0.15in}

    \begin{subfigure}[c]{0.4\textwidth}
        \begin{tikzpicture}[scale=0.816]
            \node[anchor=south west, inner sep=0] (image) at (0,0) {\includegraphics[width=\textwidth, trim=0 30px 0 30px, clip]{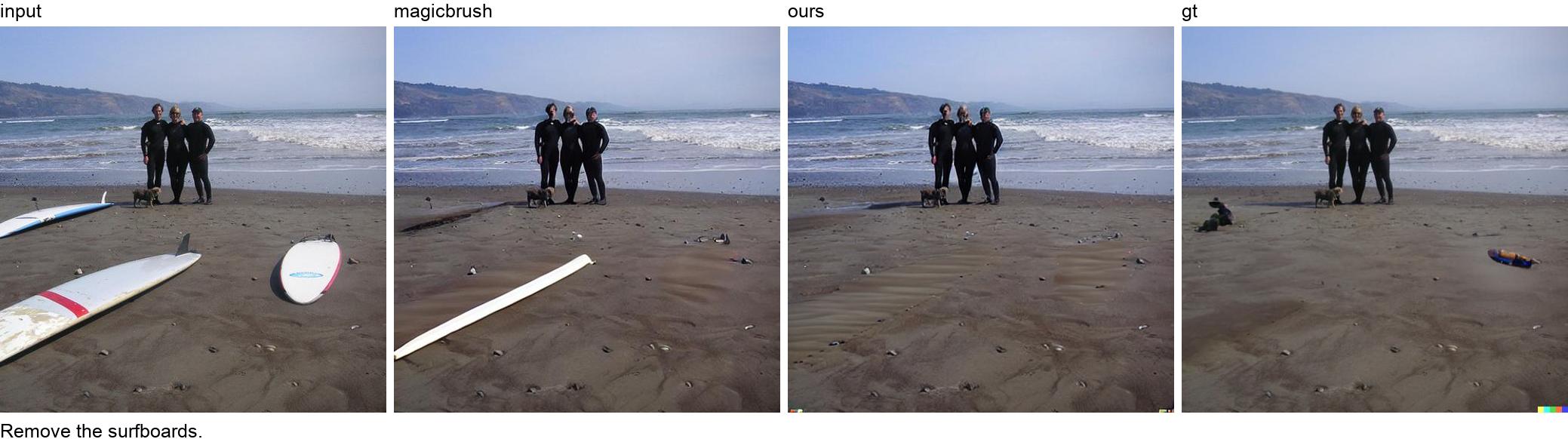}};
    
            \draw[fill=black, opacity=1.0] (0.86,1.63) rectangle (1.10,1.7);
            \draw[fill=black, opacity=1.0] (3.01,1.63) rectangle (3.25,1.7);
            \draw[fill=black, opacity=1.0] (5.16,1.63) rectangle (5.4,1.7);
            \draw[fill=black, opacity=1.0] (7.31,1.63) rectangle (7.55,1.7);
        \end{tikzpicture}
        \vspace{-0.25in}
        \caption{``Remove the surfboards''}
    \end{subfigure}
    \hspace{0.4in}
    \begin{subfigure}[c]{0.4\textwidth}
        \begin{tikzpicture}[scale=0.816]
            \node[anchor=south west, inner sep=0] (image) at (0,0) {\includegraphics[width=\textwidth, trim=0 30px 0 30px, clip]{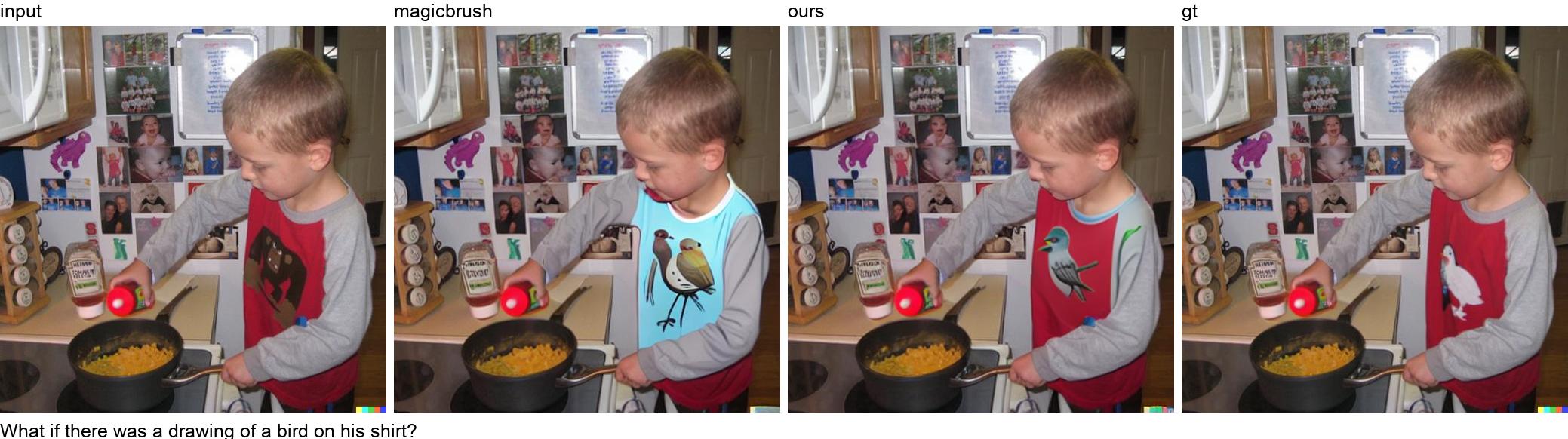}};
    
            \draw[fill=black, opacity=1.0] (1.4,1.3) circle [radius=0.15];
            \draw[fill=black, opacity=1.0] (0.66,1.35) circle [radius=0.32];
            \draw[fill=black, opacity=1.0] (3.55,1.3) circle [radius=0.15];
            \draw[fill=black, opacity=1.0] (2.81,1.35) circle [radius=0.32];
            \draw[fill=black, opacity=1.0] (5.7,1.3) circle [radius=0.15];
            \draw[fill=black, opacity=1.0] (4.96,1.35) circle [radius=0.32];
            \draw[fill=black, opacity=1.0] (7.85,1.3) circle [radius=0.15];
            \draw[fill=black, opacity=1.0] (7.11,1.35) circle [radius=0.32];
        \end{tikzpicture}
        \vspace{-0.25in}
        \caption{``What if there is a drawing of a bird on his shirt''}
    \end{subfigure}

    \begin{subfigure}[c]{0.4\textwidth}
        \includegraphics[width=\textwidth, trim=0 30px 0 30px, clip]{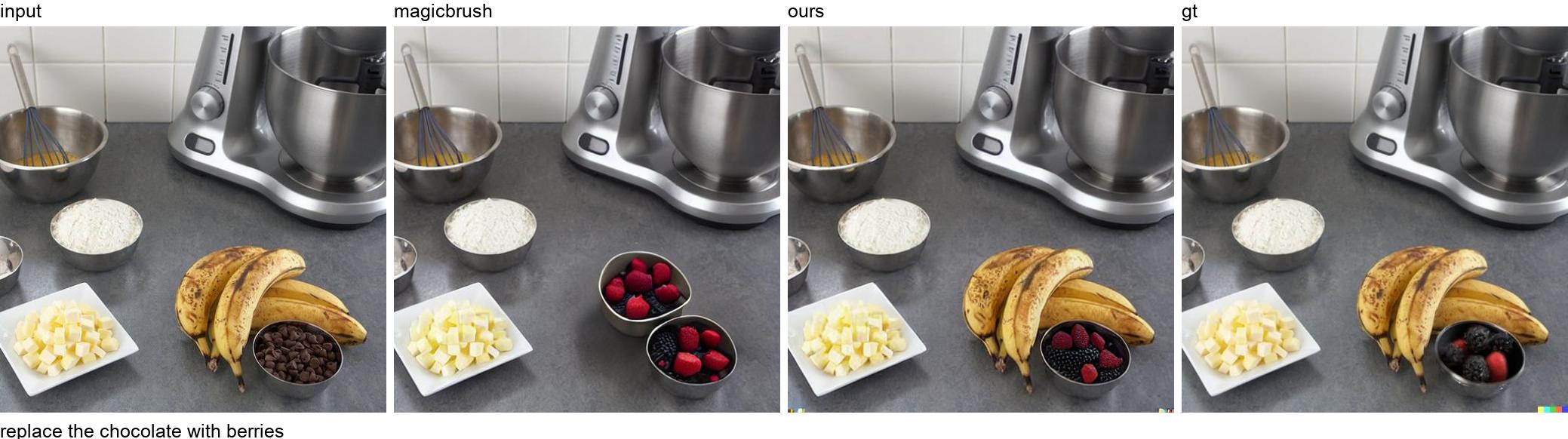}
        \vspace{-0.25in}
        \caption{``Replace the chocolate with berries''}
    \end{subfigure}
    \hspace{0.4in}
    \begin{subfigure}[c]{0.4\textwidth}
        \includegraphics[width=\textwidth, trim=0 30px 0 30px, clip]{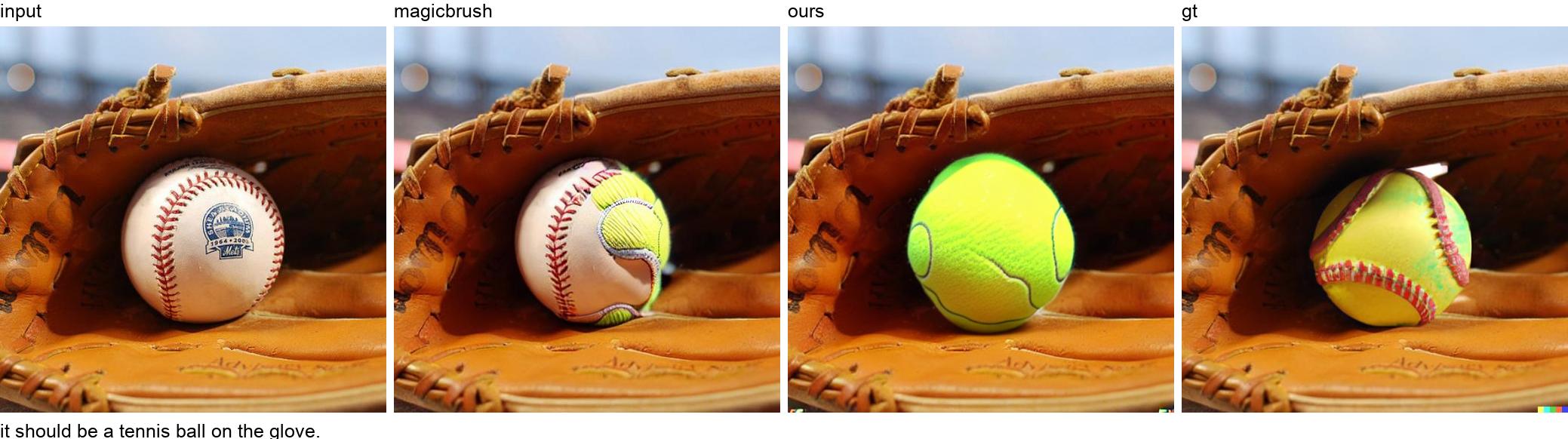}
        \vspace{-0.25in}
        \caption{``It should be a tennis ball on the glove''}
    \end{subfigure}
    \vspace{-0.1in}
    \caption{Image editing examples (face are blocked for privacy concern) from the MagicBrush editing model~\cite{zhang2024magicbrush} and our fine-tuned model using our image editing scorer as a reward model. See supplementary for more results.}
    \label{fig:image_editing_reward}
\end{figure*}
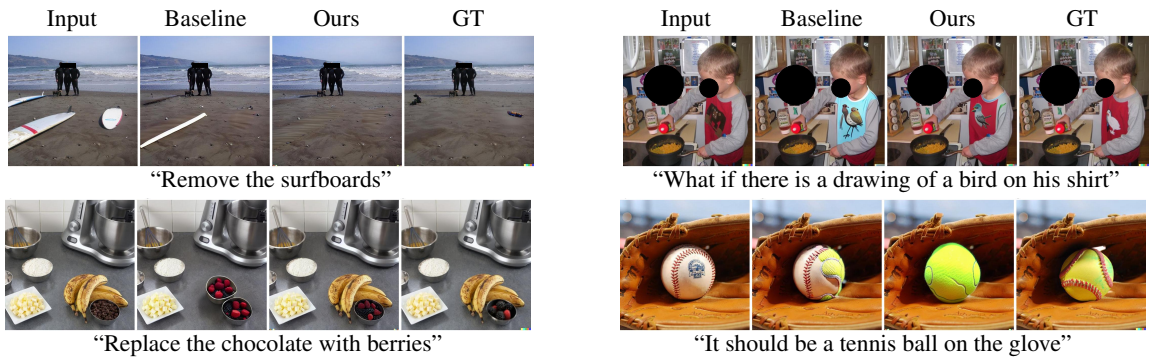

\subsection{Image Editing}

According to our evaluation scorer, the MagicBrush baseline~\cite{zhang2024magicbrush} achieves an average score of 5.90 on ImagenHub samples~\cite{ku2024imagenhub}, while our model fine-tuned with reward conditioning scores 6.43 (+8.98\%) and the model trained with reward feedback learning scores 6.30 (+6.78\%). Table~\ref{tab:image_editing_quantitative} further compares these models using standard automatic metrics, which, as shown in Tables 1–2, often poorly correlate with human judgment.

To address this gap, we conducted a user study (Table~\ref{tab:user_study}) involving 56 participants, each comparing 50 randomly sampled output pairs generated from the same input and instruction. Participants selected the better result or marked them as similar. Our method was preferred over the baseline in 41.67\% of cases vs. 32.00\%, demonstrating improved edit quality despite identical inputs.

Fig.~\ref{fig:image_editing_reward} presents visual comparisons, where our reward-conditioned model produces more faithful edits. Additional examples are included in the supplementary material.

\begin{table}[]
\renewcommand{\arraystretch}{0.9}
\centering
\small
\begin{tabular}{lcc}
    \toprule
             & Baseline & Ours \\
    \midrule
    CLIP-T   & 30.890   & 31.167 (+0.90\%) \\
    DINO-I   & 90.643   & 89.935 (-0.78\%) \\
    CLIP-I   & 95.231   & 95.215 (-0.02\%) \\
    CLIP-D   & 0.13604  & 0.13279 (-2.39\%) \\
    GPT-4o   & 4.02     & \textbf{4.36 (+8.46\%)} \\
    \bottomrule
\end{tabular}
\vspace{-0.1in}
\caption{Automatic evaluation metrics comparing baseline and our reward-conditioned image editing models.}
\label{tab:image_editing_quantitative}
\end{table}

\begin{table}[]
\renewcommand{\arraystretch}{0.9}
    \centering
    \small
    \begin{tabular}{lcc}
    \toprule
    Human Pref & Count & \% \\
    \midrule
    Baseline  & 914   & 32.00 \\
    Tied      & 752   & 26.33 \\
    Ours      & 1190  & \textbf{41.67} \\
    \bottomrule
    \end{tabular}
    \vspace{-0.1in}
    \caption{Human pairwise preference distribution comparing baseline and our reward-conditioned image editing models.}
    \label{tab:user_study}
\end{table}

\subsection{Alation Study}

\begin{table}[]
\renewcommand{\arraystretch}{0.9} 
\centering
\small
\resizebox{\linewidth}{!}{
\begin{tabular}{lcc}
\toprule
 & Ours & w/o score token \& MLP \\
\midrule
ImagenHub                  & 0.3450 & 0.2931 \\
AURORA-Bench (point-wise)             & 0.4744 & 0.3558 \\
GenAI-Bench                & 59.41  & 58.32  \\
AURORA-Bench (pair-wise)              & 52.38  & 20.50  \\
\bottomrule
\end{tabular}
}
\vspace{-0.1in}
\caption{Performance comparison between our method and the variant without score token and MLP decoder.}
\label{tab:ablation_vlm}
\end{table}

\noindent\textbf{Effect of score token and MLP decoder.} we fine-tuned the same VLM backbone as our method without the score token and MLP. As shown in Table~\ref{tab:ablation_vlm}, we see its consistent performance drop across benchmarks, which highlights the importance of the score token for isolating task-relevant information and the MLP for producing more accurate outputs and mitigating hallucinations.

\noindent\textbf{Effect of reward conditioning.} We fine-tuned the MagicBrush baseline on our dataset without reward conditioning, resulting in an score of 4.27 (vs.~6.67 from our method). This is due to the inclusion of low-score samples without reward-conditioning prompts to distinguish successful from failed edits. When training only on high-scoring samples (score $\geq$ 8 in a scale 0-10) with reward-conditioned prompts, the model reaches 5.09. Even with high-score samples, reduced diversity in the reward text and causes it to conflict with semantic meaning of the edit instruction, limiting the model’s learning capacity.

\section{Conclusion}

We introduce \ourwork{}, an automated dataset creation approach and evaluation scorer for instruction-guided image editing. We generate over 100K samples to fine-tune a VLM that surpasses all open-source VLMs and Gemini-Pro 1.5 across benchmarks. Our scorer improves score correlation with human ratings on AURORA-Bench by 0.0696 (+17.24\%) and enhances pair-wise comparison accuracy by 4.03\% (+7.21\%) on GenAI-Bench and 4.75\% (+9.35\%) on AURORA-Bench. It also improves image editing models, increasing their average score from 5.90 to 6.43 (+8.98\%).

For future work, we aim to enhance our scorer with explainability and reasoning by identifying edit artifacts and breaking down scores along axes like semantic alignment and perceptual quality~\cite{ku2023viescore}. We also plan to explore advanced fine-tuning methods such as real-time back-propagation through editing models using the inverse evaluation score as a loss signal.

{\small
\bibliographystyle{ieee_fullname}
\bibliography{egbib}

\begin{thebibliography}{10}\itemsep=-1pt

\bibitem{achiam2023gpt}
Josh Achiam, Steven Adler, Sandhini Agarwal, Lama Ahmad, Ilge Akkaya, Florencia~Leoni Aleman, Diogo Almeida, Janko Altenschmidt, Sam Altman, Shyamal Anadkat, et~al.
\newblock Gpt-4 technical report.
\newblock {\em arXiv preprint arXiv:2303.08774}, 2023.

\bibitem{cosxledit}
Stability AI.
\newblock Cos stable diffusion xl 1.0 and cos stable diffusion xl 1.0 edit, 2024.
\newblock \url{https://huggingface.co/stabilityai/cosxl}.

\bibitem{anthropic2024claude3.5sonnet}
Anthropic.
\newblock Claude 3.5 sonnet model card addendum.
\newblock Technical report, June 2024.

\bibitem{awadalla2023openflamingo}
Anas Awadalla, Irena Gao, Josh Gardner, Jack Hessel, Yusuf Hanafy, Wanrong Zhu, Kalyani Marathe, Yonatan Bitton, Samir Gadre, Shiori Sagawa, Jenia Jitsev, Simon Kornblith, Pang~Wei Koh, Gabriel Ilharco, Mitchell Wortsman, and Ludwig Schmidt.
\newblock Openflamingo: An open-source framework for training large autoregressive vision-language models.
\newblock {\em arXiv preprint arXiv:2308.01390}, 2023.

\bibitem{Qwen-VL}
Jinze Bai, Shuai Bai, Shusheng Yang, Shijie Wang, Sinan Tan, Peng Wang, Junyang Lin, Chang Zhou, and Jingren Zhou.
\newblock Qwen-vl: A versatile vision-language model for understanding, localization, text reading, and beyond.
\newblock {\em arXiv preprint arXiv:2308.12966}, 2023.

\bibitem{bar2022text2live}
Omer Bar-Tal, Dolev Ofri-Amar, Rafail Fridman, Yoni Kasten, and Tali Dekel.
\newblock Text2live: Text-driven layered image and video editing.
\newblock In {\em European Conference on Computer Vision}, pages 707--723. Springer, 2022.

\bibitem{fuyu-8b}
Rohan Bavishi, Erich Elsen, Curtis Hawthorne, Maxwell Nye, Augustus Odena, Arushi Somani, and Sa\u{g}nak Ta\c{s}\i{}rlar.
\newblock Introducing our multimodal models, 2023.

\bibitem{bianchi2024clip}
Lorenzo Bianchi, Fabio Carrara, Nicola Messina, and Fabrizio Falchi.
\newblock Is clip the main roadblock for fine-grained open-world perception?
\newblock In {\em 2024 International Conference on Content-Based Multimedia Indexing (CBMI)}, pages 1--8. IEEE, 2024.

\bibitem{brooks2023instructpix2pix}
Tim Brooks, Aleksander Holynski, and Alexei~A Efros.
\newblock Instructpix2pix: Learning to follow image editing instructions.
\newblock In {\em Proceedings of the IEEE/CVF Conference on Computer Vision and Pattern Recognition}, pages 18392--18402, 2023.

\bibitem{brown2020language}
Tom Brown, Benjamin Mann, Nick Ryder, Melanie Subbiah, Jared~D Kaplan, Prafulla Dhariwal, Arvind Neelakantan, Pranav Shyam, Girish Sastry, Amanda Askell, et~al.
\newblock Language models are few-shot learners.
\newblock {\em Advances in neural information processing systems}, 33:1877--1901, 2020.

\bibitem{caron2021emerging}
Mathilde Caron, Hugo Touvron, Ishan Misra, Herv{\'e} J{\'e}gou, Julien Mairal, Piotr Bojanowski, and Armand Joulin.
\newblock Emerging properties in self-supervised vision transformers.
\newblock In {\em Proceedings of the IEEE/CVF international conference on computer vision}, pages 9650--9660, 2021.

\bibitem{chen2024mega-bench}
Jiacheng Chen, Tianhao Liang, Sherman Siu, Zhengqing Wang, Kai Wang, Yubo Wang, Yuansheng Ni, Wang Zhu, Ziyan Jiang, Bohan Lyu, Dongfu Jiang, Xuan He, Yuan Liu, Hexiang Hu, Xiang Yue, and Wenhu Chen.
\newblock Mega-bench: Scaling multimodal evaluation to over 500 real-world tasks.
\newblock {\em arXiv preprint arXiv:2410.10563}, 2024.

\bibitem{chen2024expanding}
Zhe Chen, Weiyun Wang, Yue Cao, Yangzhou Liu, Zhangwei Gao, Erfei Cui, Jinguo Zhu, Shenglong Ye, Hao Tian, Zhaoyang Liu, et~al.
\newblock Expanding performance boundaries of open-source multimodal models with model, data, and test-time scaling.
\newblock {\em arXiv preprint arXiv:2412.05271}, 2024.

\bibitem{chen2024internvl}
Zhe Chen, Jiannan Wu, Wenhai Wang, Weijie Su, Guo Chen, Sen Xing, Muyan Zhong, Qinglong Zhang, Xizhou Zhu, Lewei Lu, et~al.
\newblock Internvl: Scaling up vision foundation models and aligning for generic visual-linguistic tasks.
\newblock In {\em Proceedings of the IEEE/CVF Conference on Computer Vision and Pattern Recognition}, pages 24185--24198, 2024.

\bibitem{couairon2022diffedit}
Guillaume Couairon, Jakob Verbeek, Holger Schwenk, and Matthieu Cord.
\newblock Diffedit: Diffusion-based semantic image editing with mask guidance.
\newblock {\em arXiv preprint arXiv:2210.11427}, 2022.

\bibitem{dai2023instructblip}
Wenliang Dai, Junnan Li, D Li, AMH Tiong, J Zhao, W Wang, B Li, P Fung, and S Hoi.
\newblock Instructblip: Towards general-purpose vision-language models with instruction tuning. arxiv 2023.
\newblock {\em arXiv preprint arXiv:2305.06500}, 2, 2023.

\bibitem{Damen2021PAMI}
Dima Damen, Hazel Doughty, Giovanni~Maria Farinella, Sanja Fidler, Antonino Furnari, Evangelos Kazakos, Davide Moltisanti, Jonathan Munro, Toby Perrett, Will Price, and Michael Wray.
\newblock The epic-kitchens dataset: Collection, challenges and baselines.
\newblock {\em IEEE Transactions on Pattern Analysis and Machine Intelligence (TPAMI)}, 43(11):4125--4141, 2021.

\bibitem{dhariwal2021diffusion}
Prafulla Dhariwal and Alexander Nichol.
\newblock Diffusion models beat {GAN}s on image synthesis.
\newblock 34:8780--8794, 2021.

\bibitem{dreamlike2023photoreal25}
Dreamlike.art.
\newblock Dreamlike photoreal 2.5, 2023.

\bibitem{esser2403scaling}
Patrick Esser, Sumith Kulal, Andreas Blattmann, Rahim Entezari, Jonas M{\"u}ller, Harry Saini, Yam Levi, Dominik Lorenz, Axel Sauer, Frederic Boesel, et~al.
\newblock Scaling rectified flow transformers for high-resolution image synthesis, march 2024.
\newblock {\em URL http://arxiv. org/abs/2403.03206}, 2024.

\bibitem{fu2023dreamsim}
Stephanie Fu, Netanel Tamir, Shobhita Sundaram, Lucy Chai, Richard Zhang, Tali Dekel, and Phillip Isola.
\newblock Dreamsim: Learning new dimensions of human visual similarity using synthetic data.
\newblock {\em arXiv preprint arXiv:2306.09344}, 2023.

\bibitem{ge2024seeddataedit}
Yuying Ge, Sijie Zhao, Chen Li, Yixiao Ge, and Ying Shan.
\newblock Seed-data-edit technical report: A hybrid dataset for instructional image editing.
\newblock {\em arXiv preprint arXiv:2405.04007}, 2024.

\bibitem{ge2024seedx}
Yuying Ge, Sijie Zhao, Jinguo Zhu, Yixiao Ge, Kun Yi, Lin Song, Chen Li, Xiaohan Ding, and Ying Shan.
\newblock Seed-x: Multimodal models with unified multi-granularity comprehension and generation.
\newblock {\em arXiv preprint arXiv:2404.14396}, 2024.

\bibitem{goyal2017something}
Raghav Goyal, Samira Ebrahimi~Kahou, Vincent Michalski, Joanna Materzynska, Susanne Westphal, Heuna Kim, Valentin Haenel, Ingo Fruend, Peter Yianilos, Moritz Mueller-Freitag, et~al.
\newblock The" something something" video database for learning and evaluating visual common sense.
\newblock In {\em Proceedings of the IEEE international conference on computer vision}, pages 5842--5850, 2017.

\bibitem{greff2021kubric}
Klaus Greff, Francois Belletti, Lucas Beyer, Carl Doersch, Yilun Du, Daniel Duckworth, David~J Fleet, Dan Gnanapragasam, Florian Golemo, Charles Herrmann, Thomas Kipf, Abhijit Kundu, Dmitry Lagun, Issam Laradji, Hsueh-Ti~(Derek) Liu, Henning Meyer, Yishu Miao, Derek Nowrouzezahrai, Cengiz Oztireli, Etienne Pot, Noha Radwan, Daniel Rebain, Sara Sabour, Mehdi S.~M. Sajjadi, Matan Sela, Vincent Sitzmann, Austin Stone, Deqing Sun, Suhani Vora, Ziyu Wang, Tianhao Wu, Kwang~Moo Yi, Fangcheng Zhong, and Andrea Tagliasacchi.
\newblock Kubric: a scalable dataset generator.
\newblock 2022.

\bibitem{gu2024multi}
Xin Gu, Ming Li, Libo Zhang, Fan Chen, Longyin Wen, Tiejian Luo, and Sijie Zhu.
\newblock Multi-reward as condition for instruction-based image editing.
\newblock {\em arXiv preprint arXiv:2411.04713}, 2024.

\bibitem{he2024videoscore}
Xuan He, Dongfu Jiang, Ge Zhang, Max Ku, Achint Soni, Sherman Siu, Haonan Chen, Abhranil Chandra, Ziyan Jiang, Aaran Arulraj, et~al.
\newblock Videoscore: Building automatic metrics to simulate fine-grained human feedback for video generation.
\newblock {\em arXiv preprint arXiv:2406.15252}, 2024.

\bibitem{hertz2022prompt}
Amir Hertz, Ron Mokady, Jay Tenenbaum, Kfir Aberman, Yael Pritch, and Daniel Cohen-Or.
\newblock Prompt-to-{P}rompt image editing with cross attention control.
\newblock {\em arXiv preprint arXiv:2208.01626}, 2022.

\bibitem{hessel2021clipscore}
Jack Hessel, Ari Holtzman, Maxwell Forbes, Ronan~Le Bras, and Yejin Choi.
\newblock Clipscore: A reference-free evaluation metric for image captioning.
\newblock {\em arXiv preprint arXiv:2104.08718}, 2021.

\bibitem{hu2022lora}
Edward~J Hu, Yelong Shen, Phillip Wallis, Zeyuan Allen-Zhu, Yuanzhi Li, Shean Wang, Lu Wang, Weizhu Chen, et~al.
\newblock Lora: Low-rank adaptation of large language models.
\newblock {\em ICLR}, 1(2):3, 2022.

\bibitem{ji2020action}
Jingwei Ji, Ranjay Krishna, Li Fei-Fei, and Juan~Carlos Niebles.
\newblock Action genome: Actions as compositions of spatio-temporal scene graphs.
\newblock In {\em Proceedings of the IEEE/CVF conference on computer vision and pattern recognition}, pages 10236--10247, 2020.

\bibitem{jiang2024genai}
Dongfu Jiang, Max Ku, Tianle Li, Yuansheng Ni, Shizhuo Sun, Rongqi Fan, and Wenhu Chen.
\newblock Genai arena: An open evaluation platform for generative models.
\newblock {\em arXiv preprint arXiv:2406.04485}, 2024.

\bibitem{johnson2017clevr}
Justin Johnson, Bharath Hariharan, Laurens Van Der~Maaten, Li Fei-Fei, C Lawrence~Zitnick, and Ross Girshick.
\newblock Clevr: A diagnostic dataset for compositional language and elementary visual reasoning.
\newblock In {\em Proceedings of the IEEE conference on computer vision and pattern recognition}, pages 2901--2910, 2017.

\bibitem{kamath2023whatsup}
Amita Kamath, Jack Hessel, and Kai-Wei Chang.
\newblock What's ``up'' with vision-language models? investigating their struggle with spatial reasoning.
\newblock In {\em EMNLP}, 2023.

\bibitem{kirstain2023pick}
Yuval Kirstain, Adam Polyak, Uriel Singer, Shahbuland Matiana, Joe Penna, and Omer Levy.
\newblock Pick-a-pic: An open dataset of user preferences for text-to-image generation.
\newblock {\em Advances in Neural Information Processing Systems}, 36:36652--36663, 2023.

\bibitem{krojer2024aurora}
Benno Krojer, Dheeraj Vattikonda, Luis Lara, Varun Jampani, Eva Portelance, Christopher Pal, and Siva Reddy.
\newblock {Learning Action and Reasoning-Centric Image Editing from Videos and Simulations}.
\newblock In {\em NeurIPS}, 2024.
\newblock Spotlight Paper.

\bibitem{ku2023viescore}
Max Ku, Dongfu Jiang, Cong Wei, Xiang Yue, and Wenhu Chen.
\newblock Viescore: Towards explainable metrics for conditional image synthesis evaluation.
\newblock {\em arXiv preprint arXiv:2312.14867}, 2023.

\bibitem{ku2024imagenhub}
Max Ku, Tianle Li, Kai Zhang, Yujie Lu, Xingyu Fu, Wenwen Zhuang, and Wenhu Chen.
\newblock Imagenhub: Standardizing the evaluation of conditional image generation models.
\newblock In {\em The Twelfth International Conference on Learning Representations}, 2024.

\bibitem{lai2023lisa}
Xin Lai, Zhuotao Tian, Yukang Chen, Yanwei Li, Yuhui Yuan, Shu Liu, and Jiaya Jia.
\newblock Lisa: Reasoning segmentation via large language model.
\newblock {\em arXiv preprint arXiv:2308.00692}, 2023.

\bibitem{laurencon2024idefics2}
Hugo Laurençon, Léo Tronchon, and Victor Sanh.
\newblock Introducing idefics2: A powerful 8b vision-language model for the community.
\newblock {\em Hugging Face Blog}, April 2024.

\bibitem{li2024llava}
Bo Li, Yuanhan Zhang, Dong Guo, Renrui Zhang, Feng Li, Hao Zhang, Kaichen Zhang, Peiyuan Zhang, Yanwei Li, Ziwei Liu, et~al.
\newblock Llava-onevision: Easy visual task transfer.
\newblock {\em arXiv preprint arXiv:2408.03326}, 2024.

\bibitem{li2023blip}
Junnan Li, Dongxu Li, Silvio Savarese, and Steven Hoi.
\newblock Blip-2: Bootstrapping language-image pre-training with frozen image encoders and large language models.
\newblock In {\em International conference on machine learning}, pages 19730--19742. PMLR, 2023.

\bibitem{li2024zone}
Shanglin Li, Bohan Zeng, Yutang Feng, Sicheng Gao, Xiuhui Liu, Jiaming Liu, Lin Li, Xu Tang, Yao Hu, Jianzhuang Liu, et~al.
\newblock {ZONE}: Zero-shot instruction-guided local editing.
\newblock pages 6254--6263, 2024.

\bibitem{richhf}
Youwei Liang, Junfeng He, Gang Li, Peizhao Li, Arseniy Klimovskiy, Nicholas Carolan, Jiao Sun, Jordi Pont-Tuset, Sarah Young, Feng Yang, Junjie Ke, Krishnamurthy~Dj Dvijotham, Katie Collins, Yiwen Luo, Yang Li, Kai~J Kohlhoff, Deepak Ramachandran, and Vidhya Navalpakkam.
\newblock Rich human feedback for text-to-image generation.
\newblock In {\em Proceedings of the IEEE/CVF Conference on Computer Vision and Pattern Recognition}, 2024.

\bibitem{lin2024evaluating}
Zhiqiu Lin, Deepak Pathak, Baiqi Li, Jiayao Li, Xide Xia, Graham Neubig, Pengchuan Zhang, and Deva Ramanan.
\newblock Evaluating text-to-visual generation with image-to-text generation.
\newblock In {\em European Conference on Computer Vision}, pages 366--384. Springer, 2024.

\bibitem{liu2024llavanext}
Haotian Liu, Chunyuan Li, Yuheng Li, Bo Li, Yuanhan Zhang, Sheng Shen, and Yong~Jae Lee.
\newblock Llava-next: Improved reasoning, ocr, and world knowledge, January 2024.

\bibitem{liu2023llava}
Haotian Liu, Chunyuan Li, Qingyang Wu, and Yong~Jae Lee.
\newblock Visual instruction tuning.
\newblock In {\em NeurIPS}, 2023.

\bibitem{liu2024visual}
Haotian Liu, Chunyuan Li, Qingyang Wu, and Yong~Jae Lee.
\newblock Visual instruction tuning.
\newblock {\em Advances in neural information processing systems}, 36, 2024.

\bibitem{luo2023latent}
Simian Luo, Yiqin Tan, Longbo Huang, Jian Li, and Hang Zhao.
\newblock Latent {C}onsistency {M}odels: Synthesizing high-resolution images with few-step inference.
\newblock {\em arXiv preprint arXiv:2310.04378}, 2023.

\bibitem{ma2024i2ebench}
Yiwei Ma, Jiayi Ji, Ke Ye, Weihuang Lin, Zhibin Wang, Yonghan Zheng, Qiang Zhou, Xiaoshuai Sun, and Rongrong Ji.
\newblock I2ebench: A comprehensive benchmark for instruction-based image editing.
\newblock {\em arXiv preprint arXiv:2408.14180}, 2024.

\bibitem{meng2021sdedit}
Chenlin Meng, Yutong He, Yang Song, Jiaming Song, Jiajun Wu, Jun-Yan Zhu, and Stefano Ermon.
\newblock Sdedit: Guided image synthesis and editing with stochastic differential equations.
\newblock {\em arXiv preprint arXiv:2108.01073}, 2021.

\bibitem{mirzaei2025watch}
Ashkan Mirzaei, Tristan Aumentado-Armstrong, Marcus~A Brubaker, Jonathan Kelly, Alex Levinshtein, Konstantinos~G Derpanis, and Igor Gilitschenski.
\newblock Watch {Y}our {S}teps: Local image and scene editing by text instructions.
\newblock pages 111--129. Springer, 2025.

\bibitem{mokady2023null}
Ron Mokady, Amir Hertz, Kfir Aberman, Yael Pritch, and Daniel Cohen-Or.
\newblock Null-text {I}nversion for editing real images using guided diffusion models.
\newblock pages 6038--6047, 2023.

\bibitem{oquab2023dinov2}
Maxime Oquab, Timoth{\'e}e Darcet, Th{\'e}o Moutakanni, Huy Vo, Marc Szafraniec, Vasil Khalidov, Pierre Fernandez, Daniel Haziza, Francisco Massa, Alaaeldin El-Nouby, et~al.
\newblock Dinov2: Learning robust visual features without supervision.
\newblock {\em arXiv preprint arXiv:2304.07193}, 2023.

\bibitem{parmar2023zero}
Gaurav Parmar, Krishna Kumar~Singh, Richard Zhang, Yijun Li, Jingwan Lu, and Jun-Yan Zhu.
\newblock Zero-shot image-to-image translation.
\newblock In {\em ACM SIGGRAPH 2023 Conference Proceedings}, pages 1--11, 2023.

\bibitem{podell2023sdxl}
Dustin Podell, Zion English, Kyle Lacey, Andreas Blattmann, Tim Dockhorn, Jonas M{\"u}ller, Joe Penna, and Robin Rombach.
\newblock {SDXL}: Improving latent diffusion models for high-resolution image synthesis.
\newblock {\em arXiv preprint arXiv:2307.01952}, 2023.

\bibitem{radford2021learning}
Alec Radford, Jong~Wook Kim, Chris Hallacy, Aditya Ramesh, Gabriel Goh, Sandhini Agarwal, Girish Sastry, Amanda Askell, Pamela Mishkin, Jack Clark, et~al.
\newblock Learning transferable visual models from natural language supervision.
\newblock In {\em International conference on machine learning}, pages 8748--8763. PMLR, 2021.

\bibitem{ramesh2022hierarchical}
Aditya Ramesh, Prafulla Dhariwal, Alex Nichol, Casey Chu, and Mark Chen.
\newblock Hierarchical text-conditional image generation with clip latents.
\newblock {\em arXiv preprint arXiv:2204.06125}, 1(2):3, 2022.

\bibitem{rombach2022high}
Robin Rombach, Andreas Blattmann, Dominik Lorenz, Patrick Esser, and Bj{\"o}rn Ommer.
\newblock High-resolution image synthesis with latent diffusion models.
\newblock pages 10684--10695, 2022.

\bibitem{sheynin2024emu}
Shelly Sheynin, Adam Polyak, Uriel Singer, Yuval Kirstain, Amit Zohar, Oron Ashual, Devi Parikh, and Yaniv Taigman.
\newblock Emu edit: Precise image editing via recognition and generation tasks.
\newblock In {\em Proceedings of the IEEE/CVF Conference on Computer Vision and Pattern Recognition}, pages 8871--8879, 2024.

\bibitem{smith2023aria}
John Smith, Jane Doe, and Michael Lee.
\newblock Aria: Advancing multimodal ai.
\newblock {\em arXiv preprint arXiv:2310.67890}, 2023.

\bibitem{song2020denoising}
Jiaming Song, Chenlin Meng, and Stefano Ermon.
\newblock Denoising diffusion implicit models.
\newblock {\em arXiv preprint arXiv:2010.02502}, 2020.

\bibitem{sun2025ie}
Shangkun Sun, Bowen Qu, Xiaoyu Liang, Songlin Fan, and Wei Gao.
\newblock Ie-bench: Advancing the measurement of text-driven image editing for human perception alignment.
\newblock {\em arXiv preprint arXiv:2501.09927}, 2025.

\bibitem{team2024gemini}
Gemini Team, Petko Georgiev, Ving~Ian Lei, Ryan Burnell, Libin Bai, Anmol Gulati, Garrett Tanzer, Damien Vincent, Zhufeng Pan, Shibo Wang, et~al.
\newblock Gemini 1.5: Unlocking multimodal understanding across millions of tokens of context.
\newblock {\em arXiv preprint arXiv:2403.05530}, 2024.

\bibitem{tumanyan2023plug}
Narek Tumanyan, Michal Geyer, Shai Bagon, and Tali Dekel.
\newblock Plug-and-{P}lay diffusion features for text-driven image-to-image translation.
\newblock pages 1921--1930, 2023.

\bibitem{wallace2023edict}
Bram Wallace, Akash Gokul, and Nikhil Naik.
\newblock {EDICT}: Exact diffusion inversion via coupled transformations.
\newblock pages 22532--22541, 2023.

\bibitem{wang2024qwen2}
Peng Wang, Shuai Bai, Sinan Tan, Shijie Wang, Zhihao Fan, Jinze Bai, Keqin Chen, Xuejing Liu, Jialin Wang, Wenbin Ge, et~al.
\newblock Qwen2-vl: Enhancing vision-language model's perception of the world at any resolution.
\newblock {\em arXiv preprint arXiv:2409.12191}, 2024.

\bibitem{wang2023cogvlm}
Weihan Wang, Qingsong Lv, Wenmeng Yu, Wenyi Hong, Ji Qi, Yan Wang, Junhui Ji, Zhuoyi Yang, Lei Zhao, Xixuan Song, Jiazheng Xu, Bin Xu, Juanzi Li, Yuxiao Dong, Ming Ding, and Jie Tang.
\newblock Cogvlm: Visual expert for pretrained language models, 2023.

\bibitem{wang2004image}
Zhou Wang, Alan~C Bovik, Hamid~R Sheikh, and Eero~P Simoncelli.
\newblock Image quality assessment: from error visibility to structural similarity.
\newblock {\em IEEE transactions on image processing}, 13(4):600--612, 2004.

\bibitem{wei2024omniedit}
Cong Wei, Zheyang Xiong, Weiming Ren, Xinrun Du, Ge Zhang, and Wenhu Chen.
\newblock Omniedit: Building image editing generalist models through specialist supervision.
\newblock {\em arXiv preprint arXiv:2411.07199}, 2024.

\bibitem{cyclediffusion}
Chen~Henry Wu and Fernando~De la Torre.
\newblock A latent space of stochastic diffusion models for zero-shot image editing and guidance.
\newblock In {\em ICCV}, 2023.

\bibitem{wu2023uncovering}
Qiucheng Wu, Yujian Liu, Handong Zhao, Ajinkya Kale, Trung Bui, Tong Yu, Zhe Lin, Yang Zhang, and Shiyu Chang.
\newblock Uncovering the disentanglement capability in text-to-image diffusion models.
\newblock pages 1900--1910, 2023.

\bibitem{wu2023humanv2}
Xiaoshi Wu, Yiming Hao, Keqiang Sun, Yixiong Chen, Feng Zhu, Rui Zhao, and Hongsheng Li.
\newblock Human preference score v2: A solid benchmark for evaluating human preferences of text-to-image synthesis.
\newblock {\em arXiv preprint arXiv:2306.09341}, 2023.

\bibitem{wu2025multimodal}
Xun Wu, Shaohan Huang, Guolong Wang, Jing Xiong, and Furu Wei.
\newblock Multimodal large language models make text-to-image generative models align better.
\newblock {\em Advances in Neural Information Processing Systems}, 37:81287--81323, 2025.

\bibitem{wu2024multimodal}
Xun Wu, Shaohan Huang, and Furu Wei.
\newblock Multimodal large language model is a human-aligned annotator for text-to-image generation.
\newblock {\em arXiv preprint arXiv:2404.15100}, 2024.

\bibitem{wu2023human}
Xiaoshi Wu, Keqiang Sun, Feng Zhu, Rui Zhao, and Hongsheng Li.
\newblock Human preference score: Better aligning text-to-image models with human preference.
\newblock In {\em Proceedings of the IEEE/CVF International Conference on Computer Vision}, pages 2096--2105, 2023.

\bibitem{xu2024imagereward}
Jiazheng Xu, Xiao Liu, Yuchen Wu, Yuxuan Tong, Qinkai Li, Ming Ding, Jie Tang, and Yuxiao Dong.
\newblock Imagereward: Learning and evaluating human preferences for text-to-image generation.
\newblock {\em Advances in Neural Information Processing Systems}, 36, 2024.

\bibitem{xu2023infedit}
Sihan Xu, Yidong Huang, Jiayi Pan, Ziqiao Ma, and Joyce Chai.
\newblock Inversion-free image editing with natural language.
\newblock 2024.

\bibitem{yao2024minicpm}
Yuan Yao, Tianyu Yu, Ao Zhang, Chongyi Wang, Junbo Cui, Hongji Zhu, Tianchi Cai, Haoyu Li, Weilin Zhao, Zhihui He, et~al.
\newblock Minicpm-v: A gpt-4v level mllm on your phone.
\newblock {\em arXiv preprint arXiv:2408.01800}, 2024.

\bibitem{yue2023mmmu}
Xiang Yue, Yuansheng Ni, Kai Zhang, Tianyu Zheng, Ruoqi Liu, Ge Zhang, Samuel Stevens, Dongfu Jiang, Weiming Ren, Yuxuan Sun, et~al.
\newblock Mmmu: A massive multi-discipline multimodal understanding and reasoning benchmark for expert agi.
\newblock In {\em Proceedings of CVPR}, 2024.

\bibitem{yuksekgonul2022and}
Mert Yuksekgonul, Federico Bianchi, Pratyusha Kalluri, Dan Jurafsky, and James Zou.
\newblock When and why vision-language models behave like bags-of-words, and what to do about it?
\newblock {\em arXiv preprint arXiv:2210.01936}, 2022.

\bibitem{zhang2024long}
Beichen Zhang, Pan Zhang, Xiaoyi Dong, Yuhang Zang, and Jiaqi Wang.
\newblock Long-clip: Unlocking the long-text capability of clip.
\newblock In {\em European Conference on Computer Vision}, pages 310--325. Springer, 2024.

\bibitem{zhang2024magicbrush}
Kai Zhang, Lingbo Mo, Wenhu Chen, Huan Sun, and Yu Su.
\newblock Magicbrush: A manually annotated dataset for instruction-guided image editing.
\newblock {\em Advances in Neural Information Processing Systems}, 36, 2024.

\bibitem{zhang2018unreasonable}
Richard Zhang, Phillip Isola, Alexei~A Efros, Eli Shechtman, and Oliver Wang.
\newblock The unreasonable effectiveness of deep features as a perceptual metric.
\newblock In {\em Proceedings of the IEEE conference on computer vision and pattern recognition}, pages 586--595, 2018.

\bibitem{zhang2024learning}
Sixian Zhang, Bohan Wang, Junqiang Wu, Yan Li, Tingting Gao, Di Zhang, and Zhongyuan Wang.
\newblock Learning multi-dimensional human preference for text-to-image generation.
\newblock In {\em Proceedings of the IEEE/CVF Conference on Computer Vision and Pattern Recognition}, pages 8018--8027, 2024.

\bibitem{zhang2024hive}
Shu Zhang, Xinyi Yang, Yihao Feng, Can Qin, Chia-Chih Chen, Ning Yu, Zeyuan Chen, Huan Wang, Silvio Savarese, Stefano Ermon, et~al.
\newblock Hive: Harnessing human feedback for instructional visual editing.
\newblock In {\em Proceedings of the IEEE/CVF Conference on Computer Vision and Pattern Recognition}, pages 9026--9036, 2024.

\bibitem{zhao2024ultraedit}
Haozhe Zhao, Xiaojian Ma, Liang Chen, Shuzheng Si, Rujie Wu, Kaikai An, Peiyu Yu, Minjia Zhang, Qing Li, and Baobao Chang.
\newblock {UltraEdit}: Instruction-based fine-grained image editing at scale.
\newblock {\em arXiv preprint arXiv:2407.05282}, 2024.

\bibitem{zheng2024llamafactory}
Yaowei Zheng, Richong Zhang, Junhao Zhang, Yanhan Ye, Zheyan Luo, Zhangchi Feng, and Yongqiang Ma.
\newblock Llamafactory: Unified efficient fine-tuning of 100+ language models.
\newblock In {\em Proceedings of the 62nd Annual Meeting of the Association for Computational Linguistics (Volume 3: System Demonstrations)}, Bangkok, Thailand, 2024. Association for Computational Linguistics.

\end{thebibliography}
}

\clearpage
\noindent\textbf{\Large Appendix}
\vspace{0.1in}
\appendix
\setcounter{figure}{8}
\setcounter{table}{3}
\setcounter{equation}{4}

In this supplementary, we first compare our model with a baseline VLM in the input and target image caption generation task for prompt-guided image editing models used in our dataset creation approach (Sec.~\ref{sec:caption_generation_supplementary}). Then, we provide more details regarding the experiment we have run, including dataset creation and model implementation in Sec.~\ref{sec:dataset_implementation_details_supplementary}, more information about the benchmarks we have tested (Sec.~\ref{sec:benchmark_details_supplementary}), and lastly additional results from our scorer and editing models in Sec.~\ref{sec:additional_results_supplementary}

\section{Caption Generation for Prompt-Guided Editing Model\label{sec:caption_generation_supplementary}}

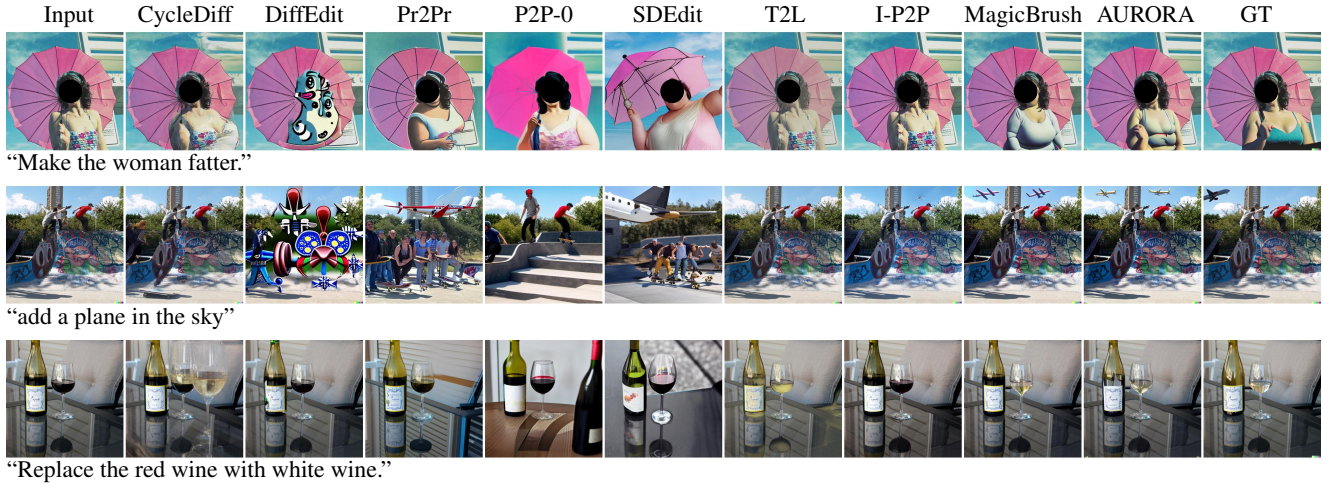
\begin{figure*}[]
    \centering

    \begin{small}
        \begin{tabbing}
            \hspace{1.3em} \= \hspace{3.7em} \= \hspace{5.1em} \= \hspace{5.3em} \= \hspace{4.7em} \= \hspace{4.7em} \= \hspace{5.3em} \= \hspace{4.4em} \= \hspace{3.5em} \= \hspace{5.3em} \= \hspace{5.8em} \=  \kill
            \> Input \> CycleDiff \> DiffEdit \> Pr2Pr \> P2P-0 \> SDEdit \> T2L \> I-P2P \> MagicBrush \> AURORA \> GT
        \end{tabbing}    
    \end{small}

    \vspace{-0.15in} 
    \begin{subfigure}{\linewidth}
        \centering
        \begin{tikzpicture}
            \node[anchor=south west, inner sep=0] (image) at (0,0) {\includegraphics[width=\textwidth, trim=0 180px 0 30px, clip]{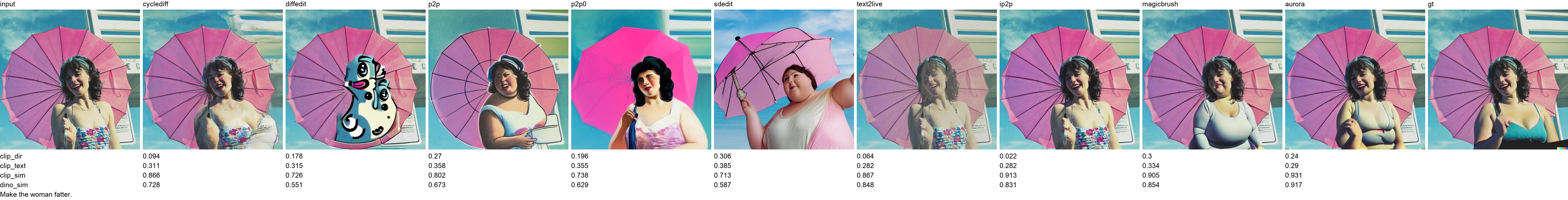}};
    
            \draw[fill=black, opacity=1.0] (0.85,0.8) circle [radius=0.15];
            \draw[fill=black, opacity=1.0] (2.45,0.8) circle [radius=0.15];
            \draw[fill=black, opacity=1.0] (5.6,0.8) circle [radius=0.15];
            \draw[fill=black, opacity=1.0] (7.25,0.8) circle [radius=0.15];
            \draw[fill=black, opacity=1.0] (8.85,0.8) circle [radius=0.15];
            \draw[fill=black, opacity=1.0] (10.4,0.8) circle [radius=0.15];
            \draw[fill=black, opacity=1.0] (12.0,0.8) circle [radius=0.15];
            \draw[fill=black, opacity=1.0] (13.55,0.8) circle [radius=0.15];
            \draw[fill=black, opacity=1.0] (15.15,0.8) circle [radius=0.15];
            \draw[fill=black, opacity=1.0] (16.75,0.8) circle [radius=0.15];
        \end{tikzpicture}
        \parbox{\linewidth}{
            \parbox{\linewidth}{\vspace{-0.1in}{\small``Make the woman fatter.''}\\\vspace{-0.15in}}
        }
    \end{subfigure}

    \begin{subfigure}{\linewidth}
        \centering
        \includegraphics[width=\linewidth, trim=0 180px 0 30px, clip]{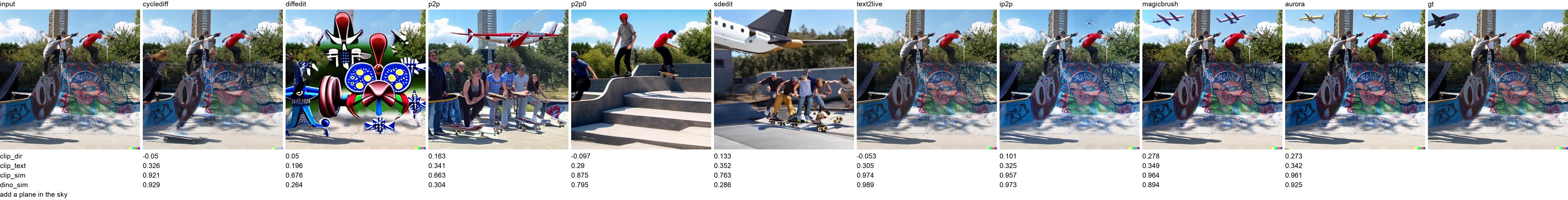}
        \parbox{\linewidth}{
            \parbox{\linewidth}{\vspace{-0.1in}{\small``add a plane in the sky''}\\\vspace{-0.15in}}
        }
    \end{subfigure}

    \begin{subfigure}{\linewidth}
        \centering
        \includegraphics[width=\linewidth, trim=0 180px 0 30px, clip]{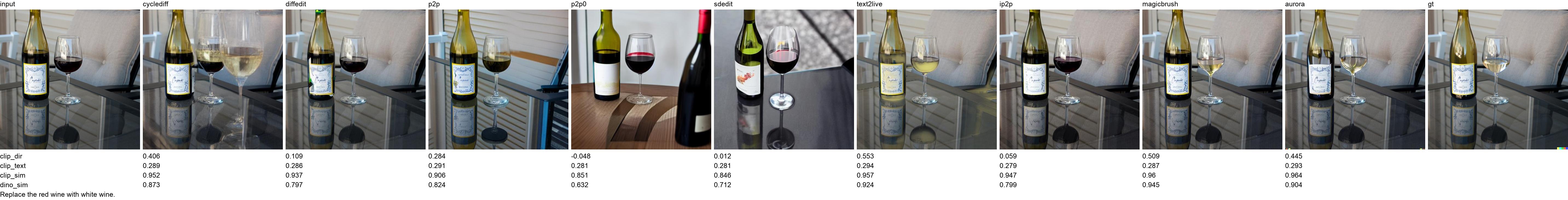}
        \parbox{\linewidth}{
            \parbox{\linewidth}{\vspace{-0.1in}{\small``Replace the red wine with white wine.''}\\\vspace{-0.15in}}
        }
    \end{subfigure}
    \vspace{-0.25in}
    \caption{We use 9 text-guided image editing methods - CycleDiffusion (CycleDiff)~\cite{cyclediffusion}, DiffEdit~\cite{couairon2022diffedit}, Prompt-to-Prompt (Pr2Pr)~\cite{hertz2022prompt}, pix2pix-zero (P2P-0)~\cite{parmar2023zero}, SDEdit~\cite{meng2021sdedit}, Text2LIVE (T2L)~\cite{bar2022text2live}, InstructPix2Pix (I-P2P)~\cite{brooks2023instructpix2pix}, MagicBrush~\cite{zhang2024magicbrush}, and AURORA~\cite{krojer2024aurora} - to generate samples (faces blocked for privacy concerns) with various editing quality as part of our evaluation training data along with the ground-truth (GT) outputs. For methods that need input and target prompts to perform edits, we trained a VLM to get these prompts as shown in Table~\ref{tab:image_captioning_more}.}
    \label{fig:data_creation_synthetic_supplementary}
\end{figure*}

As mentioned in the method section, we need to generate input and target image captions in order to generate training samples from prompt-guided image editing models. Table~\ref{tab:image_captioning_more} shows captions generated from a baseline VLM and from our fine-tuned VLM with respect to images in Fig.~\ref{fig:data_creation_synthetic_supplementary}. As we can see, our model produces captions whose differences are more aligned with the edit instructions.

\begin{table}[b]
    \centering
    \small
    \renewcommand{\arraystretch}{1.1} 
    
    \begin{tabularx}{\linewidth}{p{1.05cm}|X}
    \hline\hline
    \multicolumn{2}{c}{\textbf{Edit instruction:} Make the woman fatter.} \\
    \hline
    \multirow{2}{=}{\textbf{Baseline}} & A woman is holding a pink umbrella and smiling. \\ \cdashline{2-2}
    & A woman \textbf{in a blue dress} holds a pink umbrella, \textbf{standing in front of a building with a blue sky in the background}. \\ 
    \hline
    \multirow{2}{=}{\textbf{Ours}} & a woman holding a pink umbrella \\ \cdashline{2-2}
    & a \textbf{fat} woman holding a pink umbrella \\ 
    
    \hline\hline
    \multicolumn{2}{c}{\textbf{Edit instruction:} add a plane in the sky} \\
    \hline
    \multirow{2}{=}{\textbf{Baseline}} & \textbf{Two skateboarders are performing tricks} on a graffiti-covered ramp in an \textbf{outdoor skate} park.  \\ \cdashline{2-2}
    & \textbf{A group of people are skateboarding} on a graffiti-covered ramp in a park, \textbf{with an airplane flying overhead and a tall building in the background}. \\ 
    \hline
    \multirow{2}{=}{\textbf{Ours}} & a group of people skateboarding down a ramp \\ \cdashline{2-2}
    & a group of people skateboarding down a ramp \textbf{with a small airplane flying in the sky} \\

    \hline\hline
    \multicolumn{2}{c}{\textbf{Edit instruction:} Replace the red wine with white wine.} \\
    \hline
    \multirow{2}{=}{\textbf{Baseline}} & A bottle of Cupcake wine \textbf{is placed next to a glass of red wine} on a glass table, \textbf{with a cushioned chair in the background}. \\ \cdashline{2-2}
    & A bottle of Cupcake wine \textbf{and a partially filled wine glass sit} on a glass table \textbf{outdoors}. \\ 
    \hline
    \multirow{2}{=}{\textbf{Ours}} & a glass of \textbf{red} wine and a bottle of wine on a table \\ \cdashline{2-2}
    & a glass of \textbf{white} wine and a bottle of wine on a table \\ 
    \hline\hline
    \end{tabularx}
    \caption{Input prompt (above the dashline) and target prompt (below the dashline) generated by a baseline VLM~\cite{wang2024qwen2} and our model corresponding to the edit instruction and input/groundtruth (GT) images in Fig.~\ref{fig:data_creation_synthetic_supplementary}. The difference between each pair of input and target prompt is in bold. As we can see, our model produces prompt pairs whose difference is much more closely aligned with the edit instructions compared to the baseline.}
    \label{tab:image_captioning_more}
\end{table}

\section{Dataset and Implementation Details\label{sec:dataset_implementation_details_supplementary}}

In our training dataset, for samples generate using text-guided image editing models, we use the model configuration as specified in ImagenHub~\cite{ku2024imagenhub}. We also use the same configuration for our fine-tuned MagicBrush model~\cite{zhang2024magicbrush} to ensure a fair comparison with the baseline model. 

When using LLaMA-Factory~\cite{zheng2024llamafactory} to fine-tune a Qwen2-VL-7B-Instruct~\cite{wang2024qwen2} for prompt generation, we follow the existing supervised fine-tuning (SFT) setting in the provided example in their code-base, which uses LoRA rank 8, batch size 8, and learning rate 1e-4. 

When training the evaluation scorer, we create several different question templates to prompt the VLM:
\begin{itemize}
    \item ``Can you rate how successful the edit instruction [INSTRUCTION] has been executed from the first image to the second image with a score from 0 to 10?''
    \item ``Please rate how successful the edit instruction [INSTRUCTION] has been executed from the first image to the second image with a score from 0 to 10.'' 
    \item "How successful the edit instruction [INSTRUCTION] has been executed from the first image to the second image? Please respond with a score from 0 to 10."
    \item "How successful the edit instruction [INSTRUCTION] has been executed from the first image to the second image? Please output a score from 0 to 10.",
\end{itemize}
where ``[INSTRUCTION]'' is a place-holder that will be replaced by the actual edit instruction with respect to the input and edited images. 

Similar, we also create a few answering templates:
\begin{itemize}
    \item ``It is [SCORE].'' 
    \item ``Sure, [SCORE]''
    \item ``Sure, it is [SCORE]''
    \item ``Sure, the score is [SCORE]''
    \item ``[SCORE]'',
\end{itemize}
where ``[SCORE]'' is a special token that will be decoded to the final evaluation score. During training, we randomly select a question and an answering template for each sample so the scorer is not over-fitted on fixed prompts. 

The scorer training loss consists of two parts. The first part is the auto-regressive cross-entropy loss for VLM generated text $\widehat{\mathbf{y}}_{txt}$ with respect to the selected answering template $\mathbf{y}_{txt}$, which is defined in LISA~\cite{lai2023lisa} as:
\begin{equation}
    \mathcal{L}_{txt} = \text{CE}(\widehat{\mathbf{y}}_{txt}, \mathbf{y}_{txt}).
\end{equation}

The second part is the loss between the predicted score $\hat{s}$ and the ground-truth score $s$:
\begin{equation}
    \mathcal{L}_{score} = \text{L1}(\hat{s}, s).
\end{equation}

The total training loss is 
\begin{equation}
    \mathcal{L} = \lambda_{text}\mathcal{L}_{txt} + \lambda_{score}\mathcal{L}_{score},
\end{equation}
where we set $\lambda_{text} = 1$ and $\lambda_{score} = 10$.

To fine-tune MagicBrush~\cite{zhang2024magicbrush} using our scorer as a reward model in the reward condition setting, we take all samples in our evaluation training set that corresponds to input images and the edit instructions in MagicBrush training set to re-label them with reward scores predicted by our scorer. These samples become the training set to fine-tune MagicBrush on. 

To fine-tune the model in the reward feedback learning setting, we train the model with a weighted sum of the reward feedback learning loss $\mathcal{L}_{reward}$ and the original diffusion model MSE loss $\mathcal{L}){pre}$ as discussed in ImageReward~\cite{xu2024imagereward}, where the total training loss is:
\begin{equation}
    \mathcal{L}_{total} = \mathcal{L}_{pre} + \lambda_{reward}\mathcal{L}_{reward}
\end{equation}
for $\lambda_{reward} = 0.001$ so the weighted reward learning loss is balanced with the MSE loss for stable training. 

\section{Benchmarks Details\label{sec:benchmark_details_supplementary}}

ImagenHub consists of 179 text-guided image editing samples, each containing an input image, edit instruction, input and target prompts, and a ground-truth output. Each sample includes editing outputs from eight methods~\cite{cyclediffusion,couairon2022diffedit,brooks2023instructpix2pix,zhang2024magicbrush,hertz2022prompt,parmar2023zero,meng2021sdedit,bar2022text2live}, evaluated by three human raters following a quantized scoring scheme: 0, 0.5, and 1. Following VIEScore~\cite{ku2023viescore}, we compute the Spearman correlation between human ratings and predicted scores per method, applying Fisher Z-transformation to obtain the average correlation. The inter-rater (Human-to-Human) Spearman correlation serves as the upper bound for evaluator performance. 

GenAI-Bench comprises 919 samples, each with an input image, edit instruction, two edit outputs from different methods, and a human preference label indicating which output is preferred or if both are good/bad. As it uses images in ImagenHub, each output is paired with a ground-truth edit.

AURORA-Bench contains two parts: (1) 2,000 point-wise evaluation samples with input images, edit instructions, edit outputs, and human-averaged quality scores ranging from 0–2, which we refer to as AURORA-Bench (point-wise); (2) 1,600 pair-wise comparison samples in the same format as GenAI-Bench but only distinguishes between preferred and tied outputs, and we refer them as AURORA-Bench (pair-wise). Samples from AURORA-Bench are collected from multiple sources: MagicBrush~\cite{zhang2024magicbrush}, Action-Genome~\cite{ji2020action}, Something-Something~\cite{goyal2017something}, Epic-Kitchen~\cite{Damen2021PAMI}, Kubric~\cite{greff2021kubric}, CLEVR~\cite{johnson2017clevr}, WhatsUp~\cite{kamath2023whatsup}, and Emu-Edit~\cite{sheynin2024emu}.

\section{Additional Results\label{sec:additional_results_supplementary}}

\begin{figure*}
    \centering
    \captionsetup[subfigure]{labelformat=empty}
    
    \begin{small}
        \begin{tabbing}
            \hspace{2.0em} \= \hspace{5.7em} \= \hspace{7.4em} \= \hspace{6.8em} \= \hspace{7.2em} \= \hspace{5.8em} \= \hspace{7.4em} \= \hspace{6.7em} \=  \kill
            \> Input \> Baseline \> Ours \> GT \> Input \> Baseline \> Ours \> GT
        \end{tabbing}    
    \end{small}
    \vspace{-0.15in}

    \begin{subfigure}[c]{0.49\textwidth}
        \includegraphics[width=\textwidth, trim=0 30px 0 30px, clip]{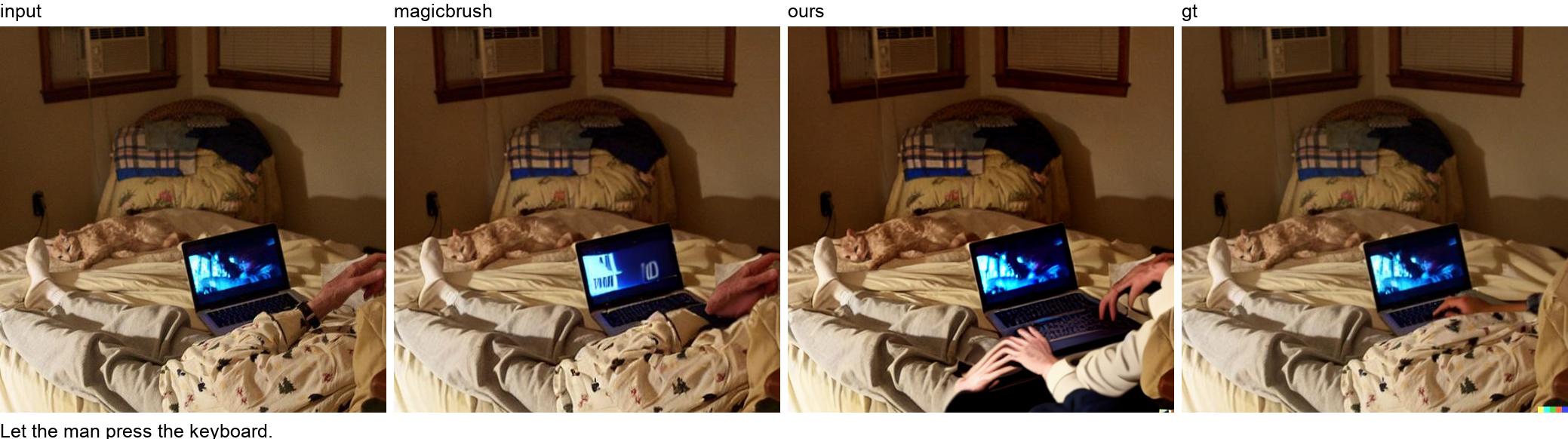}
        \vspace{-0.25in}
        \caption{``Let the man press the keyboard.''}
    \end{subfigure}
    \hfill
    \begin{subfigure}[c]{0.49\textwidth}
        \includegraphics[width=\textwidth, trim=0 30px 0 30px, clip]{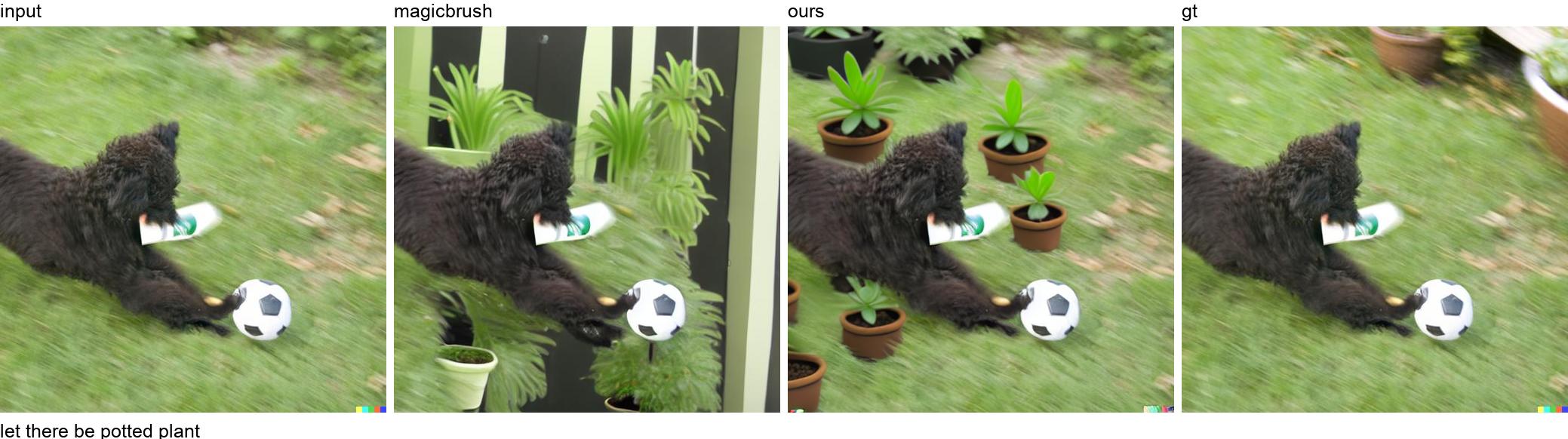}
        \vspace{-0.25in}
        \caption{``Let there be potted plants'}
    \end{subfigure}

    \begin{subfigure}[c]{0.49\textwidth}
        \includegraphics[width=\textwidth, trim=0 30px 0 30px, clip]{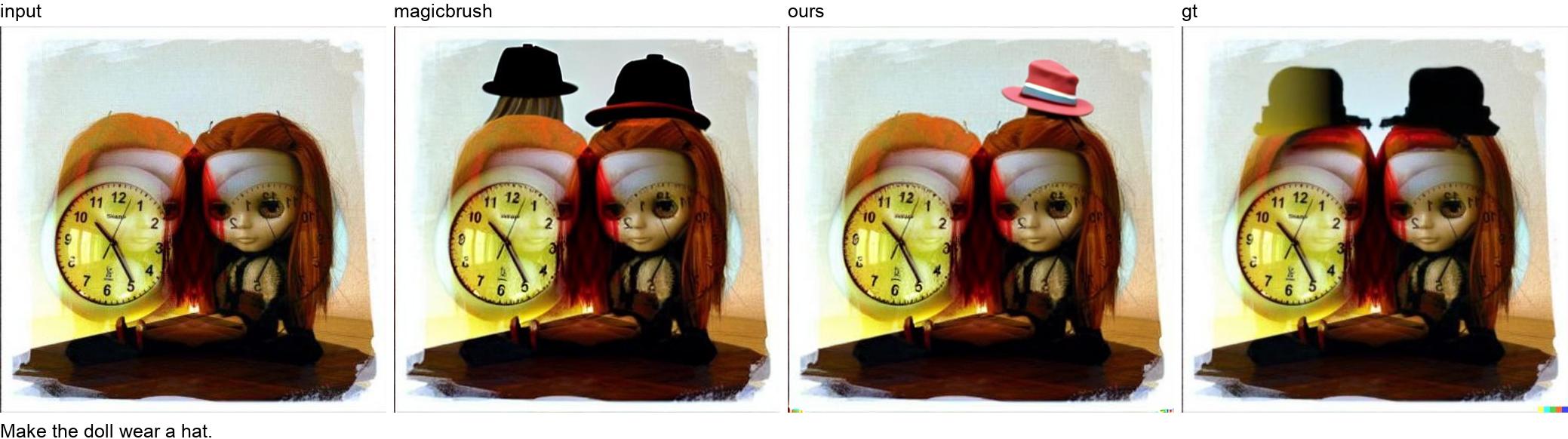}
        \vspace{-0.25in}
        \caption{``Make the doll wear a hat.''}
    \end{subfigure}
    \hfill
    \begin{subfigure}[c]{0.49\textwidth}
        \includegraphics[width=\textwidth, trim=0 30px 0 30px, clip]{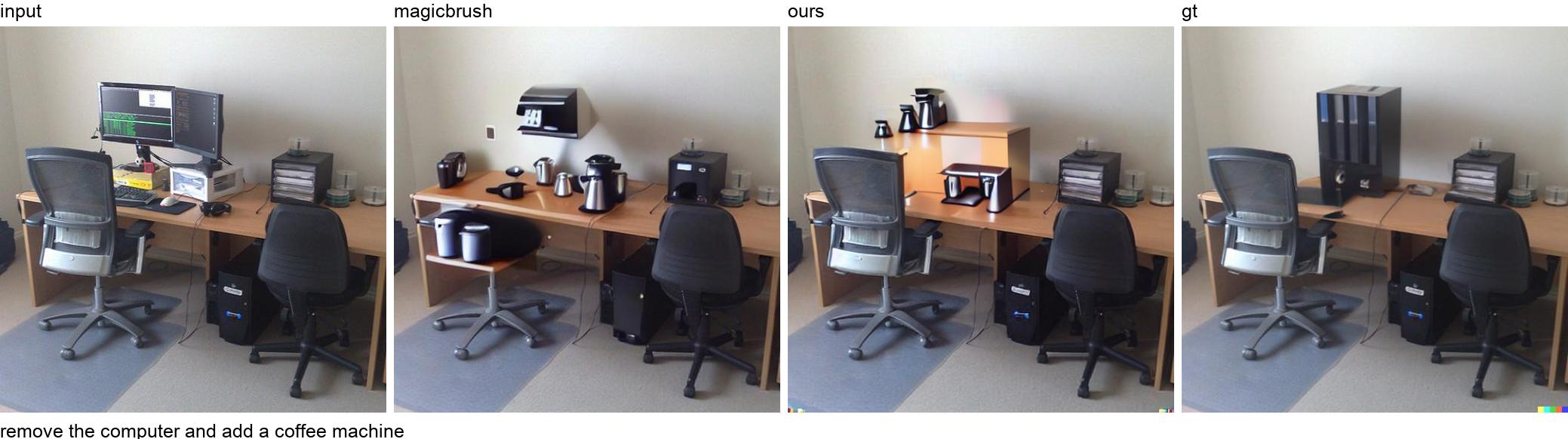}
        \vspace{-0.25in}
        \caption{``remove the computer and add a coffee machine''}
    \end{subfigure}

    \begin{subfigure}[c]{0.49\textwidth}
        \includegraphics[width=\textwidth, trim=0 30px 0 30px, clip]{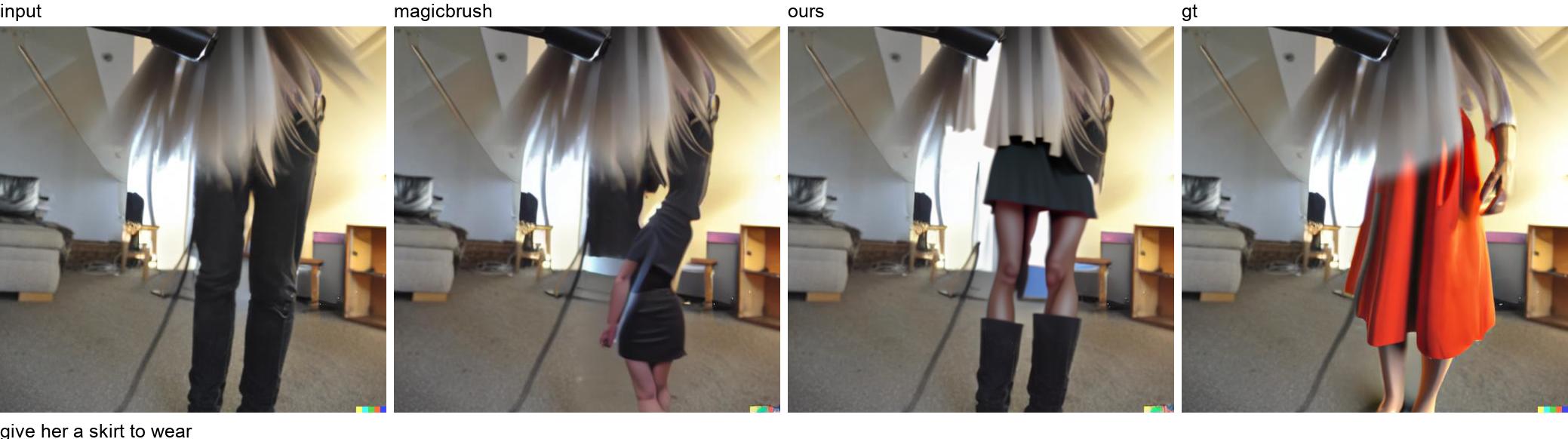}
        \vspace{-0.25in}
        \caption{``give her a skirt to wear''}
    \end{subfigure}
    \hfill
    \begin{subfigure}[c]{0.49\textwidth}
        \includegraphics[width=\textwidth, trim=0 30px 0 30px, clip]{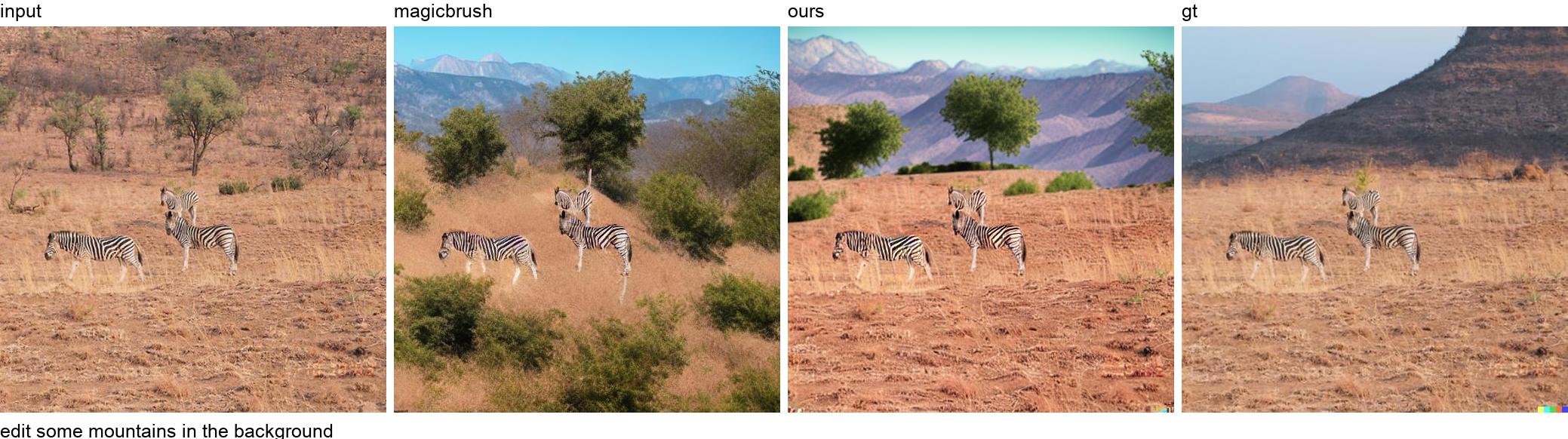}
        \vspace{-0.25in}
        \caption{``edit some mountains in the background''}
    \end{subfigure}

    \begin{subfigure}[c]{0.49\textwidth}
        \includegraphics[width=\textwidth, trim=0 30px 0 30px, clip]{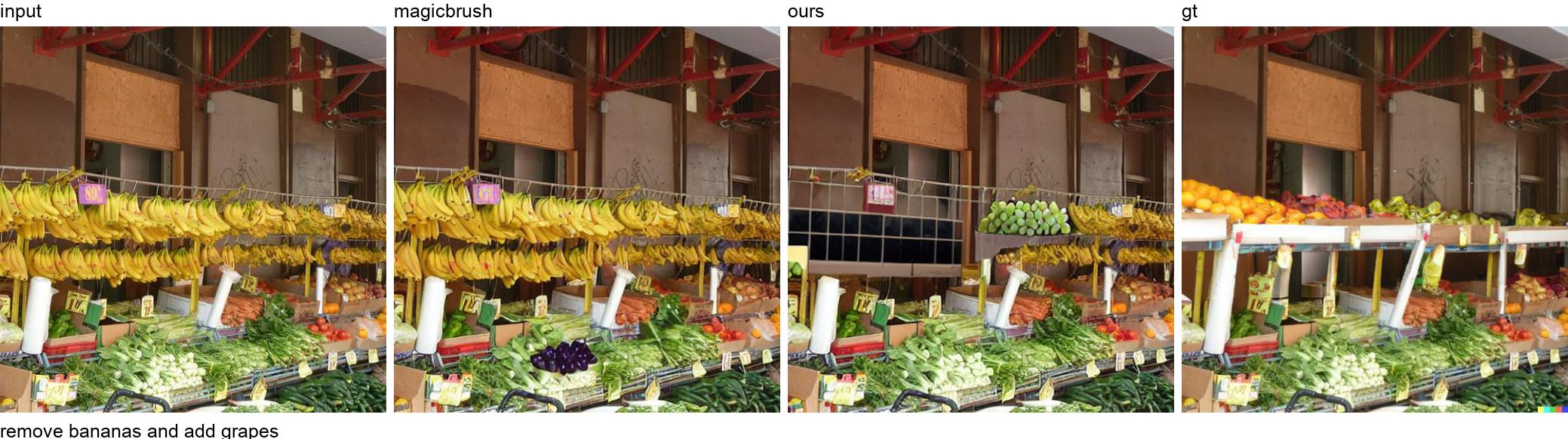}
        \vspace{-0.25in}
        \caption{``remove bananas and add grapes''}
    \end{subfigure}
    \hfill
    \begin{subfigure}[c]{0.49\textwidth}
        \includegraphics[width=\textwidth, trim=0 30px 0 30px, clip]{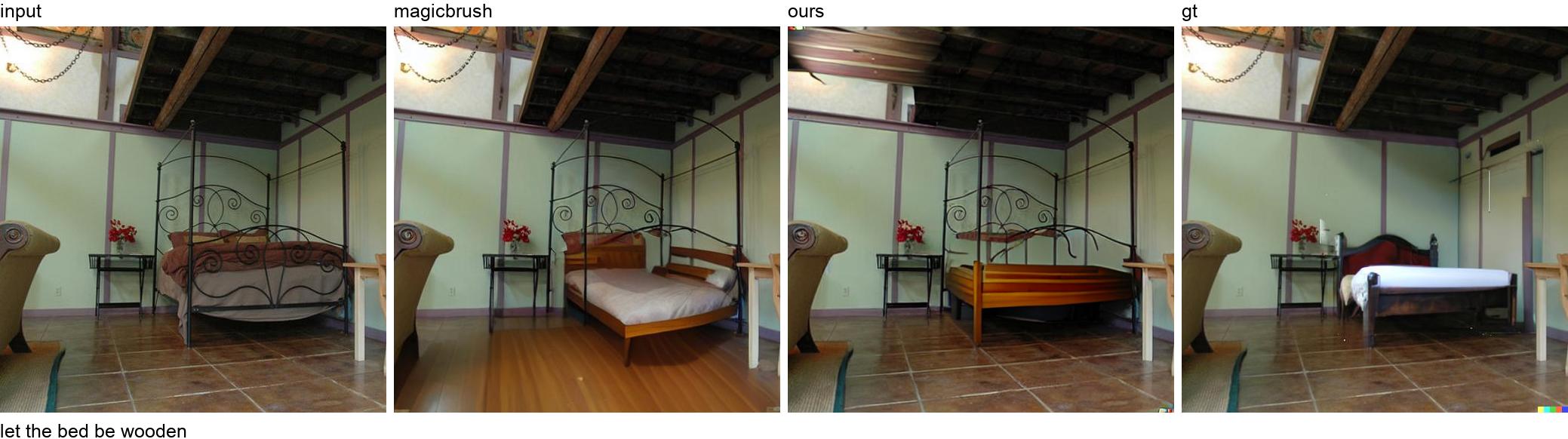}
        \vspace{-0.25in}
        \caption{``let the bed be wooden''}
    \end{subfigure}

    \begin{subfigure}[c]{0.49\textwidth}
        \includegraphics[width=\textwidth, trim=0 30px 0 30px, clip]{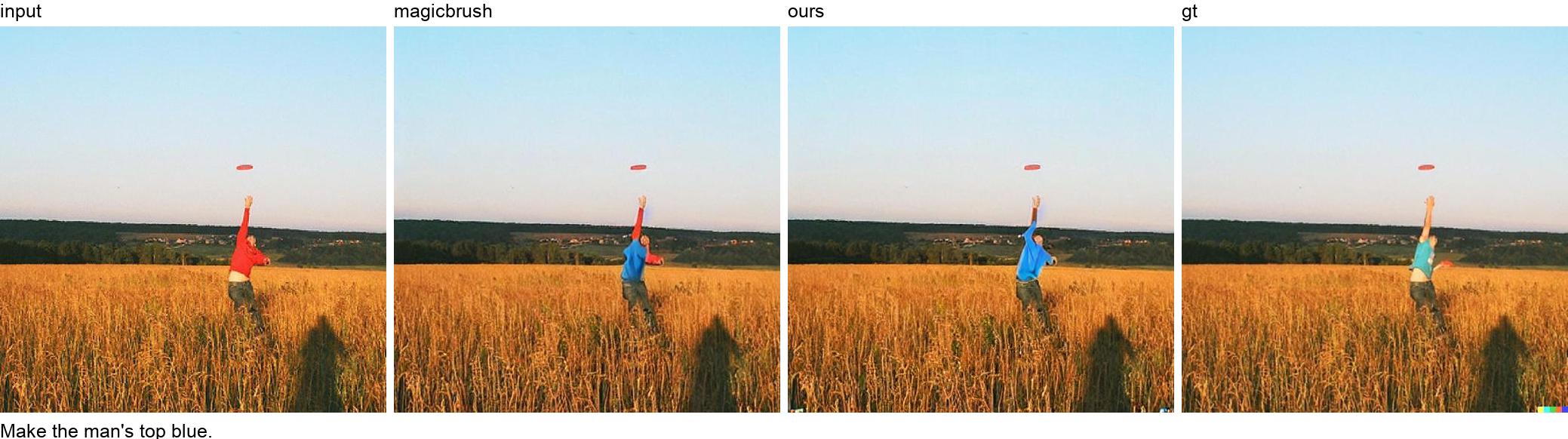}
        \vspace{-0.25in}
        \caption{``Make the man's top blue''}
    \end{subfigure}
    \hfill
    \begin{subfigure}[c]{0.49\textwidth}
        \includegraphics[width=\textwidth, trim=0 30px 0 30px, clip]{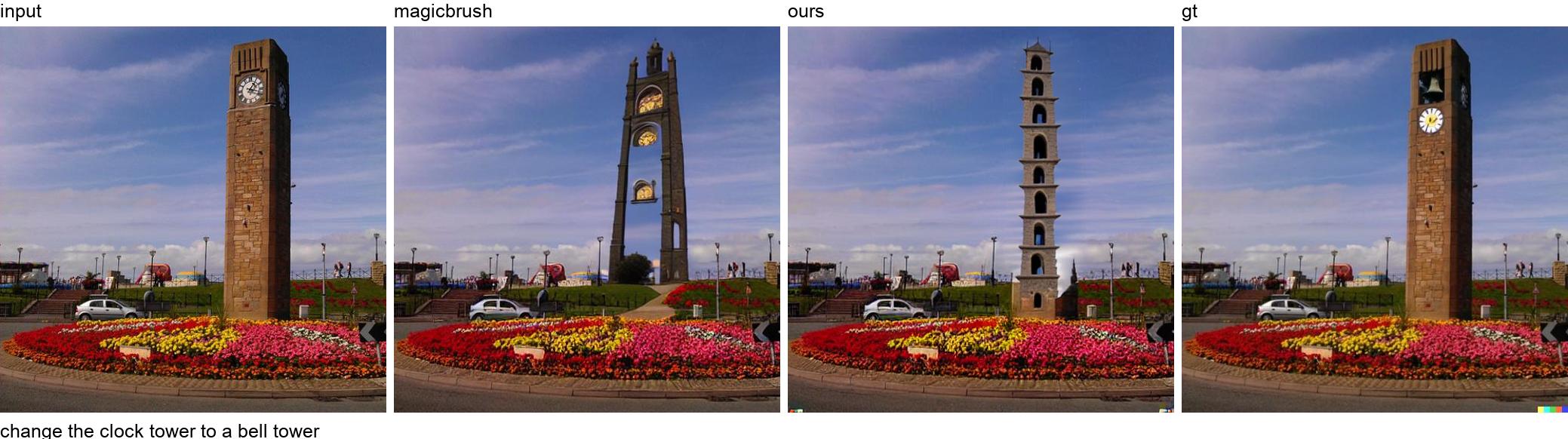}
        \vspace{-0.25in}
        \caption{``change the clock tower to a bell tower''}
    \end{subfigure}

    \begin{subfigure}[c]{0.49\textwidth}
        \includegraphics[width=\textwidth, trim=0 30px 0 30px, clip]{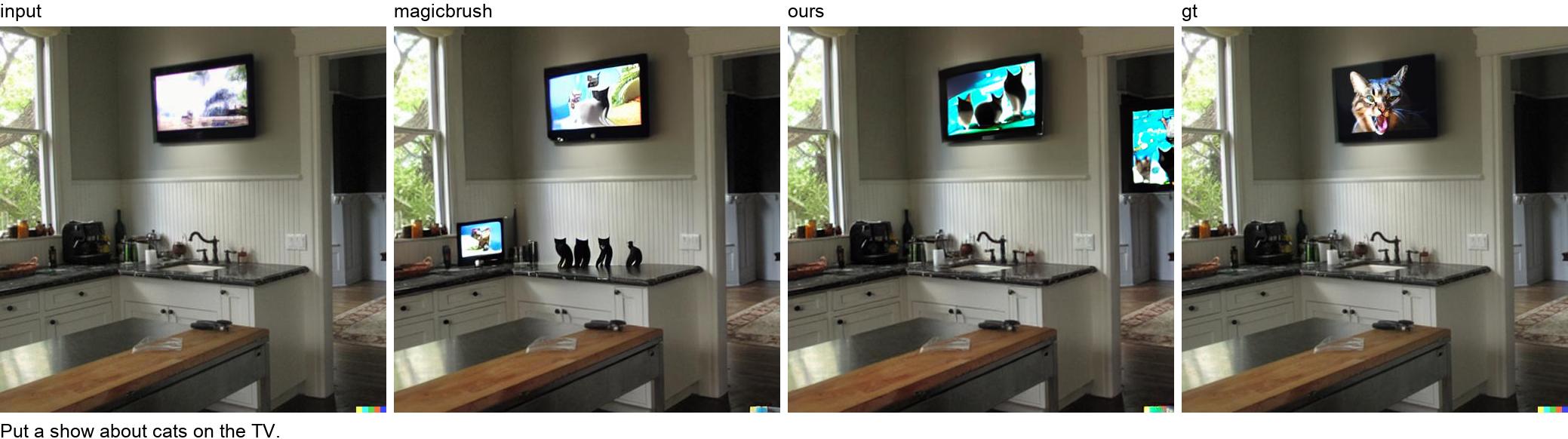}
        \vspace{-0.25in}
        \caption{``Put a show about cats on the TV.''}
    \end{subfigure}
    \hfill
    \begin{subfigure}[c]{0.49\textwidth}
    \begin{tikzpicture}
        \node[anchor=south west, inner sep=0] (image) at (0,0) {\includegraphics[width=\textwidth, trim=0 30px 0 30px, clip]{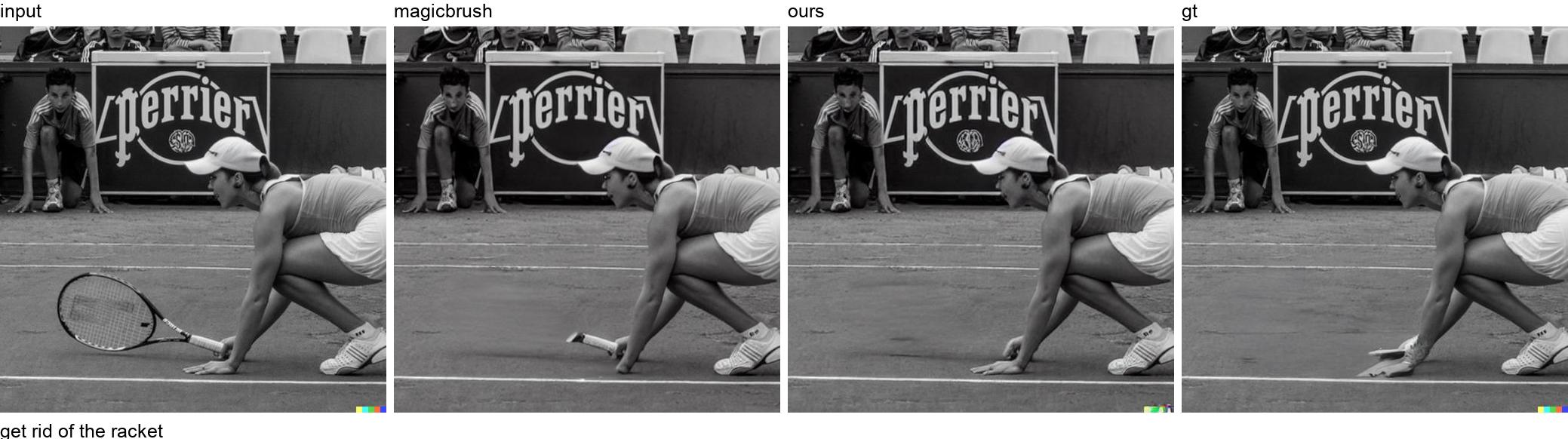}};
        \draw[fill=black, opacity=1.0] (1.2,1.25) circle [radius=0.1];
        \draw[fill=black, opacity=1.0] (0.35,1.75) circle [radius=0.1];

        \draw[fill=black, opacity=1.0] (3.35,1.25) circle [radius=0.1];
        \draw[fill=black, opacity=1.0] (2.5,1.75) circle [radius=0.1];

        \draw[fill=black, opacity=1.0] (5.5,1.25) circle [radius=0.1];
        \draw[fill=black, opacity=1.0] (4.65,1.75) circle [radius=0.1];

        \draw[fill=black, opacity=1.0] (7.65,1.25) circle [radius=0.1];
        \draw[fill=black, opacity=1.0] (6.8,1.75) circle [radius=0.1];
    \end{tikzpicture}
        \vspace{-0.25in}
        \caption{``get rid of the racket''}
    \end{subfigure}

    \begin{subfigure}[c]{0.49\textwidth}
        \includegraphics[width=\textwidth, trim=0 30px 0 30px, clip]{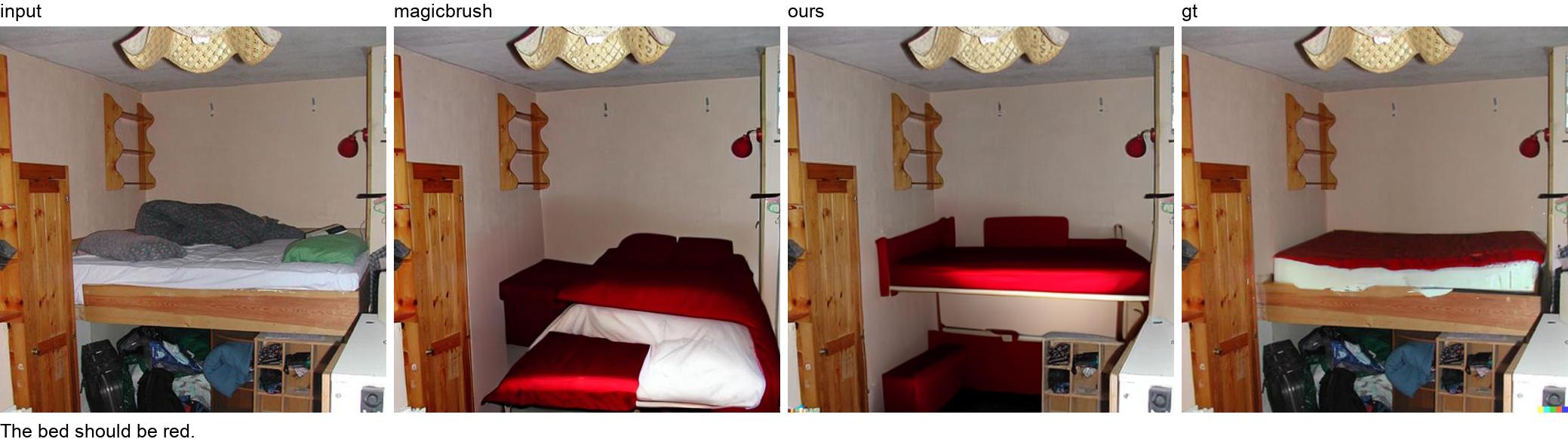}
        \vspace{-0.25in}
        \caption{``The bed should be red.''}
    \end{subfigure}
    \hfill
    \begin{subfigure}[c]{0.49\textwidth}
        \includegraphics[width=\textwidth, trim=0 30px 0 30px, clip]{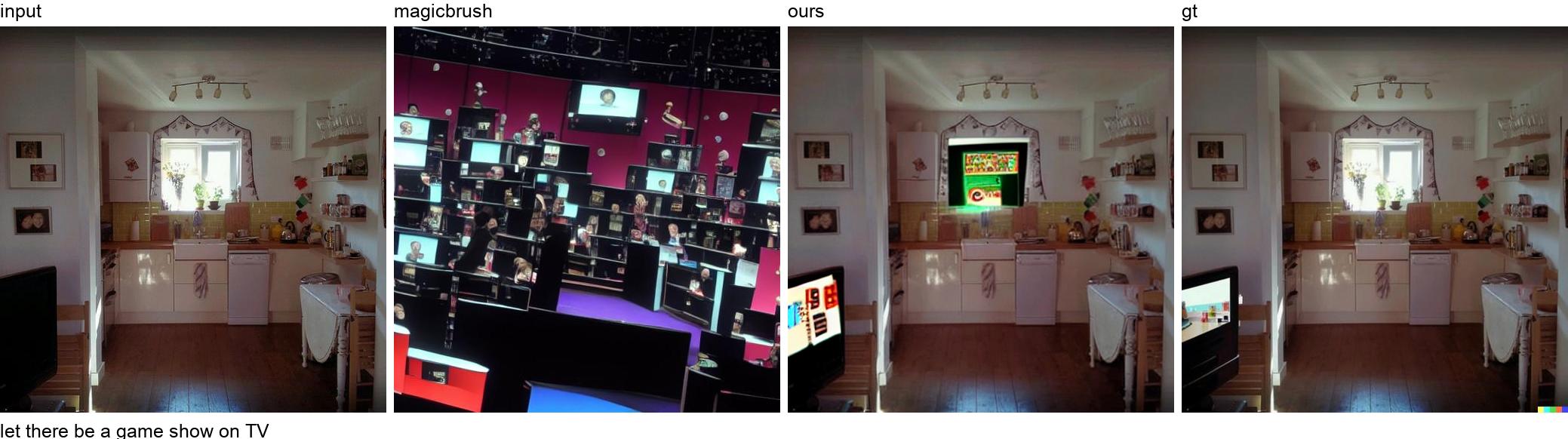}
        \vspace{-0.25in}
        \caption{``Let there be a game show on TV''}
    \end{subfigure}

    \begin{subfigure}[c]{0.49\textwidth}
        \includegraphics[width=\textwidth, trim=0 30px 0 30px, clip]{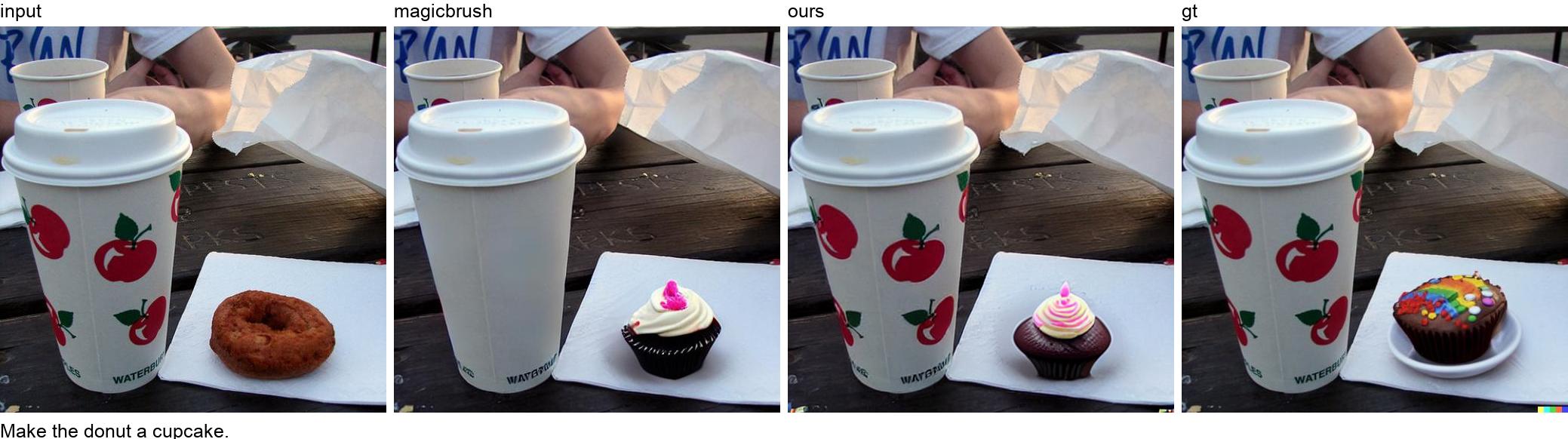}
        \vspace{-0.25in}
        \caption{``Make the donut a cupcake.''}
    \end{subfigure}
    \hfill
    \begin{subfigure}[c]{0.49\textwidth}
        \includegraphics[width=\textwidth, trim=0 30px 0 30px, clip]{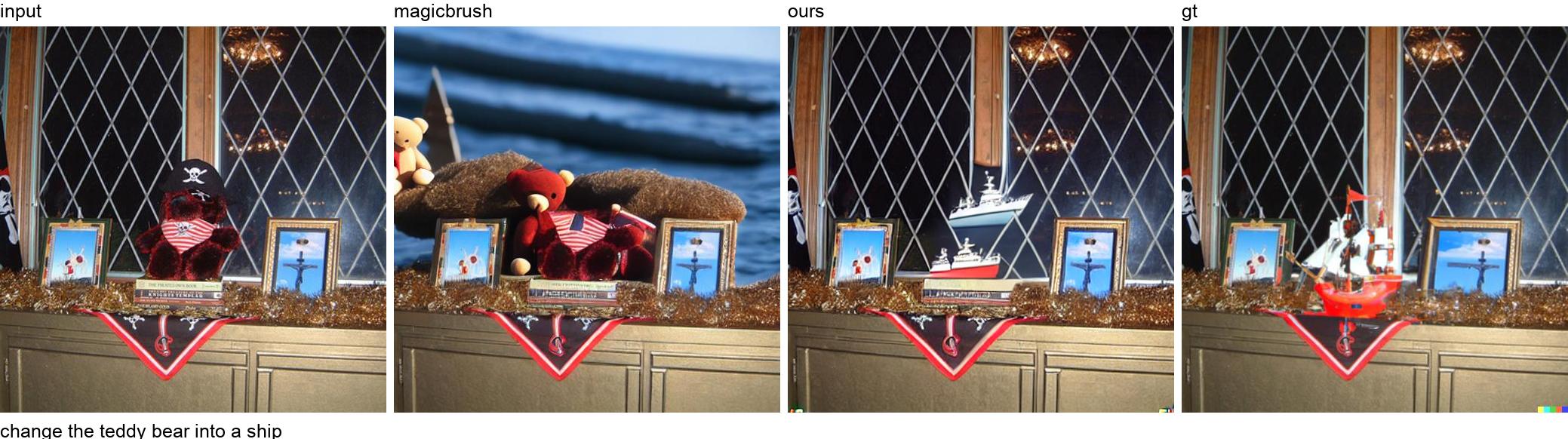}
        \vspace{-0.25in}
        \caption{``change the teddy bear into a ship''}
    \end{subfigure}

    \caption{More mage editing examples (faces are blocked due to privacy concerns) from the MagicBrush editing model~\cite{zhang2024magicbrush} and our fine-tuned model using our image editing scorer as a reward model.}
    \label{fig:image_editing_reward_supplementary}
\end{figure*}
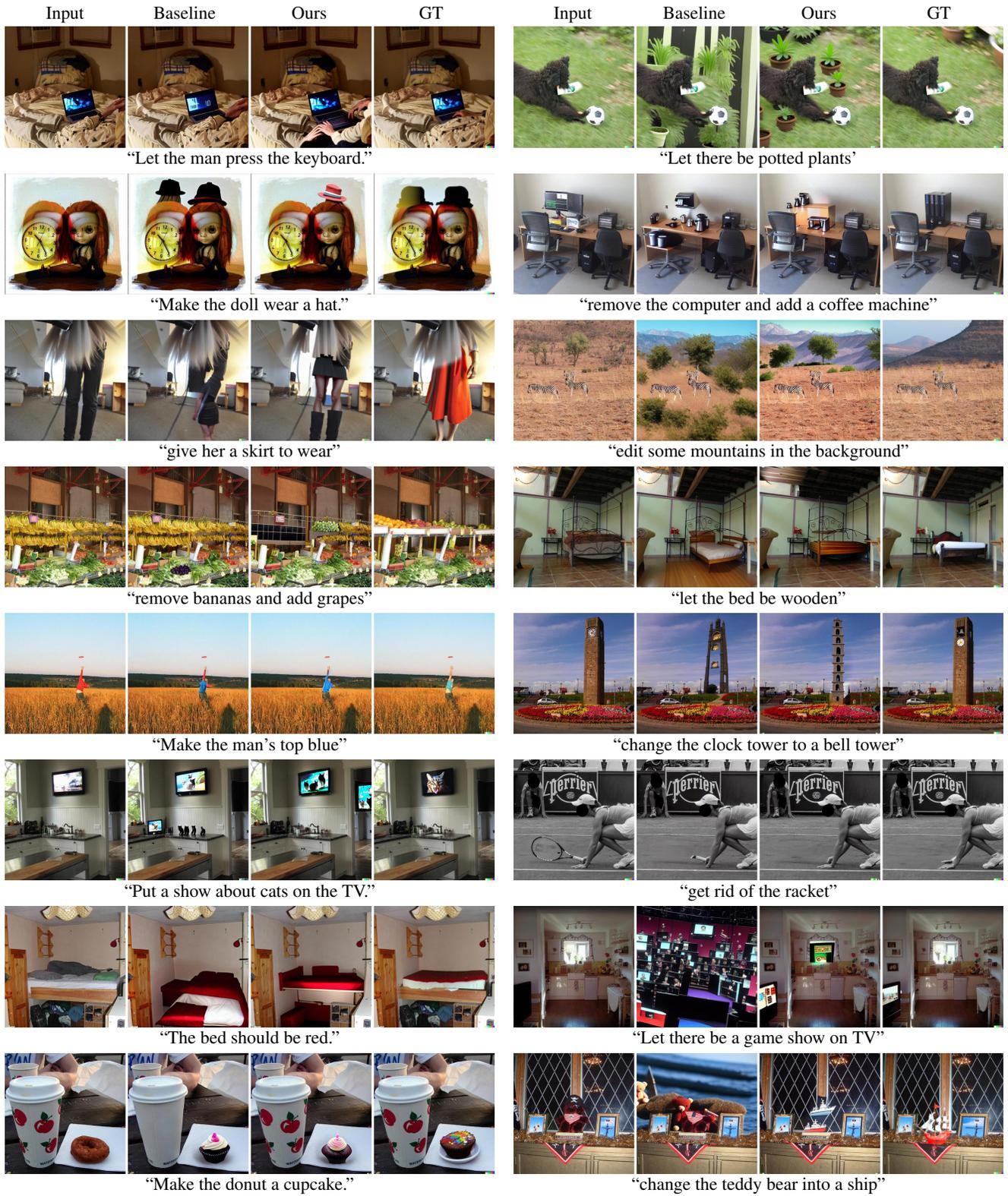

We include additional qualitative comparisons in Table~\ref{tab:image_editing_evaluation_correlation_supplementary} and~\ref{tab:image_editing_evaluation_accuracy_supplementary}. Notably, we include ImagenHub and AURORA-Bench (point-wise) results from baseline methods under the 1-shot setting defined in VIEScore~\cite{ku2023viescore}, where a single image editing evaluation example with ground-truth outputs is included in the prompt input. Additionally, we include more qualitative results to compare our scorer with propriety models for image editing evaluation (Fig.~\ref{fig:image_editing_evaluation_results_qualitative_supplementary}) as well as image editing comparisons between MagicBrush~\cite{zhang2024magicbrush} and our fine-tuned editing model in Fig.~\ref{fig:image_editing_reward_supplementary}.

\begin{table}[]
    \centering
    \small
    \begin{tabular}{lccc}
    \toprule
     Method           & Avg score  & Rank & GenAI-Arena rank \\
     \midrule
     MagicBrush       & 6.15       & 1    & 1 \\
     CosXL Edit       & 5.74       & 2    & 4 \\
     UltraEdit        & 4.63       & 3    & 2 \\
     InstructPix2Pix  & 4.18       & 4    & 5 \\
     Plug-and-Play    & 3.70       & 5    & 6 \\
     InfEdit          & 3.40       & 6    & 3 \\
     CycleDiffusion   & 3.20       & 7    & 8 \\
     Prompt-to-Prompt & 3.00       & 8    & 7 \\
     SDEdit           & 1.41       & 9    & 9 \\
     pix2pix-zero     & 0.71       & 10   & 10 \\
    \bottomrule
    \end{tabular}
    \caption{Editing models in GenAI-Arena leaderboard~\cite{jiang2024genai} ranked based on the average score of their outputs with respect to ImagenHub~\cite{ku2024imagenhub} assessed by our scorer. The resulting ranking is closely aligned with the GenAI-Arena ranking, even though our scorer has not been trained on samples from four of the ten models: CosXL Edit~\cite{cosxledit}, UltraEdit~\cite{zhao2024ultraedit}, Plug-and-Play~\cite{tumanyan2023plug}, and InfEdit~\cite{xu2023infedit}.}
    \label{tab:leaderboard}
\end{table}

Lastly, we take all the editing models listed on the GenAI-Arena leaderboard~\cite{jiang2024genai} and rank them based on the average score of outputs generated by each model with respect to ImagenHub samples, as assessed by our scorer. The resulting scores and ranking is shown in Tab.~\ref{tab:leaderboard}, where our ranking aligns closely with the one from GenAI-Arena, despite our scorer not being trained on samples from four of the ten models: CosXL Edit~\cite{cosxledit}, UltraEdit~\cite{zhao2024ultraedit}, Plug-and-Play~\cite{tumanyan2023plug}, and InfEdit~\cite{xu2023infedit}.

\begin{table}[]
\renewcommand{\arraystretch}{0.9} 
\setlength{\tabcolsep}{2pt}
\centering
\small
\resizebox{\linewidth}{!}{
\begin{tabular}{lcc}
\toprule
                        & ImagenHub & AURORA-Bench (point-wise) \\
\midrule
Human-to-Human          & 0.4184  &  -       \\
\midrule
CLIP-D                  & 0.2117  &  0.3080  \\ 
CLIP-T                  & 0.1894  &  0.1847  \\
CLIP-I                  & 0.1261  &  -       \\
DINO-I                  & 0.0441  &  -       \\ 
\midrule
GPT-4o                  & 0.3821  &  0.4038  \\
GPT-4o (1 shot)         & 0.3438  &  0.4779  \\
Gemini-Pro 1.5          & 0.2728  &  0.1052  \\
Gemini-Pro 1.5 (1 shot) & 0.2648  &  0.2315  \\
\midrule
LLaVA                   & 0.0273  &  0.0073  \\  
LLaVA (1 shot)          & 0.0258  &  -0.0110 \\ 
LLaVA-NeXT              & 0.0356  &  -0.0491 \\
LLaVA-NeXT (1 shot)     & 0.0468  &  0.0130  \\
LLaVA-OneVision         & 0.0829  &  0.0555  \\
LLaVA-OneVision (1 shot) & 0.3225 &  0.0896  \\
Qwen-VL                 & 0.0404  &  0.0118  \\
Qwen-VL (1 shot)        & 0.0037  &  0.0357  \\
Qwen2-VL                & 0.1445  &  0.1783  \\
Qwen2-VL (1 shot)       & 0.0914  &  0.1421  \\
Qwen2.5-VL              & 0.1859  &  0.2351  \\
Qwen2.5-VL (1 shot)     & 0.3467  &  0.2867  \\
Phi3.5-vision-instruct  & 0.1126  &  -0.0107 \\
Phi3.5-vision-instruct (1 shot) & 0.2605 & 0.0381 \\
Pixtral                 & 0.0123  &  -0.0005 \\
Pixtral (1 shot)        & 0.0243  &  -0.0005 \\
BLIP-2                  & 0.0378  & -0.0003  \\
BLIP-2 (1 shot)         & -0.0085 & 0.0011   \\
InstructBLIP            &  0.0212 & -0.0351  \\
Fuyu                    & 0.0206  & -0.0044  \\
CogVLM                  & -0.0288 & 0.0199   \\
OpenFlamingo            & -0.0577 & 0.0065   \\
\textbf{\ourwork{} (Ours)}       & 0.3450  & 0.4734   \\
\bottomrule
\end{tabular}}
\caption{Correlations of predicted scores with human ratings.
}
\label{tab:image_editing_evaluation_correlation_supplementary}
\end{table}

\begin{table}[]
\renewcommand{\arraystretch}{0.9} 
\setlength{\tabcolsep}{2pt}
\centering
\small
\resizebox{\linewidth}{!}{
\begin{tabular}{lcc}
\toprule
                & GenAI-Bench & AURORA-Bench (pair-wise) \\
\midrule
random          & 25.90       &  33.43 \\
\midrule
CLIP-D          & 43.09       &  31.63  \\
CLIP-T          & 39.39       &  42.93  \\
CLIP-I          & 38.96       & -       \\  
DINO-I          & 36.78       & -       \\
\midrule
GPT-4o          & 53.54       & 50.81  \\
Gemini-Pro 1.5  & \underline{55.93}       & 28.13  \\
LLaVA           & 26.12       & 27.50  \\
LLaVA-NeXT      & 25.35       & 27.19  \\
LLaVA-OneVision & 22.31       & 33.25  \\
LLaVA-OneVision (1 shot) & 22.31 & 33.25 \\
Qwen-VL         & 14.91       & 12.69  \\
Qwen2-VL        & 26.12       & 27.38  \\
Qwen2-VL (1 shot) & 26.12     & 27.38  \\
Qwen2.5-VL      & 32.10       & 30.69  \\
Qwen2.5-VL (1 shot) & 0.0     & 31.31  \\
Phi3.5-vision-instruct & 21.87 & 32.25 \\
Phi3.5-vision-instruct (1 shot) & 21.87 & 32.25 \\
Pixtral         & 26.12       & 27.38  \\
Pixtral (1 shot) & 26.12      & 27.38  \\
BLIP-2          & 26.01       & 26.25  \\
InstructBLIP    & 19.80       & 16.69  \\
Fuyu            & 0.0         & 0.0    \\
CogVLM          & 0.0         & 0.0    \\
OpenFlamingo    & 0.0         & 0.0    \\
\textbf{\ourwork{} (Ours)} & 59.41       & 52.88 \\
\bottomrule
\end{tabular}}
\caption{Accuracy of predicted comparison labels with human preference.}
\label{tab:image_editing_evaluation_accuracy_supplementary}
\end{table}

\begin{figure*}[]
    \centering
    \small
    \setlength{\tabcolsep}{2pt}
    \resizebox{\linewidth}{!}{
    \begin{tabular}{cc|cc|cc|cc}
    \multicolumn{1}{c}{Input} &  Output & \multicolumn{1}{c}{Input} &  Output & \multicolumn{1}{c}{Input} &  Output & \multicolumn{1}{c}{Input} &  Output \\ 
    
    \includegraphics[width=0.12\textwidth]{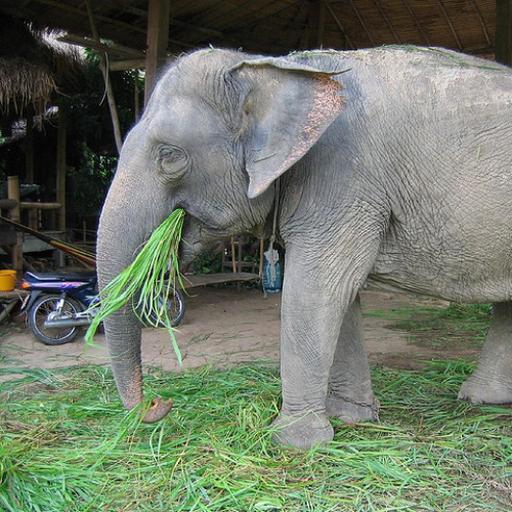} &  
    \includegraphics[width=0.12\textwidth]{fig/4_experiments/image_editing_evaluation/218_input.jpg} & 
    \includegraphics[width=0.12\textwidth]{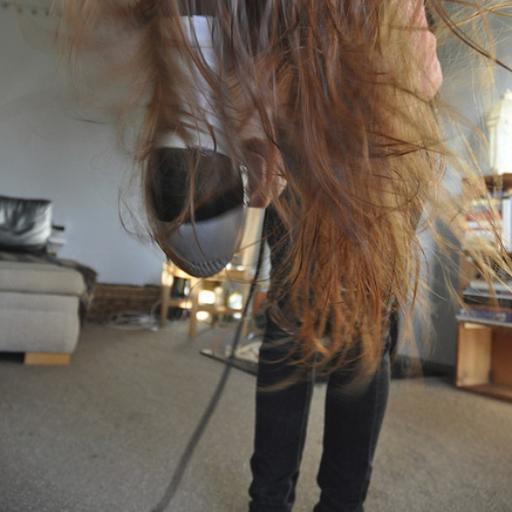} & 
    \includegraphics[width=0.12\textwidth]{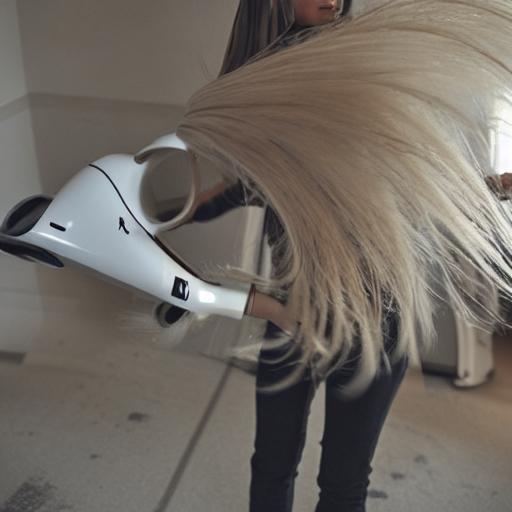} & 
    \includegraphics[width=0.12\textwidth]{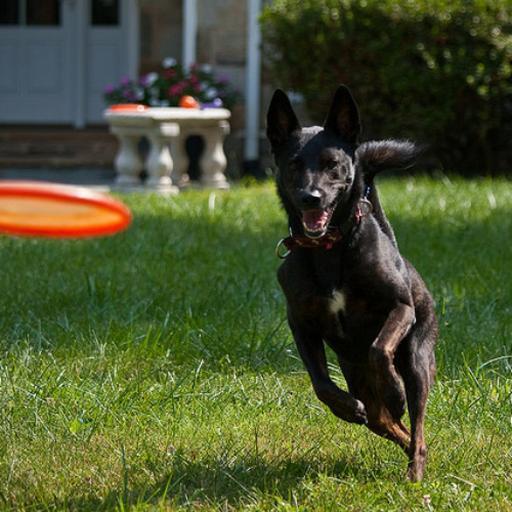} & 
    \includegraphics[width=0.12\textwidth]{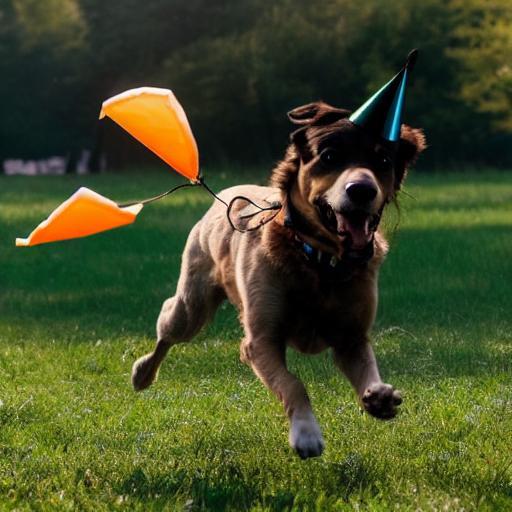} & 
    \begin{tikzpicture}
            \node[anchor=south west, inner sep=0] (image) at (0,0) {\includegraphics[width=0.12\textwidth]{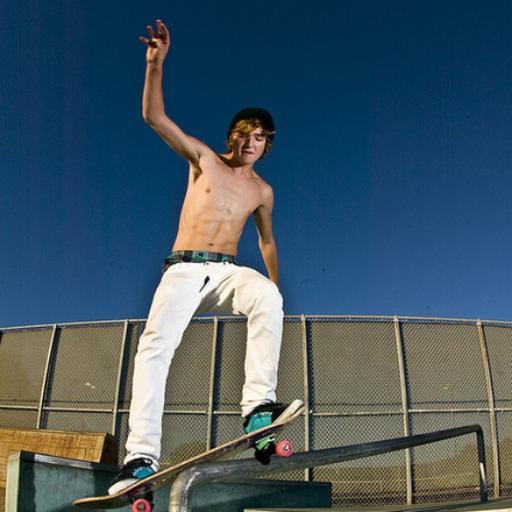}};
            \draw[fill=black, opacity=1.0] (1.05,1.55) circle [radius=0.12];
        \end{tikzpicture} & 
    \includegraphics[width=0.12\textwidth]{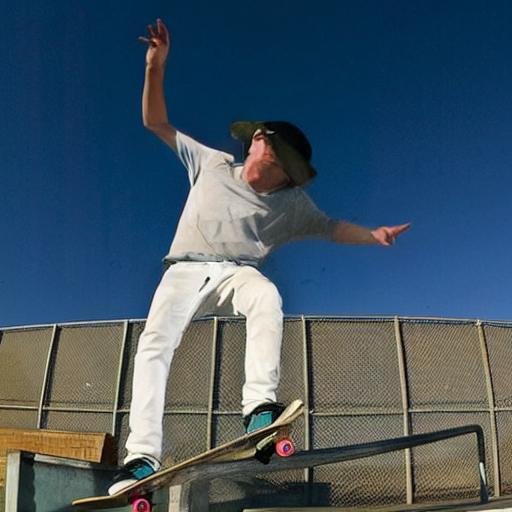} \\
    
    \multicolumn{2}{l|}{\makecell[l]{
    ``He should be eating a watermelon''
    }} & 
    \multicolumn{2}{l|}{\makecell[l]{
    ``turn her hair white''
    }} & 
    \multicolumn{2}{l|}{\makecell[l]{
    ``put a party hat on the dog''
    }} & 
    \multicolumn{2}{l}{\makecell[l]{
    ``What if the man had a hat?''
    }} \\

    \makecell[l]{\textbf{GT}: 0.0} & \makecell[l]{\textbf{Ours}: 0.94} & 
    \makecell[l]{\textbf{GT}: 0.0} & \makecell[l]{\textbf{Ours}: 2.06} & 
    \makecell[l]{\textbf{GT}: 0.0} & \makecell[l]{\textbf{Ours}: 2.71} &  
    \makecell[l]{\textbf{GT}: 2.36} & \makecell[l]{\textbf{Ours}: 5.39} \\

    \makecell[l]{\textbf{GPT-4o}: 1.41} & \makecell[l]{\textbf{Gemini}: 3.87} &
    \makecell[l]{\textbf{GPT-4o}: 4.90} & \makecell[l]{\textbf{Gemini}: 5.20} &
    \makecell[l]{\textbf{GPT-4o}: 5.65} & \makecell[l]{\textbf{Gemini}: 8.37} &
    \makecell[l]{\textbf{GPT-4o}: 6.0} & \makecell[l]{\textbf{Gemini}: 7.75} \\

    \midrule
    \includegraphics[width=0.12\textwidth]{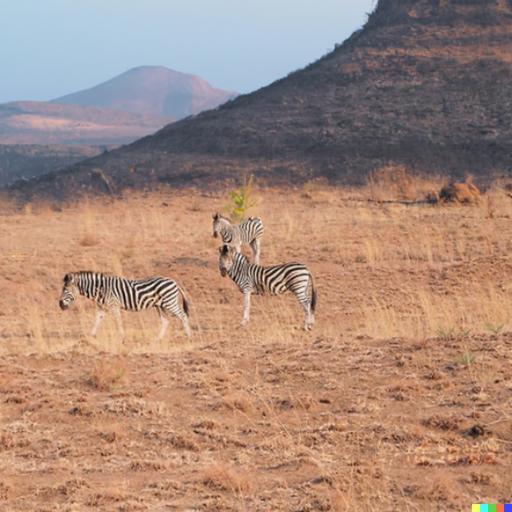} &  
    \includegraphics[width=0.12\textwidth]{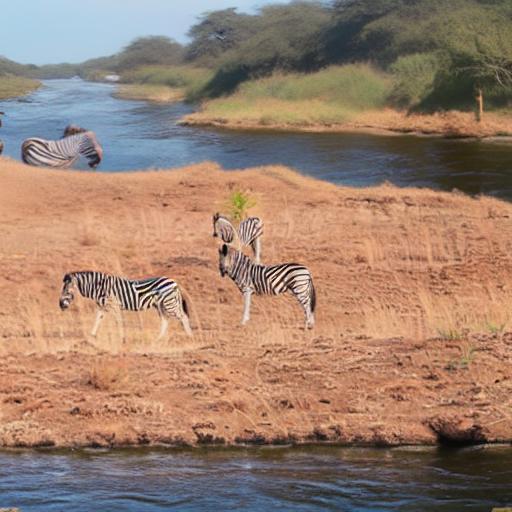} & 
    \includegraphics[width=0.12\textwidth]{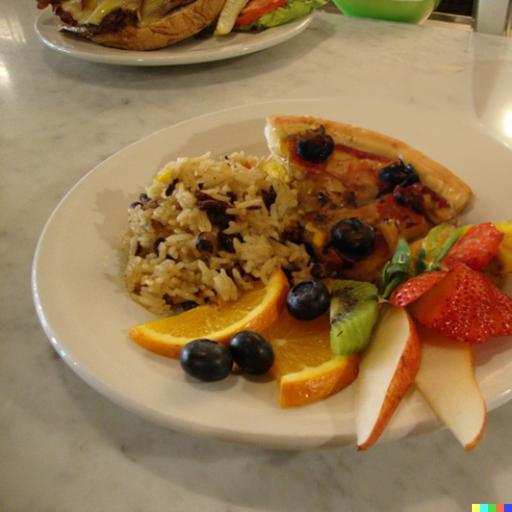} & 
    \includegraphics[width=0.12\textwidth]{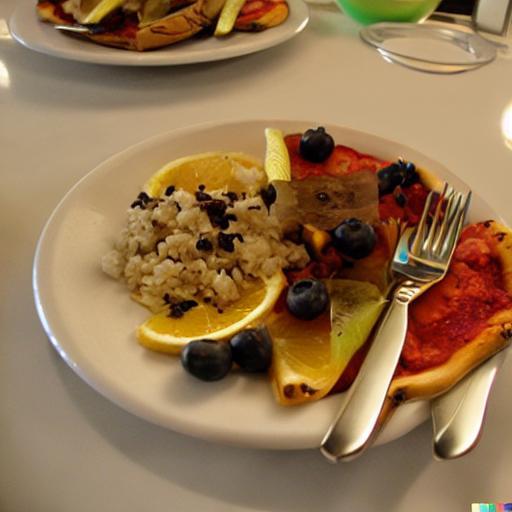} & 
    \includegraphics[width=0.12\textwidth]{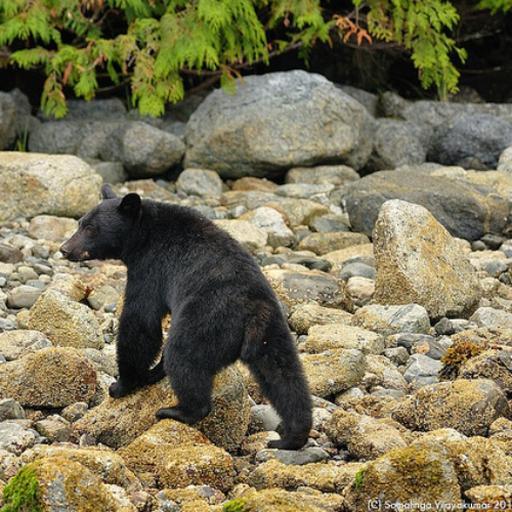} & 
    \includegraphics[width=0.12\textwidth]{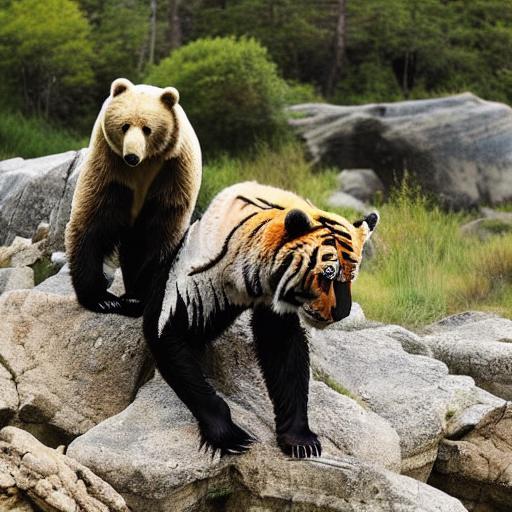} & 
    \begin{tikzpicture}
        \node[anchor=south west, inner sep=0] (image) at (0,0) {\includegraphics[width=0.12\textwidth]{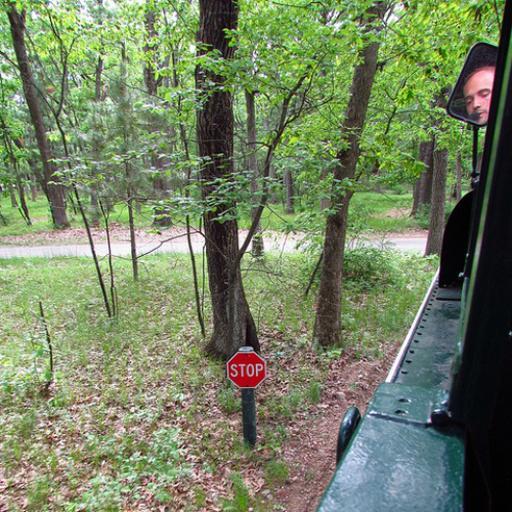}};
        \draw[fill=black, opacity=1.0] (2.,1.7) circle [radius=0.1];
    \end{tikzpicture} & 
    \begin{tikzpicture}
        \node[anchor=south west, inner sep=0] (image) at (0,0) {\includegraphics[width=0.12\textwidth]{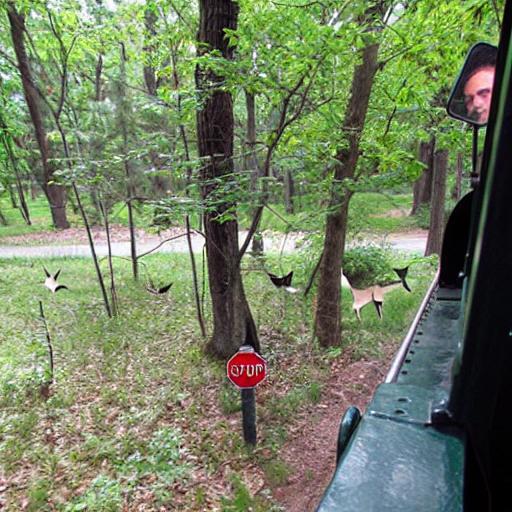}};
        \draw[fill=black, opacity=1.0] (2.,1.7) circle [radius=0.1];
    \end{tikzpicture} \\
    
    \multicolumn{2}{l|}{\makecell[l]{
    ``put the zebras next to a river''
    }} & 
    \multicolumn{2}{l|}{\makecell[l]{
    ``There should be some cutlery on\\ the table.''
    }} & 
    \multicolumn{2}{l|}{\makecell[l]{
    ``put a robot tiger next to the bear''
    }} & 
    \multicolumn{2}{l}{\makecell[l]{
    ``Add a dear on the grass.''
    }} \\

    \makecell[l]{\textbf{GT}: 5.0} & \makecell[l]{\textbf{Ours}: 6.09} & 
    \makecell[l]{\textbf{GT}: 6.67} & \makecell[l]{\textbf{Ours}: 5.94} & 
    \makecell[l]{\textbf{GT}: 0.0} & \makecell[l]{\textbf{Ours}: 0.23} &  
    \makecell[l]{\textbf{GT}: 3.33} & \makecell[l]{\textbf{Ours}: 3.67} \\

    \makecell[l]{\textbf{GPT-4o}: 3.74} & \makecell[l]{\textbf{Gemini}: 7.07} &
    \makecell[l]{\textbf{GPT-4o}: 4.90} & \makecell[l]{\textbf{Gemini}: 4.90} &
    \makecell[l]{\textbf{GPT-4o}: 2.45} & \makecell[l]{\textbf{Gemini}: 2.82} &
    \makecell[l]{\textbf{GPT-4o}: 0.0} & \makecell[l]{\textbf{Gemini}: 0.0} \\

    \midrule
    \includegraphics[width=0.12\textwidth]{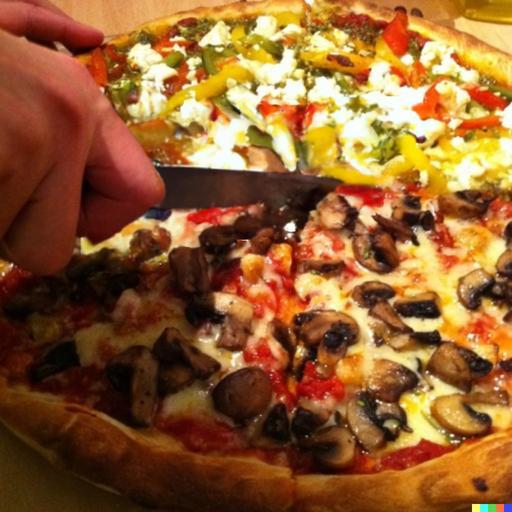} &  
    \includegraphics[width=0.12\textwidth]{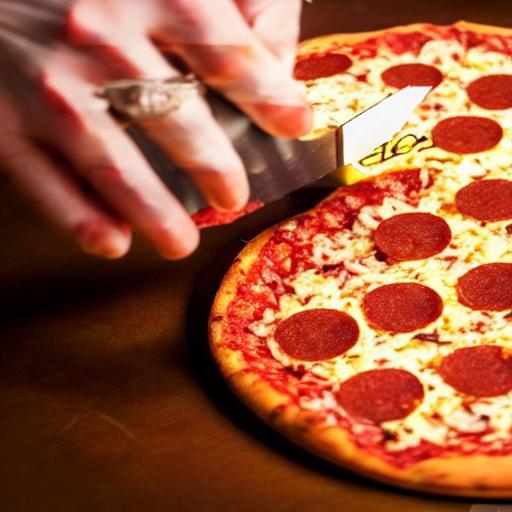} & 
    \includegraphics[width=0.12\textwidth]{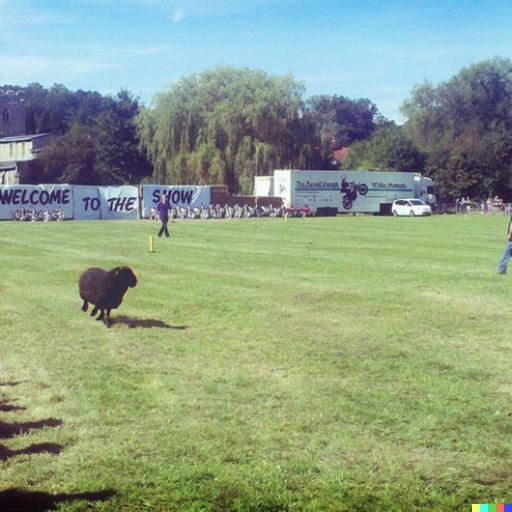} & 
    \includegraphics[width=0.12\textwidth]{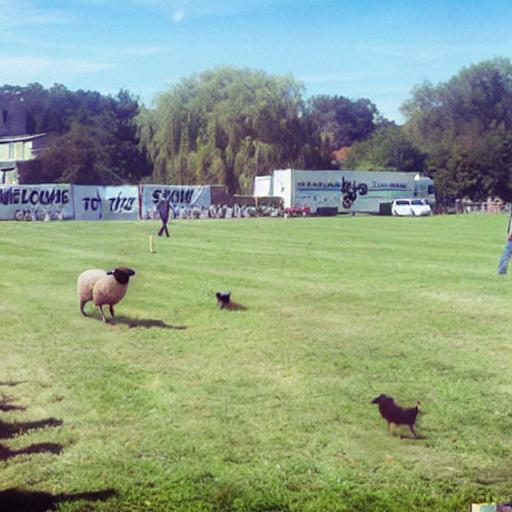} & 
    \includegraphics[width=0.12\textwidth]{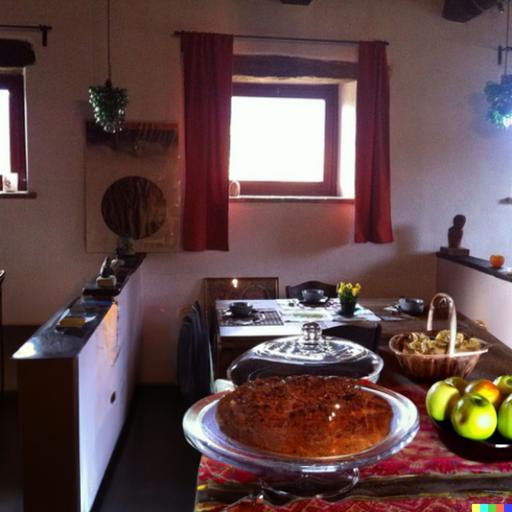} & 
    \includegraphics[width=0.12\textwidth]{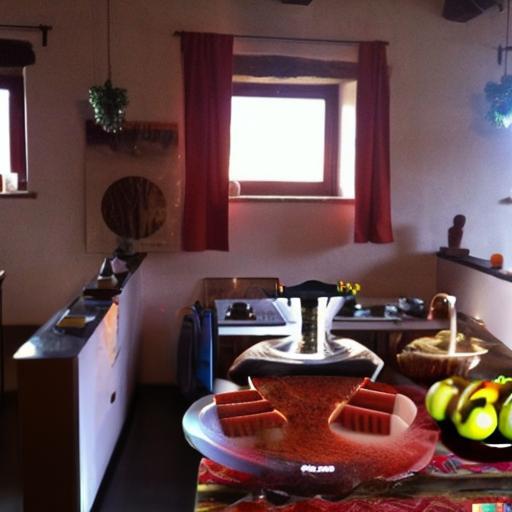} & 
    \begin{tikzpicture}
        \node[anchor=south west, inner sep=0] (image) at (0,0) {\includegraphics[width=0.12\textwidth]{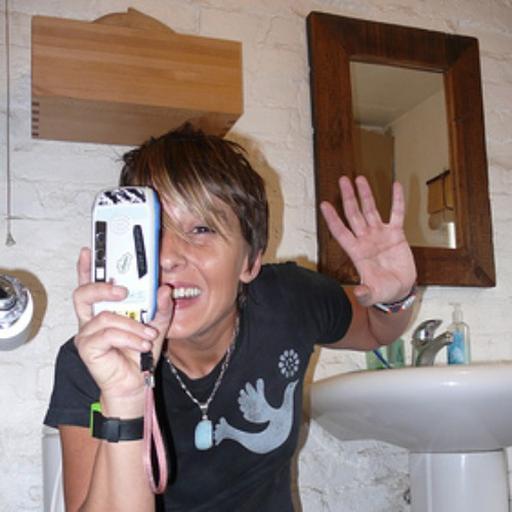}};
        \draw[fill=black, opacity=1.0] (0.9,1) circle [radius=0.25];
    \end{tikzpicture} & 
    \begin{tikzpicture}
        \node[anchor=south west, inner sep=0] (image) at (0,0) {\includegraphics[width=0.12\textwidth]{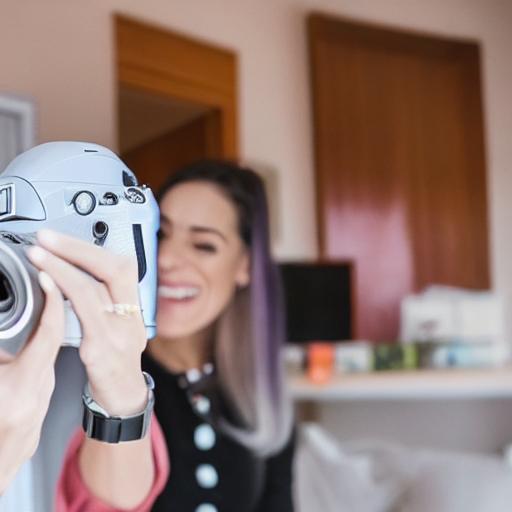}};
        \draw[fill=black, opacity=1.0] (0.9,1) circle [radius=0.25];
    \end{tikzpicture} \\
    
    \multicolumn{2}{l|}{\makecell[l]{
    ``make it a pepperoni pizza''
    }} & 
    \multicolumn{2}{l|}{\makecell[l]{
    ``A dog should be near the sheep.''
    }} & 
    \multicolumn{2}{l|}{\makecell[l]{
    ``Make the cake a chocolate cake''
    }} & 
    \multicolumn{2}{l}{\makecell[l]{
    ``make the woman hold a camera''
    }} \\

    \makecell[l]{\textbf{GT}: 0.0} & \makecell[l]{\textbf{Ours}: 1.66} & 
    \makecell[l]{\textbf{GT}: 3.33} & \makecell[l]{\textbf{Ours}: 5.63} & 
    \makecell[l]{\textbf{GT}: 0.0} & \makecell[l]{\textbf{Ours}: 2.37} &  
    \makecell[l]{\textbf{GT}: 0.0} & \makecell[l]{\textbf{Ours}: 1.17} \\

    \makecell[l]{\textbf{GPT-4o}: 4.90} & \makecell[l]{\textbf{Gemini}: 6.71} &
    \makecell[l]{\textbf{GPT-4o}: 5.66} & \makecell[l]{\textbf{Gemini}: 6.32} &
    \makecell[l]{\textbf{GPT-4o}: 3.46} & \makecell[l]{\textbf{Gemini}: 3.46} &
    \makecell[l]{\textbf{GPT-4o}: 2.65} & \makecell[l]{\textbf{Gemini}: 7.07} \\

    \midrule
    \begin{tikzpicture}
        \node[anchor=south west, inner sep=0] (image) at (0,0) {\includegraphics[width=0.12\textwidth]{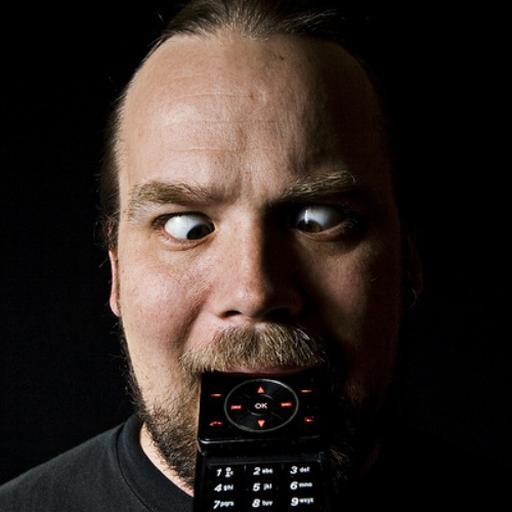}};
        \draw[fill=black, opacity=1.0] (1.,1) circle [radius=0.45];
    \end{tikzpicture} &  
    \begin{tikzpicture}
        \node[anchor=south west, inner sep=0] (image) at (0,0) {\includegraphics[width=0.12\textwidth]{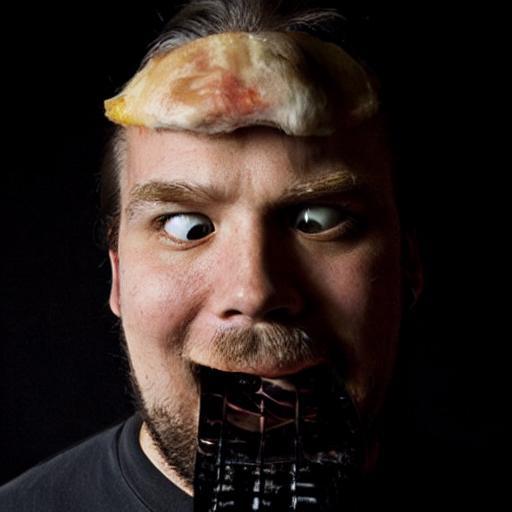}};
        \draw[fill=black, opacity=1.0] (1.,1) circle [radius=0.45];
    \end{tikzpicture} & 
    \includegraphics[width=0.12\textwidth]{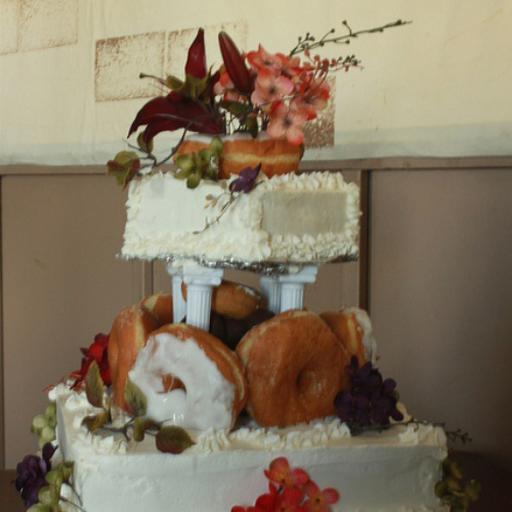} & 
    \includegraphics[width=0.12\textwidth]{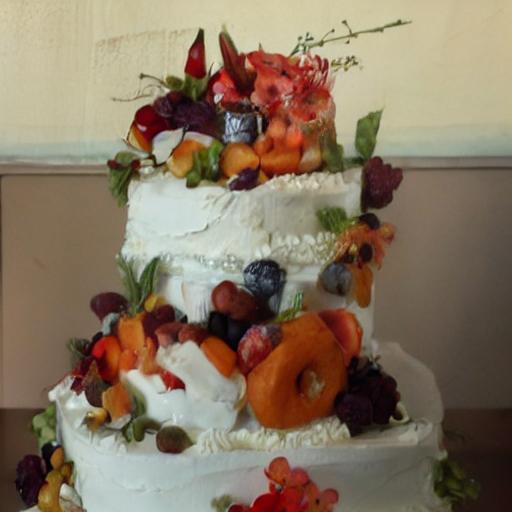} & 
    \includegraphics[width=0.12\textwidth]{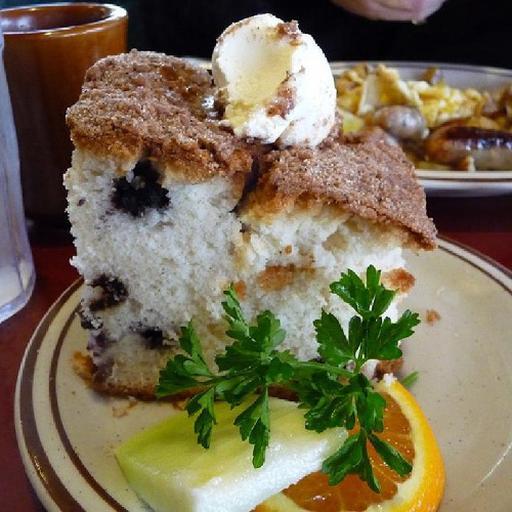} & 
    \includegraphics[width=0.12\textwidth]{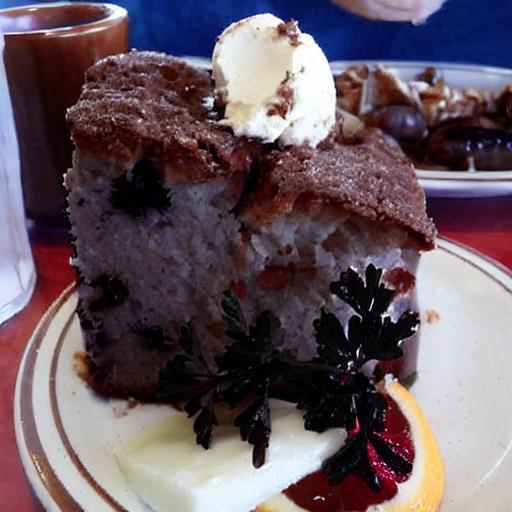} & 
    \includegraphics[width=0.12\textwidth]{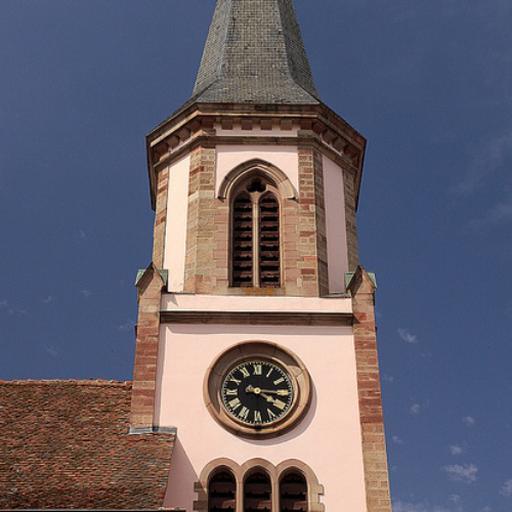} & 
    \includegraphics[width=0.12\textwidth]{fig/4_experiments/image_editing_evaluation/579_input.jpg} \\
    
    \multicolumn{2}{l|}{\makecell[l]{
    ``turn the remote into a pizza''
    }} & 
    \multicolumn{2}{l|}{\makecell[l]{
    ``replace the donuts with fruits''
    }} & 
    \multicolumn{2}{l|}{\makecell[l]{
    ``Let the bluebery cake be topped\\ with chocolate syrup.''
    }} & 
    \multicolumn{2}{l}{\makecell[l]{
    ``Let's add a cat on the roof.''
    }} \\

    \makecell[l]{\textbf{GT}: 0.0} & \makecell[l]{\textbf{Ours}: 1.29} & 
    \makecell[l]{\textbf{GT}: 9.02} & \makecell[l]{\textbf{Ours}: 6.76} & 
    \makecell[l]{\textbf{GT}: 7.07} & \makecell[l]{\textbf{Ours}: 6.41} &  
    \makecell[l]{\textbf{GT}: 3.33} & \makecell[l]{\textbf{Ours}: 4.41} \\

    \makecell[l]{\textbf{GPT-4o}: 2.45} & \makecell[l]{\textbf{Gemini}: 5.20} &
    \makecell[l]{\textbf{GPT-4o}: 6.0} & \makecell[l]{\textbf{Gemini}: 6.0} &
    \makecell[l]{\textbf{GPT-4o}: 2.83} & \makecell[l]{\textbf{Gemini}: 0.0} &
    \makecell[l]{\textbf{GPT-4o}: 0.0} & \makecell[l]{\textbf{Gemini}: 6.93} \\

    \midrule
    \includegraphics[width=0.12\textwidth]{fig/4_experiments/image_editing_evaluation/771_input.jpg} &  
    \includegraphics[width=0.12\textwidth]{fig/4_experiments/image_editing_evaluation/771_output.jpg} & 
    \includegraphics[width=0.12\textwidth]{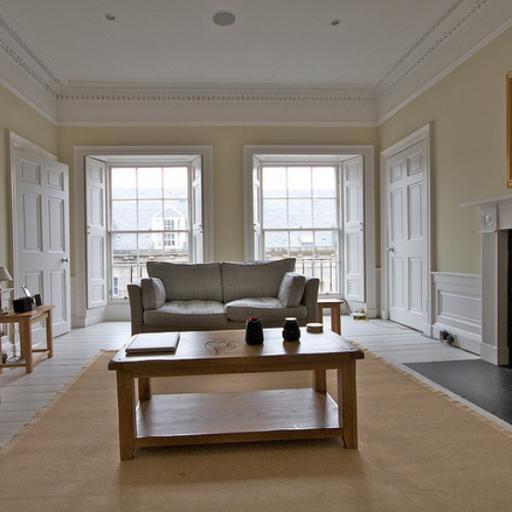} & 
    \includegraphics[width=0.12\textwidth]{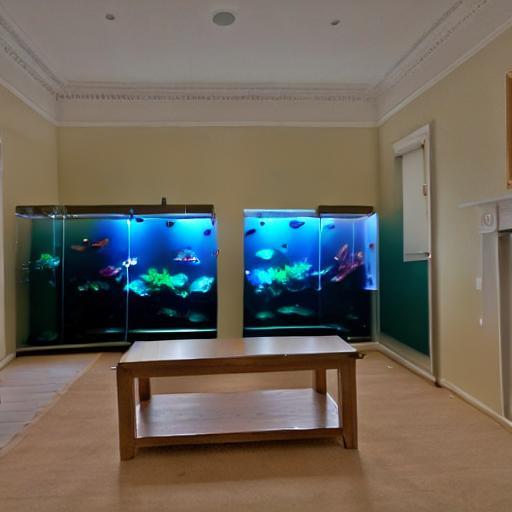} & 
    \includegraphics[width=0.12\textwidth]{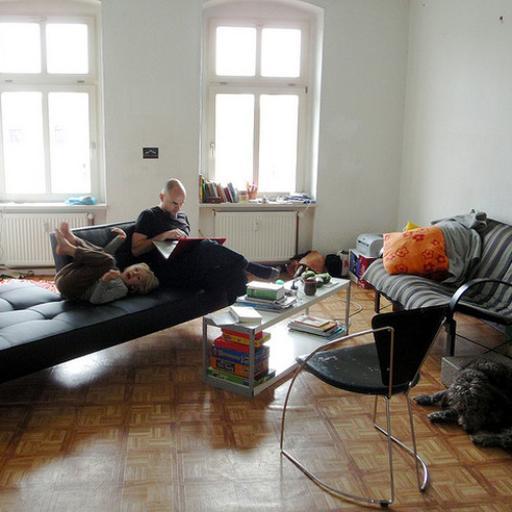} & 
    \includegraphics[width=0.12\textwidth]{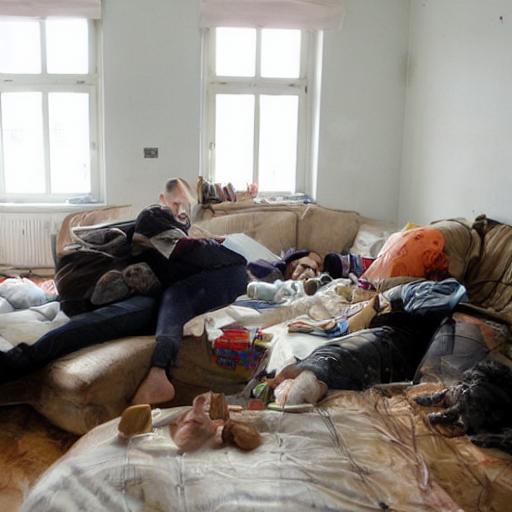} & 
    \includegraphics[width=0.12\textwidth]{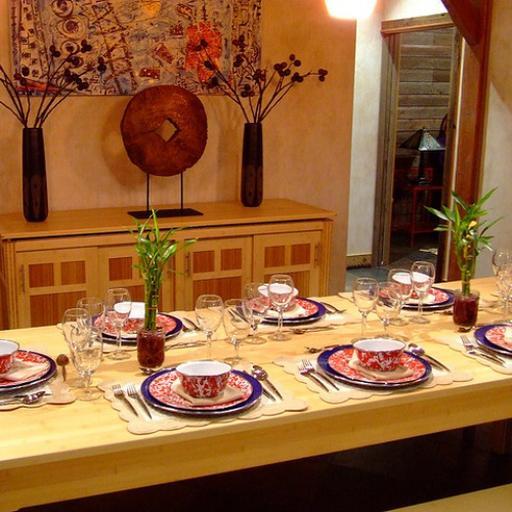} & 
    \includegraphics[width=0.12\textwidth]{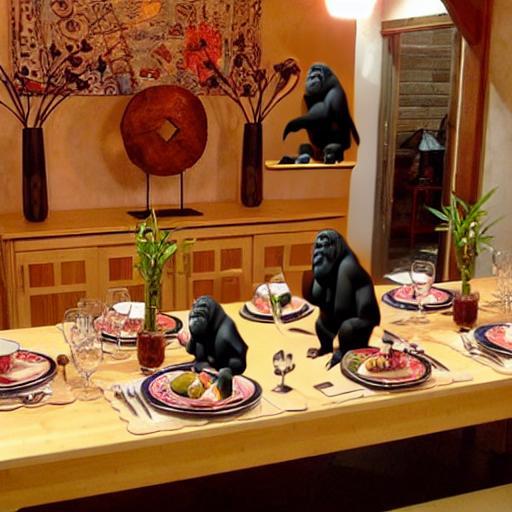} \\
    
    \multicolumn{2}{l|}{\makecell[l]{
    ``edit the background by removing\\ the museum and placing a castle''
    }} & 
    \multicolumn{2}{l|}{\makecell[l]{
    ``remove the table and add an\\ aquarium''
    }} & 
    \multicolumn{2}{l|}{\makecell[l]{
    ``let the kid sleep''
    }} & 
    \multicolumn{2}{l}{\makecell[l]{
    ``Have a gorilla sit at the dinner\\ table.''
    }} \\

    \makecell[l]{\textbf{GT}: 6.38} & \makecell[l]{\textbf{Ours}: 6.09} & 
    \makecell[l]{\textbf{GT}: 5.0} & \makecell[l]{\textbf{Ours}: 4.49} & 
    \makecell[l]{\textbf{GT}: 0.0} & \makecell[l]{\textbf{Ours}: 1.45} &  
    \makecell[l]{\textbf{GT}: 1.67} & \makecell[l]{\textbf{Ours}: 5.47} \\

    \makecell[l]{\textbf{GPT-4o}: 5.66} & \makecell[l]{\textbf{Gemini}: 5.66} &
    \makecell[l]{\textbf{GPT-4o}: 0.0} & \makecell[l]{\textbf{Gemini}: 0.0} &
    \makecell[l]{\textbf{GPT-4o}: 3.16} & \makecell[l]{\textbf{Gemini}: 7.75} &
    \makecell[l]{\textbf{GPT-4o}: 5.66} & \makecell[l]{\textbf{Gemini}: 6.32} \\

    \midrule
    \includegraphics[width=0.12\textwidth]{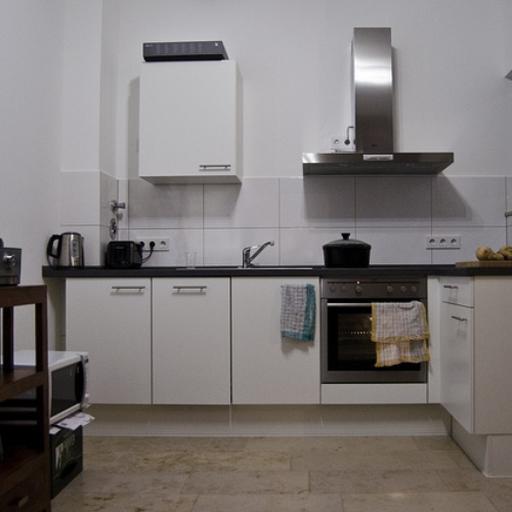} &  
    \includegraphics[width=0.12\textwidth]{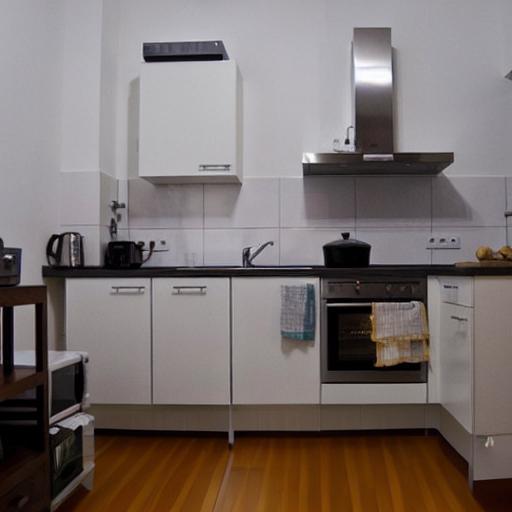} & 
    \includegraphics[width=0.12\textwidth]{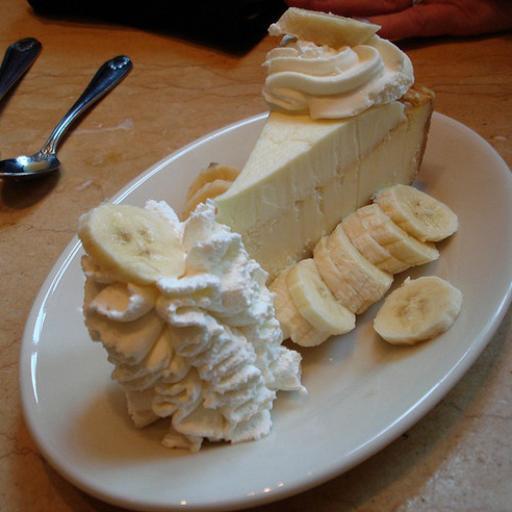} & 
    \includegraphics[width=0.12\textwidth]{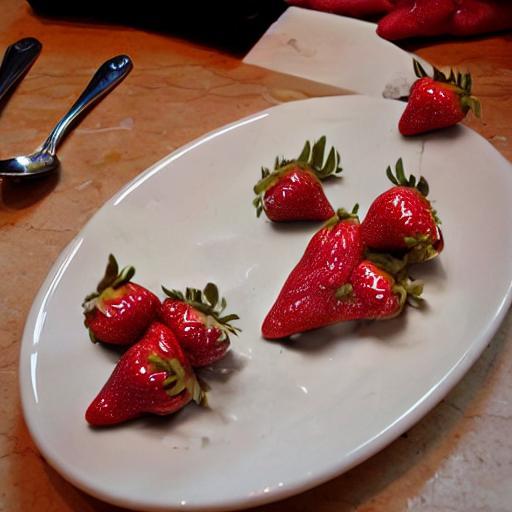} & 
    \includegraphics[width=0.12\textwidth]{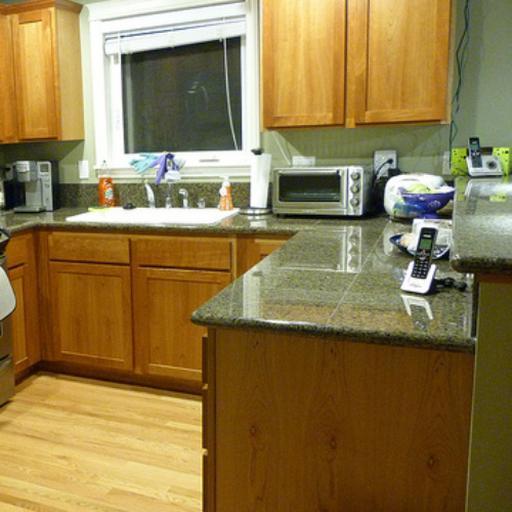} & 
    \includegraphics[width=0.12\textwidth]{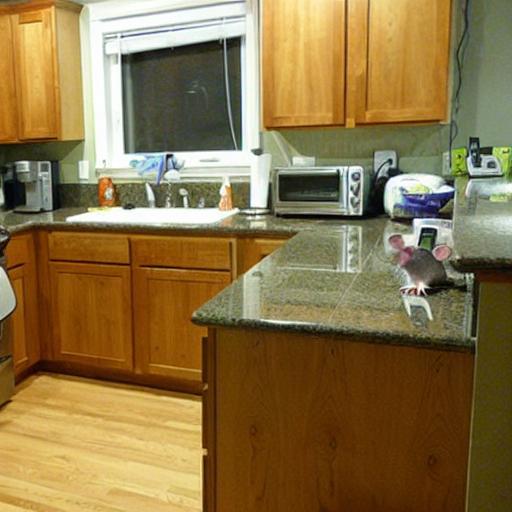} & 
    \includegraphics[width=0.12\textwidth]{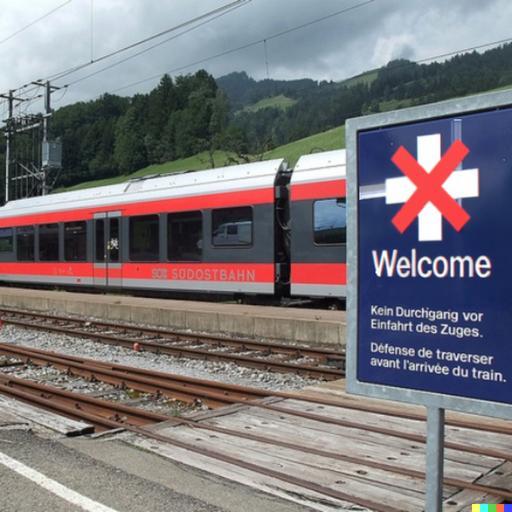} & 
    \includegraphics[width=0.12\textwidth]{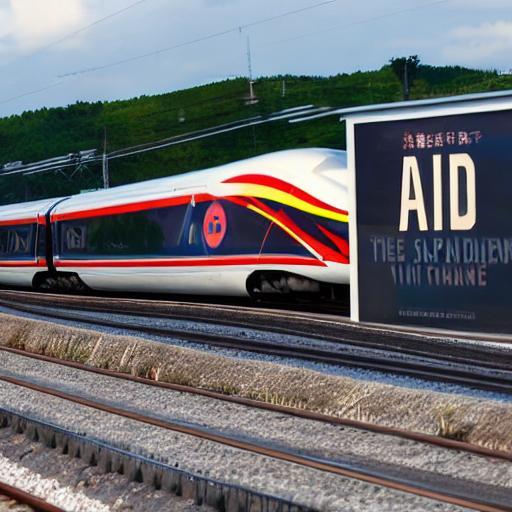} \\
    
    \multicolumn{2}{l|}{\makecell[l]{
    ``Put a wooden floor on the kitchen.''
    }} & 
    \multicolumn{2}{l|}{\makecell[l]{
    ``put strawberry on the plate''
    }} & 
    \multicolumn{2}{l|}{\makecell[l]{
    ``Put a rat on the counter.''
    }} & 
    \multicolumn{2}{l}{\makecell[l]{
    `let it be a bullet train''
    }} \\

    \makecell[l]{\textbf{GT}: 10.0} & \makecell[l]{\textbf{Ours}: 8.59} & 
    \makecell[l]{\textbf{GT}: 6.38} & \makecell[l]{\textbf{Ours}: 4.20} & 
    \makecell[l]{\textbf{GT}: 5.69} & \makecell[l]{\textbf{Ours}: 5.63} &  
    \makecell[l]{\textbf{GT}: 0.0} & \makecell[l]{\textbf{Ours}: 2.97} \\

    \makecell[l]{\textbf{GPT-4o}: 8.49} & \makecell[l]{\textbf{Gemini}: 7.75} &
    \makecell[l]{\textbf{GPT-4o}: 3.0} & \makecell[l]{\textbf{Gemini}: 0.0} &
    \makecell[l]{\textbf{GPT-4o}: 6.71} & \makecell[l]{\textbf{Gemini}: 7.07} &
    \makecell[l]{\textbf{GPT-4o}: 5.92} & \makecell[l]{\textbf{Gemini}: 6.48} \\
    
    \end{tabular}}
    \caption{More evaluation examples (faces are blocked due to privacy concerns) from GPT-4o~\cite{achiam2023gpt}, Gemini-Pro 1.5 (Gemini)~\cite{team2024gemini}, and our method on ImagenHub~\cite{ku2024imagenhub}, where the ground-truth (GT) scores are presented below edit instructions.}
    \label{fig:image_editing_evaluation_results_qualitative_supplementary}
\end{figure*}

\end{document}